\documentclass{article} 
\usepackage{iclr2023_conference,times}

\usepackage{amsmath,amsfonts,bm}

\def\eqref#1{equation~\ref{#1}}

\def\1{\bm{1}}

\DeclareMathAlphabet{\mathsfit}{\encodingdefault}{\sfdefault}{m}{sl}
\SetMathAlphabet{\mathsfit}{bold}{\encodingdefault}{\sfdefault}{bx}{n}

\DeclareMathOperator*{\argmax}{arg\,max}
\DeclareMathOperator*{\argmin}{arg\,min}

\newcommand{\eat}[1]{}

\usepackage{multirow}
\usepackage{balance}

\usepackage{subfig}
\usepackage{algorithm}
\usepackage{bm}
\usepackage{algpseudocode}
\usepackage{amsthm,amsmath,amssymb}
\usepackage{mathrsfs}

\algnewcommand{\Initialize}[1]{%
  \State \textbf{Initialize:}
  \Statex \hspace*{\algorithmicindent}\parbox[t]{.8\linewidth}{\raggedright #1}
}
\algnewcommand{\Inputs}[1]{%
  \State \textbf{Inputs:}
  \Statex \hspace*{\algorithmicindent}\parbox[t]{.8\linewidth}{\raggedright #1}
}
\algnewcommand{\Outputs}[1]{%
  \State \textbf{Outputs:}
  \Statex \hspace*{\algorithmicindent}\parbox[t]{.8\linewidth}{\raggedright #1}
}
\newtheorem{theorem}{Theorem}[section]
\newtheorem*{remark}{Remark}

\newtheorem{definition}{Definition}

\usepackage{hyperref}
\usepackage{url}
\usepackage{graphicx}
\usepackage{amsmath}
\usepackage{booktabs}
\usepackage{multirow}
\usepackage{balance}
\usepackage{wrapfig,lipsum}

\usepackage{subfig}

\usepackage{mathtools}
\DeclarePairedDelimiterX{\infdivx}[2]{(}{)}{%
  #1\;\delimsize\|\;#2%
}
\newcommand{\infdiv}{D_{KL}\infdivx}

\title{Exploring Active 3D Object Detection from a Generalization Perspective}

\author{Yadan Luo$^*$, Zhuoxiao Chen\thanks{Equal contribution. Correspondence to Yadan Luo $<$y.luo@uq.edu.au$>$.}~~, Zijian Wang, Xin Yu, Zi Huang, Mahsa Baktashmotlagh \\
The University of Queensland, Australia\\
}

\iclrfinalcopy 
\begin{document}

\maketitle

\begin{abstract}\vspace{-1.5ex}
To alleviate the high annotation cost in LiDAR-based 3D object detection, active learning is a promising solution that learns to select only a small portion of unlabeled data to annotate, without compromising model performance. Our empirical study, however, suggests that mainstream uncertainty-based and diversity-based active learning policies are not effective when applied in the 3D detection task, as they \eat{overlook the generalization performance and }fail to balance the trade-off between point cloud informativeness and box-level annotation costs. To overcome this limitation, we jointly investigate three novel criteria in our framework \textbf{\textsc{Crb}} for point cloud acquisition - \textit{label \underline{c}onciseness}, \textit{feature \underline{r}epresentativeness} and \textit{geometric \underline{b}alance}, which hierarchically filters out the point clouds of redundant 3D bounding box labels, latent features and geometric characteristics (\textit{e.g.}, point cloud density) from the unlabeled sample pool and greedily selects informative ones with fewer objects to annotate. Our theoretical analysis demonstrates that the proposed criteria aligns the marginal distributions of the selected subset and the prior distributions of the unseen test set, and minimizes the upper bound of the generalization error. To validate the effectiveness and applicability of \textsc{Crb}, we conduct extensive experiments on the two benchmark 3D object detection datasets of KITTI and Waymo and examine both one-stage (\textit{i.e.}, \textsc{Second}) and two-stage 3D detectors (\textit{i.e.}, \textsc{Pv-rcnn}). Experiments evidence that the proposed approach outperforms existing active learning strategies and achieves fully supervised performance requiring $1\%$ and $8\%$ annotations of bounding boxes and point clouds, respectively. Source code: \href{https://github.com/Luoyadan/CRB-active-3Ddet}{https://github.com/Luoyadan/CRB-active-3Ddet}.
\end{abstract}

\section{Introduction}\vspace{-2ex}
LiDAR-based 3D object detection plays an indispensable role in 3D scene understanding with a wide range of applications such as autonomous driving \citep{DBLP:conf/nips/DengQNFZA21,DBLP:conf/eccv/WangLGD20} and robotics \citep{DBLP:conf/iros/AhmedTCMW18, DBLP:conf/iros/MontesLCD20, DBLP:journals/sensors/WangLSLZSQT19}. The emerging stream of 3D detection models enables accurate recognition at the cost of large-scale labeled point clouds, where 7-degree of freedom (DOF) 3D bounding boxes - consisting of a position, size, and orientation information- for each object are annotated. In the benchmark datasets like Waymo \citep{DBLP:conf/cvpr/SunKDCPTGZCCVHN20}, there are over 12 million LiDAR boxes, for which, labeling a precise 3D box takes more than 100 seconds for an annotator \citep{DBLP:conf/cvpr/SongLX15}. This prerequisite for the performance boost greatly hinders the feasibility of applying models to the wild, especially when the annotation budget is limited.

To alleviate this limitation, active learning (AL) aims to reduce labeling costs by querying labels for only a small portion of unlabeled data. The criterion-based query selection process iteratively selects the most beneficial samples for the subsequent model training until the labeling budget is run out. The criterion is expected to quantify the sample informativeness using the heuristics derived from \textit{sample uncertainty} \citep{DBLP:conf/icml/GalIG17, DBLP:conf/iccv/DuZCC0021, DBLP:conf/cvpr/CaramalauBK21, DBLP:conf/cvpr/YuanWFLXJY21, DBLP:conf/iccv/ChoiELFA21, DBLP:conf/cvpr/Zhang0YWZH20, DBLP:conf/ijcai/ShiL19} and \textit{sample diversity} \citep{DBLP:conf/iclr/MaZMS21, DBLP:conf/cvpr/GudovskiyHYT20, DBLP:conf/eccv/GaoZYADP20, DBLP:conf/iccv/SinhaED19, DBLP:conf/nips/Pinsler0NH19}. In particular, uncertainty-driven approaches focus on the samples that the model is the least confident of their labels, thus searching for the candidates with: maximum entropy \citep{DBLP:journals/neco/MacKay92b,DBLP:journals/bstj/Shannon48,DBLP:conf/nips/KimSJM21, DBLP:conf/cvpr/SiddiquiVN20, DBLP:conf/nips/Shi019}, disagreement among different experts \citep{DBLP:conf/nips/FreundSST92, DBLP:conf/icml/TranDRC19}, minimum posterior probability of a predicted class \citep{DBLP:journals/tcsv/WangZLZL17}, or the samples with reducible yet maximum estimated error \citep{DBLP:conf/icml/RoyM01,DBLP:conf/cvpr/YooK19, DBLP:conf/cvpr/KimPKC21}. On the other hand, diversity-based methods try to find the most representative samples to avoid sample redundancy. To this end, they form subsets that are sufficiently diverse to describe the entire data pool by making use of the greedy coreset algorithms~\citep{DBLP:conf/iclr/SenerS18}, or the clustering algorithms \citep{DBLP:conf/icml/NguyenS04}. Recent works \citep{DBLP:conf/iccv/LiuDZLDH21, DBLP:conf/nips/CitovskyDGKRRK21, DBLP:conf/nips/KirschAG19, DBLP:journals/corr/abs-1112-5745} combine the aforementioned heuristics: they measure uncertainty as the gradient magnitude of samples \citep{DBLP:conf/iclr/AshZK0A20} or its second-order metrics \citep{DBLP:conf/iccv/LiuDZLDH21} at the final layer of neural networks, and then select samples with gradients spanning a diverse set of directions. While effective, the hybrid approaches commonly cause heavy computational overhead, since gradient computation is required for each sample in the unlabeled pool. Another stream of works apply active learning to 2D/3D object detection tasks\textcolor{black}{~\citep{DBLP:conf/ivs/FengWRMD19, DBLP:conf/ivs/SchmidtRTK20, wang2022weaklySupervisedObject, DBLP:conf/cvpr/WuC022, 9548667}}, by leveraging ensemble \citep{DBLP:conf/cvpr/BeluchGNK18} or Monte Carlo (MC) dropout \citep{DBLP:conf/icml/GalG16} algorithms to estimate the classification and localization uncertainty of bounding boxes for images/point clouds acquisition (more details in \textcolor{black}{Appendix I}). Nevertheless, those AL methods generally favor the point clouds with more objects, which have a higher chance of containing uncertain and diverse objects. With a fixed annotation budget, it is far from optimal to select such point clouds, since more clicks are required to form 3D box annotations.


To overcome the above limitations, we propose to learn AL criteria for cost-efficient sample acquisition at the 3D box level by empirically studying its relationship with optimizing the generalization upper bound. Specifically, we propose three selection criteria for cost-effective point cloud acquisition, termed as \textsc{Crb}, \textit{i.e.,} \textit{label \underline{c}onciseness}, \textit{feature \underline{r}epresentativeness} and \textit{geometric \underline{b}alance}.  Specifically, we divide the sample selection process into three stages: (1) To alleviate the issues of label redundancy and class imbalance, and to ensure \textit{label conciseness}, we firstly calculate the entropy of bounding box label predictions and only pick top $\mathcal{K}_1$ point clouds for Stage 2; (2) We then examine the \textit{feature representativeness} of candidates by formulating the task as the $\mathcal{K}_2$-medoids problem on the gradient space. To jointly consider the impact of classification and regression objectives on gradients, we enable the Monte Carlo dropout (\textsc{Mc-dropout}) and construct the hypothetical labels by averaging predictions from multiple stochastic forward passes. (3) Finally, to maintain the \textit{geometric balance} property, we minimize the KL divergence between the marginal distributions of point cloud density of each predicted bounding box. This makes the trained detector predict more accurate localization and size of objects, and recognize both close (\textit{i.e.}, dense) and distant (\textit{i.e.}, sparse) objects at the test time, using minimum number of annotations. We base our criterion design on our theoretical analysis of optimizing the upper bound of the generalization risk, which can be reformulated as distribution alignment of the selected subset and the test set. Note that since the empirical distribution of the test set is not observable during training, WLOG, we make an appropriate assumption of its prior distribution.

\noindent\textbf{Contributions}. Our work is a pioneering study in active learning for 3D object detection, aiming to boost the detection performance at the \textbf{lowest cost of bounding box-level annotations}. To this end, we propose a hierarchical active learning scheme for 3D object detection, which progressively filters candidates according to the derived selection criteria without triggering heavy computation. Extensive experiments conducted demonstrate that the proposed \textsc{Crb} strategy can consistently outperform all the state-of-the-art AL baselines on two large-scale 3D detection datasets irrespective of the detector architecture. To enhance the reproducibility of our work and accelerate future work in this new research direction, we develop a \texttt{active-3D-det} toolbox, which accommodates various AL approaches and 3D detectors. The source code is available in the supplementary material, and will be publicly shared upon acceptance of the paper.

\vspace{-2ex}
\section{Methodology}\vspace{-2ex}
\subsection{Problem Formulation}
\begin{figure}[!t]\vspace{-7ex}
\centering
\includegraphics[width=1\linewidth]{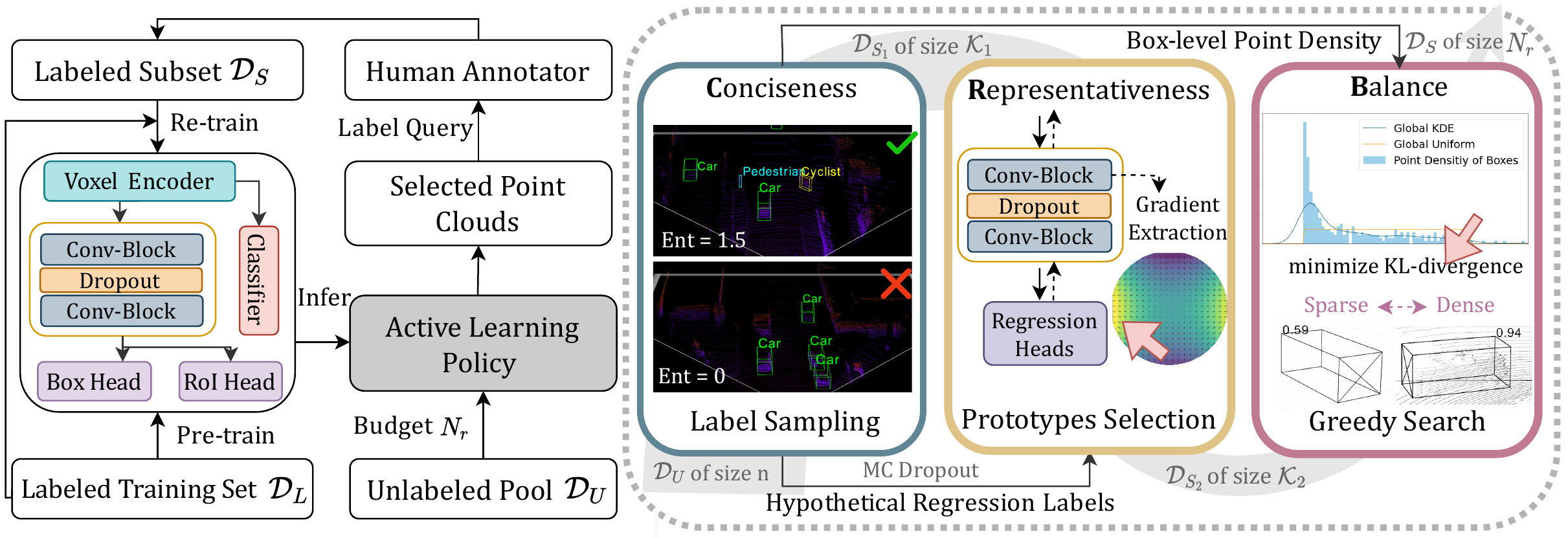}\vspace{-2ex}
\caption{An illustrative flowchart of the proposed \textsc{Crb} framework for active selection of point clouds. Motivated by optimizing the generalization risk, the derived strategy hierarchically selects point clouds that have non-redundant bounding box labels, latent gradients and geometric characteristics to mitigate the gap with the test set and minimize annotation costs.\vspace{-1ex}}
\label{fig:flowchart}
\vspace{-2ex}
\end{figure}

\vspace{-2ex}
In this section, we mathematically formulate the problem of active learning for 3D object detection and set up the notations. Given an orderless LiDAR point cloud $\mathcal{P} = \{x, y, z, e\}$ with 3D location $(x, y, z)$ and reflectance $e$, the goal of 3D object detection is to localize the objects of interest as a set of 3D bounding boxes $\mathcal{B} = \{b_k\}_{k\in[N_B]}$ with $N_B$ indicating the number of detected bounding boxes, and predict the associated box labels $Y = \{y_k\}_{k\in[N_B]} \in\mathcal{Y} = \{1,\ldots,C\}$, with $C$ being the number of classes to predict. Each bounding box $b$ represents the relative center position $(p_x, p_y, p_z)$ to the object ground planes, the box size $(l, w, h)$, and the heading angle $\theta$. Mainstream 3D object detectors \textcolor{black}{use point clouds $\mathcal{P}$ to extract point-level features $\bm{x}\in\mathbb{R}^{W\cdot L\cdot  F}$ ~}\citep{DBLP:conf/cvpr/ShiWL19,DBLP:conf/iccv/YangS0SJ19,DBLP:conf/cvpr/YangS0J20} or by voxelization \citep{DBLP:conf/cvpr/ShiGJ0SWL20}, with $W$, $L$, $F$ representing width, length, and channels of the feature map. The feature map $\bm{x}$ is passed to a classifier $f(\cdot; \bm{w}_{f})$ parameterized by $\bm{w}_{f}$ and regression heads $g(\cdot; \bm{w}_{g})$ (\textit{e.g.,} box refinement and ROI regression) parameterized by $\bm{w}_{g}$. The output of the model is the detected bounding boxes $\widehat{\mathcal{B}} = \{\hat{b}_k\}$ with the associated box labels $\widehat{Y}=\{\hat{y}_k\}$ from anchored areas. \textcolor{black}{The loss functions $\ell^{cls}$ and $\ell^{reg}$ for classification (\textit{e.g.}, regularized cross entropy loss \cite{DBLP:journals/corr/abs-1808-09540}) and regression (\textit{e.g.}, mean absolute error/$L_1$ regularization \cite{DBLP:journals/spl/QiDSML20}) are assumed to be Lipschitz continuous.} As shown in the left half of Figure \ref{fig:flowchart}, in an active learning pipeline, a small set of labeled point clouds $\mathcal{D}_L=\{(\mathcal{P}, \mathcal{B}, Y)_i\}_{i\in[m]}$ and a large pool of raw point clouds $\mathcal{D}_U=\{(\mathcal{P})_j\}_{j\in[n]}$ are provided at training time, with $n$ and $m$ being a total number of point clouds and $m\ll n$. For each active learning round $r\in[R]$, and based on the criterion defined by an active learning policy, we select a subset of raw data $\{\mathcal{P}_j\}_{j\in[N_r]}$ from $\mathcal{D}_U$ and query the labels of 3D bounding boxes from an oracle $\bm{\Omega}: \mathcal{P}\rightarrow \mathcal{B}\times\mathcal{Y}$ to construct $\mathcal{D}_S=\{(\mathcal{P}, \mathcal{B}, Y)_j\}_{j\in[N_r]}$. The 3D detection model is pre-trained with $\mathcal{D}_L$ for active selection, and then retrained with $\mathcal{D}_{S}\cup \mathcal{D}_L$ until the selected samples reach the final budget $B$, \textit{i.e.,} $\sum_{r=1}^{R}N_{r} = B$.

\subsection{Theoretical Motivation}\label{sec:motivation}
\vspace{-1ex}
The core question of active 3D detection is how to design a proper criterion, based on which a fixed number of unlabeled point clouds can be selected to achieve minimum empirical risk $\mathfrak{R}_T[\ell(f, g;\bm{w})]$ on the test set $\mathcal{D}_T$ and minimum annotation time. Below, inspired by \citep{DBLP:conf/colt/MansourMR09, DBLP:journals/ml/Ben-DavidBCKPV10}, we derive the following \textbf{generalization bound} for active 3D detection so that the desired acquisition criteria can be obtained by optimizing the generalization risk.
\begin{theorem}
\label{theo:crb}
Let $\mathcal{H}$ be a hypothesis space of Vapnik-Chervonenkis (VC) dimension $d$, with $f$ and $g$ being the classification and regression branches, respectively. The $\widehat{\mathcal{D}}_{S}$ and $\widehat{\mathcal{D}}_T$ represent the empirical distribution induced by samples drawn from the acquired subset $\mathcal{D}_S$ and the test set $\mathcal{D}_T$, and $\ell$ the loss function bounded by $\mathcal{J}$. It is proven that $\forall$ $\delta \in (0, 1)$, and $\forall f, g\in \mathcal{H}$, with probability at least $1-\delta$ the following inequality holds, 
\vspace{-1ex}
\begin{align*}
        \mathfrak{R}_T[\ell(f, g;\bm{w})]\leq \mathfrak{R}_S[\ell(f, g;\bm{w})] +  \frac{1}{2}disc(\widehat{\mathcal{D}}_{S}, \widehat{\mathcal{D}}_T) +\lambda^* + \text{const},
\end{align*}
where $\text{const} = 3 \mathcal{J} (\sqrt{\frac{\log \frac{4}{\delta}}{2 N_r}} + \sqrt{\frac{\log \frac{4}{\delta}}{2 N_t}}) + \sqrt{\frac{2d \log(e N_r/d)}{N_r}} + \sqrt{\frac{2d \log(e N_t/d)}{N_t}}$.

Notably, $\lambda^* = \mathfrak{R}_T[\ell(f^*, g^*; \bm{w}^*)] +  \mathfrak{R}_S[\ell(f^*, g^*; \bm{w}^*)]$ denotes the joint risk of the optimal hypothesis $f^*$ and $g^*$, with $\bm{w}^*$ being the model weights. $N_r$ and $N_t$ indicate the number of samples in the $\mathcal{D}_S$ and $\mathcal{D}_T$. The proof can be found in the supplementary material.
\end{theorem}
\begin{remark}
The first term indicates the training error on the selected subsets, which is assumed to be trivial based on the zero training assumption \citep{DBLP:conf/iclr/SenerS18}. To obtain a tight upper bound of the generalization risk, the \textbf{optimal subset} $\mathcal{D}_S^*$ can be determined via minimizing the discrepancy distance of empirical distribution of two sets, \textit{i.e.,}
\begin{equation*}
    \mathcal{D}^*_S = \argmin_{\mathcal{D}_S\subset\mathcal{D}_U}disc(\widehat{\mathcal{D}}_{S}, \widehat{\mathcal{D}}_T).
\end{equation*}
Below, we define the discrepancy distance for the 3D object detection task.
\end{remark}

\begin{definition}[] For any $f, g, f', g'\in \mathcal{H}$, the discrepancy between the distribution of the selected sets $\mathcal{D}_S$ and unlabeled pool $\mathcal{D}_T$ can be formulated as,
\begin{align*}
    disc(\widehat{\mathcal{D}}_{S}, \widehat{\mathcal{D}}_T) &= \sup_{f, f'\in\mathcal{H}}|\mathbb{E}_{\widehat{\mathcal{D}}_{S}}\ell(f, f') - \mathbb{E}_{\widehat{\mathcal{D}}_T}\ell(f, f')|  +  \sup_{g, g'\in\mathcal{H}}|\mathbb{E}_{\widehat{\mathcal{D}}_{S}}\ell(g, g') - \mathbb{E}_{\widehat{\mathcal{D}}_T}\ell(g, g')|,
\end{align*}
where the bounded expected loss $\ell$ for any classification and regression functions are symmetric and satisfy the triangle inequality.
\end{definition}
\begin{remark}
As 3D object detection is naturally an integration of classification and regression tasks, mitigating the set discrepancy is basically aligning the inputs and outputs of each branch. Therefore, with the detector freezed during the active selection, finding an optimal $\mathcal{D}^*_S$ can be interpreted as enhancing the acquired set's (1) \textbf{Label Conciseness}: aligning marginal label distribution of bounding boxes, (2) \textbf{Feature Representativeness}: aligning marginal distribution of the latent representations of point clouds, and (3) \textbf{Geometric Balance}: aligning marginal distribution of geometric characteristics of point clouds and predicted bounding boxes, and can be written as:

\begin{equation}\label{eq:optimal}
\mathcal{D}^*_S  \approx \argmin_{\mathcal{D}_S\subset\mathcal{D}_U} \underbrace{
     d_{\mathcal{A}}(P_{\widehat{Y}_S}, P_{Y_T})}_{\text{Conciseness}} +\underbrace{d_{\mathcal{A}}(P_{X_S}, P_{X_T})}_{\text{Representativeness}} + \underbrace{d_{\mathcal{A}}(P_{\phi(\mathcal{P}_S, \strut\widehat{\mathcal{B}}_S)}, P_{\phi(\mathcal{P}_T, \mathcal{B}_T)})}_{\text{Balance}}.    
\end{equation}
Here, \textcolor{black}{$\mathcal{P}_S$ and $\mathcal{P}_T$ represent the point clouds in the selected set and the ones in the test set.} $\phi(\cdot)$ indicates the geometric descriptor of point clouds and $d_{\mathcal{A}}$ distance \citep{DBLP:conf/vldb/KiferBG04} which can be estimated by a finite set of samples. For latent features $X_S$ and $X_T$, we only focus on the features that differ from the training sets, since $\mathbb{E}_{\widehat{D}_L}\ell^{cls}=0$ and $\mathbb{E}_{\widehat{D}_L}\ell^{reg} = 0$ based on the zero training error assumption. Considering that test samples and their associated labels are not observable during training, we make an assumption on the prior distributions of test data. WLOG, we assume that the prior distribution of bounding box labels and geometric features are uniform. Note that we can adopt the KL-divergence for the implementation of $d_{\mathcal{A}}$ assuming that latent representations follow the univariate Gaussian distribution.
\end{remark}
\vspace{-1ex}
\noindent\textbf{Connections with existing AL approaches.} The proposed criteria jointly optimize the discrepancy distance for both tasks with three objectives, which shows the connections with existing AL strategies. The uncertainty-based methods focus strongly on the first term, based on the assumption that learning more difficult samples will help to improve the suprema of the loss. This rigorous assumption can result in a bias towards hard samples, which will be accumulated and amplified across iterations.  Diversity-based methods put more effort into minimizing the second term, aiming to align the distributions in the latent subspace. However, the diversity-based approaches are unable to discover the latent features specified for regression, which can be critical when dealing with a detection problem. We introduce the third term for the 3D detection task, motivated by the fact that aligning the geometric characteristics of point clouds helps to preserve the fine-grained details of objects, leading to more accurate regression. Our empirical study provided in Sec. \ref{sec:ablaton} suggests jointly optimizing three terms can lead to the best performance.

\subsection{Our Approach} 
\vspace{-1ex}
To optimize the three criteria outlined in Eq. \ref{eq:optimal}, we derive an AL scheme consisting of three components. In particular, to reduce the computational overhead, we hierarchically filter the samples that meet the selection criteria (illustrated in Fig. \ref{fig:flowchart}): we first pick $\mathcal{K}_1$ candidates by concise label sampling (\textbf{Stage 1}), from which we select $\mathcal{K}_2$ representative prototypes (\textbf{Stage 2}), with $\mathcal{K}_1, \mathcal{K}_2 << n$. Finally, we leverage greedy search (\textbf{Stage 3}) to find the $N_r$ prototypes that match with the prior marginal distribution of test data. The hierarchical sampling scheme can save $\mathcal{O}((n-\mathcal{K}_1)T_2 + (n-\mathcal{K}_2)T_3)$ cost, with $T_2$ and $T_3$ indicating the runtime of criterion evaluation. The algorithm is summarized in the supplemental material. In the following, we describe the details of the three stages.

\noindent\textbf{Stage 1: Concise Label Sampling (\textsc{Cls}).}\label{sec:cls} 
By using \textit{label conciseness} as a sampling criterion, we aim to alleviate label redundancy and align the source label distribution with the target prior label distribution. Particularly, we find a subset $\mathcal{D}^*_{S_1}$ of size $\mathcal{K}_1$ that minimizes Kullback-Leibler (KL) divergence between the probability distribution $P_{Y_S}$ and the uniform distribution $P_{Y_T}$. To this end, we formulate the KL-divergence with Shannon entropy $H(\cdot)$ and define an optimization problem of maximizing the entropy of the label distributions:
\begin{align}
 &\infdiv{P_{\widehat{Y}_{S_1}}}{P_{Y_T}} = -H(\widehat{Y}_{S_1}) + \log |\widehat{Y}_{S_1}|,\\
&\mathcal{D}^{*}_{S_1} = \argmin_{\mathcal{D}_{S_1}\subset\mathcal{D}_U} \infdiv{P_{\widehat{Y}_{S_1}}}{P_{Y_T}} =\argmax_{\mathcal{D}_{S_1}\subset\mathcal{D}_U} H(\widehat{Y}_{S_1}),
\end{align}
where $\log |\widehat{Y}_{S_1}| = log \mathcal{K}_1$ indicates the number of values $Y_{S_1}$ can take on, which is a constant.
Note that $P_{Y_T}$ is a uniform distribution, and we removed the constant values from the formulations. We pass all point clouds $\{(\mathcal{P})_j\}_{i\in[n]}$ from the unlabeled pool to the detector and extract the predictive labels $\{\hat{y}_i\}_{i=1}^{N_B}$ for $N_B$ bounding boxes, with $\hat{y}_i = \argmax_{y\in[C]} f(x_i; \bm{w}_f)$. The label entropy of the $j$-th point cloud $H(\widehat{Y}_{j, S})$ can be calculated as,
\begin{align}
   \quad H(\widehat{Y}_{j, S}) = -\sum_{c=1}^C\bm{p}_{i,c}\log \bm{p}_{i,c},\quad  \bm{p}_{i,c} = \frac{e^{|{\hat{y}_i=c}|/ N_B}}{\sum^C_{c=1}e^{|{\hat{y}_i=c}|/ N_B}}.
\end{align}
Based on the calculated entropy scores, we filter out the top-$\mathcal{K}_1$ candidates and validate them through the \textbf{Stage 2} representative prototype selection.

\noindent\textbf{Stage 2: Representative Prototype Selection (\textsc{Rps}).}\label{sec:rps} In this stage, we aim to to identify whether the subsets cover the \textit{unique} knowledge encoded only in $\mathcal{D}_U$ and not in $\mathcal{D}_L$ by measuring the \textit{feature representativeness} with gradient vectors of point clouds. Motivated by this, we find the representative prototypes on the gradient space $\mathcal{G}$ to form the subset $\mathcal{D}_{S_2}$, where magnitude and orientation represent the uncertainty and diversity of the new knowledge. For a classification problem, gradients can be retrieved by feeding the hypothetical label $\hat{y} = \argmax_{y\in[C]} \bm{p}(y|x)$ to the networks. However, the gradient extraction for regression problem is not explored yet in the literature, due to the fact that the hypothetical labels for regression heads cannot be directly obtained. To mitigate this, we propose to enable Monte Carlo dropout \textsc{(Mc-dropout)} at the \textbf{Stage 1}, and get the averaging predictions $\bar{B}$ of $M$ stochastic forward passes through the model as the hypothetical labels for regression loss: 
\vspace{-2ex}
\begin{align}
    &\bar{B} \approx \frac{1}{M}\sum_{i=1}^M g(\bm{x};\bm{w}_{d}, \bm{w}_{g}), \bm{w}_{d}\sim \texttt{Bernoulli}(1-p),\\
    &G_{S_2} = \{\nabla_{\Theta} \ell^{reg}(g(\bm{x}), \bar{B};\bm{w}_g), \bm{x}\sim\mathcal{D}_{S_2}\},
\end{align}
with $p$ indicating the dropout rate, $\bm{w}_{d}$ the random variable of the dropout layer, and $\Theta$ the parameters of the convolutional layer of the shared block. The gradient maps $G_{S_2}\in\mathcal{G}$ can be extracted from shared layers and calculated by the chain rule. Since the gradients for test samples are not observable, we make an assumption that its prior distribution follows a Gaussian distribution, which allows us to rewrite the optimization function as,
\begin{equation}\label{eq:rps}
    \begin{split}
    \mathcal{D}_{S_2}^* &= \argmin_{\mathcal{D}_{S_2}\subset\mathcal{D}_{S_1}}\infdiv{P_{X_{S_2}}}{P_{X_T}} \approx \argmin_{\mathcal{D}_{S_2}\subset\mathcal{D}_{S_1}}\infdiv{P_{G_{S_2}}}{P_{G_T}}\\&= \argmin_{\mathcal{D}_{S_2}\subset\mathcal{D}_{S_1}} \log\frac{\sigma_T}{\sigma_{S_2}} +\frac{\sigma_{S_2}^2 + (\mu_{S_2} -\mu_T)}{2\delta^2_{T}} - \frac{1}{2} \approx \mathcal{K}_2\texttt{-medoids}(G_{S_1}),
    \end{split} 
\end{equation}
with $\mu_{S_2}$, $\sigma_{S_2}$ ($\mu_T$, and $\sigma_T$) being the mean and a standard deviation of the univariate Gaussian distribution of the selected set (test set), respectively. Based on Eq. \ref{eq:rps}, the task of finding a representative set can be viewed as picking $\mathcal{K}_2$ prototypes (\textit{i.e.,} ${\mathcal{K}_2}$-medoids) from the clustered data, so that the centroids (mean value) of the selected subset and the test set can be naturally matched. The variance $\sigma_{S_2}$ and $\sigma_{T}$, basically, the distance of each point to its prototypes will be minimized simultaneously. We test different approaches for selecting prototypes in Sec. \ref{sec:ablaton}.

\noindent\textbf{Stage 3: Greedy Point Density Balancing (\textsc{Gpdb}).}\label{sec:gpd} 
The third criterion adopted is \textit{geometric balance}, which targets at aligning the distribution of selected prototypes with the marginal distribution of testing point clouds. As point clouds typically consist of thousands (if not millions) of points, it is computationally expensive to directly align the meta features (\textit{e.g.,} coordinates) of points. Furthermore, in representation learning for point clouds, the common practice of using voxel-based architecture typically relies on quantized representations of point clouds and loses the object details due to the limited perception range of voxels. Therefore, we utilize the point density $\phi(\cdot, \cdot)$ within each bounding box to preserve the geometric characteristics of an object in 3D point clouds. By aligning the geometric characteristic of the selected set and unlabeled pool, the fine-tuned detector is expected to predict more accurate localization and size of bounding boxes and recognize both close (\textit{i.e.,} dense) and distant (\textit{i.e.,} sparse) objects at the test time. The probability density function (PDF) of the point density is not given and has to be estimated from the bounding box predictions. To this end, we adopt Kernel Density Estimation (KDE) using a finite set of samples from each class which can be computed as:
\vspace{-1ex}
\begin{align}
    \bm{p}(\phi(\mathcal{P},\widehat{\mathcal{B}})) = \frac{1}{N_B h}\sum_{j=1}^{N_B} \mathcal{K}er (\frac{\phi(\mathcal{P},\widehat{\mathcal{B}}) - \phi(\mathcal{P},\widehat{\mathcal{B}}_j)}{h}),
\end{align}
with $h>0$ being the pre-defined bandwidth that can determine the smoothing of the resulting density function. We use Gaussian kernel for the kernel function $\mathcal{K}er(\cdot)$. With the PDF defined, the optimization problem of selecting the final candidate sets $\mathcal{D}_{S}$ of size $N_r$ for the label query is:
\vspace{-1ex}
\begin{equation}
    \mathcal{D}_{S}^* = \argmin_{\mathcal{D}_{S}\subset\mathcal{D}_{S_2}}\infdiv{\phi(\mathcal{P}_S,\widehat{\mathcal{B}}_S)}{\phi(\mathcal{P}_T, \mathcal{B}_T)},
\end{equation}
where $\phi(\cdot, \cdot)$ measures the point density for each bounding box. We use greedy search to find the optimal combinations from the subset $\mathcal{D}_{S_2}$ that can minimize the KL distance to the uniform distribution $\bm{p}(\phi(\mathcal{P}_T,\mathcal{B}_T)) \sim \texttt{uniform}(\alpha_{lo}, \alpha_{hi})$. The upper bound $\alpha_{hi}$ and lower bound $\alpha_{lo}$ of the uniform distribution are set to the 95\% density interval, \textit{i.e.,} $ \bm{p}(\alpha_{lo} <\phi(\mathcal{P},\widehat{\mathcal{B}}_j) < \alpha_{hi}) = 95\%$ for every predicted bounding box $j$. Notably, the density of each bounding box is recorded during the \textbf{Stage 1}, which will not cause any computation overhead. The analysis of time complexity against other active learning methods is presented in Sec. \ref{sec:complexity}.
\vspace{-2ex}
\section{Experiments}
\vspace{-2ex}
\subsection{Experimental Setup}\label{sec:setup}
\vspace{-1ex}

\noindent\textbf{Datasets.} KITTI \citep{DBLP:conf/cvpr/GeigerLU12} is one of the most representative datasets for point cloud based object detection. The dataset consists of 3,712 training samples (\textit{i.e.,} point clouds) and 3,769 \textit{val} samples. The dataset includes a total of 80,256 labeled objects with three commonly used classes for autonomous driving: cars, pedestrians, and cyclists. The Waymo Open dataset \citep{DBLP:conf/cvpr/SunKDCPTGZCCVHN20} is a challenging testbed for autonomous driving, containing 158,361 training samples and 40,077 testing samples. The sampling intervals for KITTI and Waymo are set to 1 and 10, respectively.


\noindent \textbf{Generic AL Baselines}. We implemented the following five generic AL baselines of which the implementation details can be found in the supplementary material.
(1) \textbf{\textsc{Rand}}: is a basic sampling method that selects $N_r$ samples at random for each selection round; (2) \textbf{\textsc{Entropy}} \citep{DBLP:conf/ijcnn/WangS14}: is an \textit{uncertainty}-based active learning approach that targets the \textit{classification} head of the detector, and selects the top $N_r$ ranked samples based on the entropy of the sample's predicted label; (3) \textbf{\textsc{Llal}} \citep{DBLP:conf/cvpr/YooK19}: is an \textit{uncertainty}-based method that adopts an auxiliary network to predict an indicative loss and enables to select samples for which the model is likely to produce wrong predictions; (4) \textbf{\textsc{Coreset}} \citep{DBLP:conf/iclr/SenerS18}: is a \textit{diversity}-based method performing the core-set selection that uses the greedy furthest-first search on both labeled and unlabeled embeddings at each round; and (5) \textbf{\textsc{Badge}} \citep{DBLP:conf/iclr/AshZK0A20}: is a \textit{hybrid} approach that samples instances that are disparate and of high magnitude when presented in a hallucinated gradient space.

\noindent \textbf{Applied AL Baselines for 2D and 3D Detection}. For a fair comparison, we also compared three variants of the deep active learning method for 3D detection and adapted one 2D active detection method to our 3D detector. (6) \textbf{\textsc{Mc-mi}} \citep{DBLP:conf/ivs/FengWRMD19} utilized Monte Carlo dropout associated with mutual information to determine the uncertainty of point clouds. (7) \textbf{\textsc{Mc-reg}}: Additionally, to verify the importance of the uncertainty in regression, we design an \textit{uncertainty}-based baseline that determines the \textit{regression} uncertainty via conducting $M$-round \textsc{Mc-dropout} stochastic passes at the test time. The variances of predictive results are then calculated, and the samples with the top-$N_r$ greatest variance will be selected for label acquisition. We further adapted two applied AL methods for 2D detection to a 3D detection setting, where (8) \textbf{\textsc{Lt/c}} \citep{DBLP:conf/accv/KaoLS018} measures the class-specific localization tightness, \textit{i.e.}, the changes from the intermediate proposal to the final bounding box and (9) \textbf{\textsc{Consensus}} \citep{DBLP:conf/ivs/SchmidtRTK20} calculates the variation ratio of minimum IoU value for each RoI-match of 3D boxes.

\begin{table}[h]\vspace{-1ex} 
\centering 
\caption{Performance comparisons (3D AP scores) with generic AL and applied AL for detection on KITTI \textit{val} set with 1\% queried bounding boxes.}
\resizebox{1\linewidth}{!}{%
\begin{tabular}{l l ccc ccc c c c c c c c c}
\toprule 
& &\multicolumn{3}{c}{\textsc{Car}}&\multicolumn{3}{c}{\textsc{Pedestrian}}&\multicolumn{3}{c}{\textsc{Cyclist}} & \multicolumn{3}{c}{\textsc{Average}} \\ 
\cmidrule(l){3-5}\cmidrule(l){6-8} \cmidrule(l){9-11} \cmidrule(l){12-14}  
 & Method &\textsc{Easy} &\textsc{Mod.} &\textsc{Hard} &\textsc{Easy} &\textsc{Mod.} &\textsc{Hard} &\textsc{Easy} &\textsc{Mod.} &\textsc{Hard} &\textsc{Easy} &\textsc{Mod.} &\textsc{Hard}\\ 
\midrule
\parbox[t]{2mm}{\multirow{3}{*}{\rotatebox[origin=c]{90}{Generic}}} 

&\textsc{Coreset}             & 87.77                    & 77.73                   & 72.95                    & 47.27                    & 41.97                   & 38.19                    & 81.73                    & 59.72                   & 55.64                    & 72.26                    & 59.81                   & 55.59                    \\
&\textsc{Badge}                & 89.96                    & 75.78                   & 70.54                    & 51.94                    & 46.24                   & 40.98                    & 84.11                    & 62.29                   & 58.12                    & 75.34                    & 61.44                   & 56.55                    \\
&\textsc{LLAL}                 & 89.95                    & 78.65                   & \textbf{75.32}           & 56.34                    & 49.87                   & 45.97                    & 75.55                    & 60.35                   & 55.36                    & 73.94                    & 62.95                   & 58.88                    \\
\midrule
\parbox[t]{2mm}{\multirow{4}{*}{\rotatebox[origin=c]{90}{AL-Det}}} 

& \textsc{Mc-reg} & 88.85 & 76.21 & 73.47 & 35.82 & 31.81 & 29.79 & 73.98 & 55.23 & 51.85 & 66.21 & 54.41 & 51.70  \\
& \textsc{Mc-mi} &86.28	&75.58	&71.56 &41.05	&37.50	&33.83 &86.26	&60.22	&56.04 &71.19	&57.77	&53.81\\
& \textsc{Consensus}  & 90.14 & 78.01 & 74.28 & 56.43 & 49.50 & 44.80 & 78.46 & 55.77 & 53.73 & 75.01 & 61.09 & 57.60 \\
& \textsc{Lt/c} & 88.73 & 78.12 & 73.87 & 55.17 & 48.37 & 43.63 & 83.72 & 63.21 & 59.16 & 75.88 & 63.23 & 58.89                     \\

\midrule
\midrule
&\textsc{Crb}                  & \textbf{90.98}         & \textbf{79.02}          & 74.04                    & \textbf{64.17}           & \textbf{54.80}          & \textbf{50.82}           & \textbf{86.96}           & \textbf{67.45}          & \textbf{63.56}           & \textbf{80.70}           & \textbf{67.81}          & \textbf{62.81}           \\
\bottomrule \vspace{-0.7cm}
\end{tabular}
}
\label{tab:generic_applied} 
\end{table}   

\begin{figure}[t]%
\vspace{-3ex}
    \subfloat{{\includegraphics[width=0.33\textwidth]{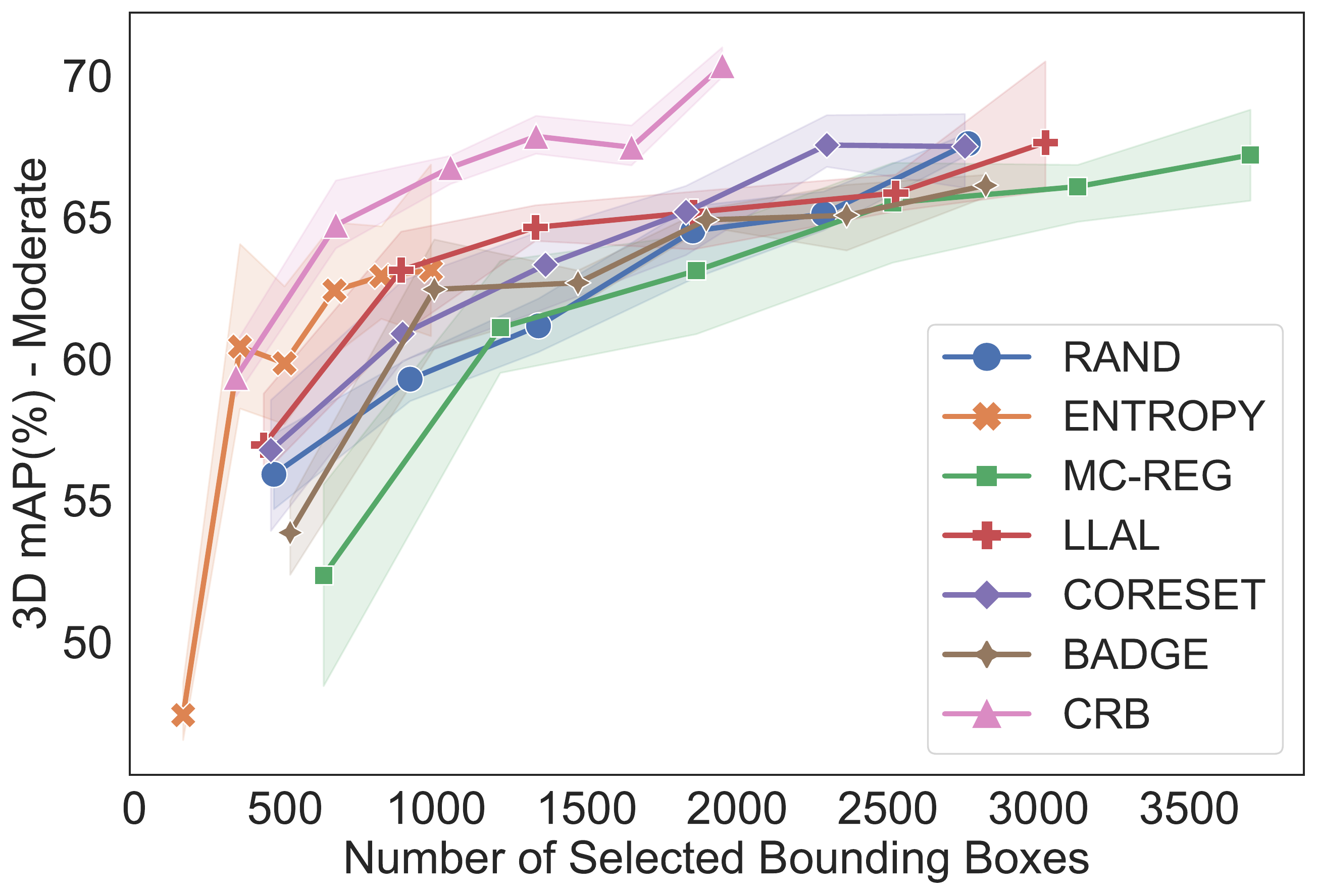} }\vspace{-3ex}}%
    \subfloat{{\includegraphics[width=0.33\textwidth]{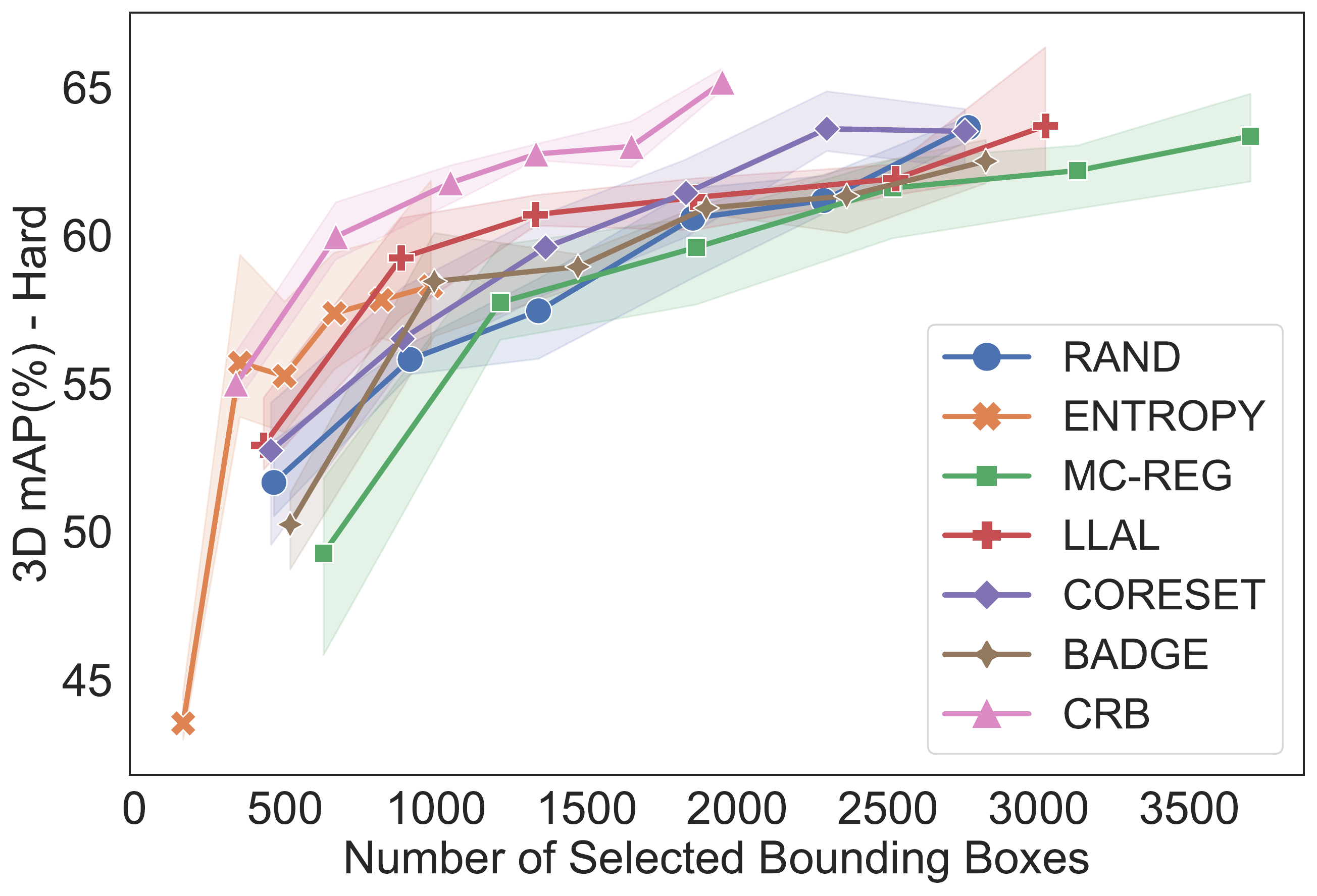} }\vspace{-3ex}}%
    \subfloat{{\includegraphics[width=0.33\textwidth]{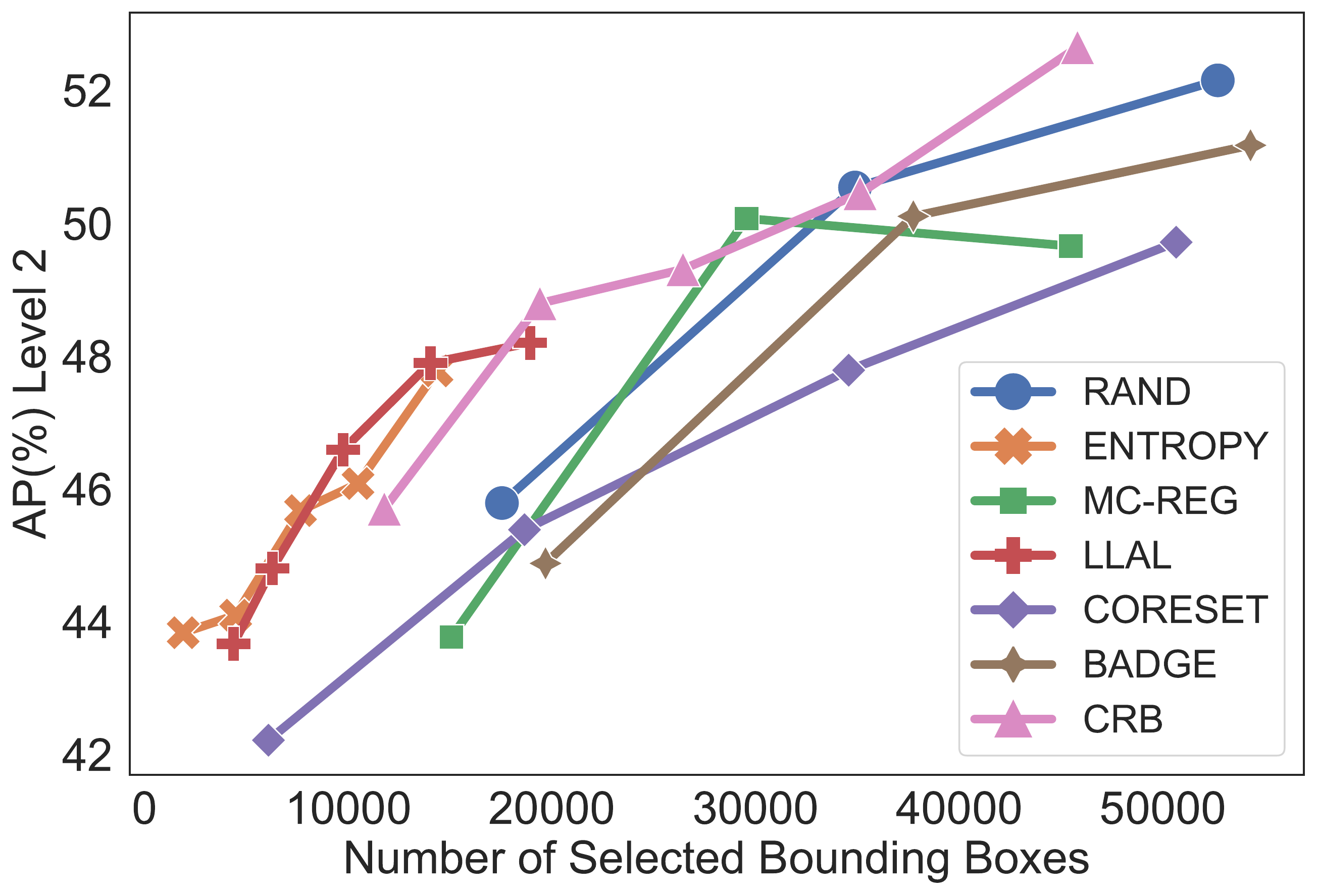} }\vspace{-3ex}}%
   \hfill
    \subfloat{{\includegraphics[width=0.33\textwidth]{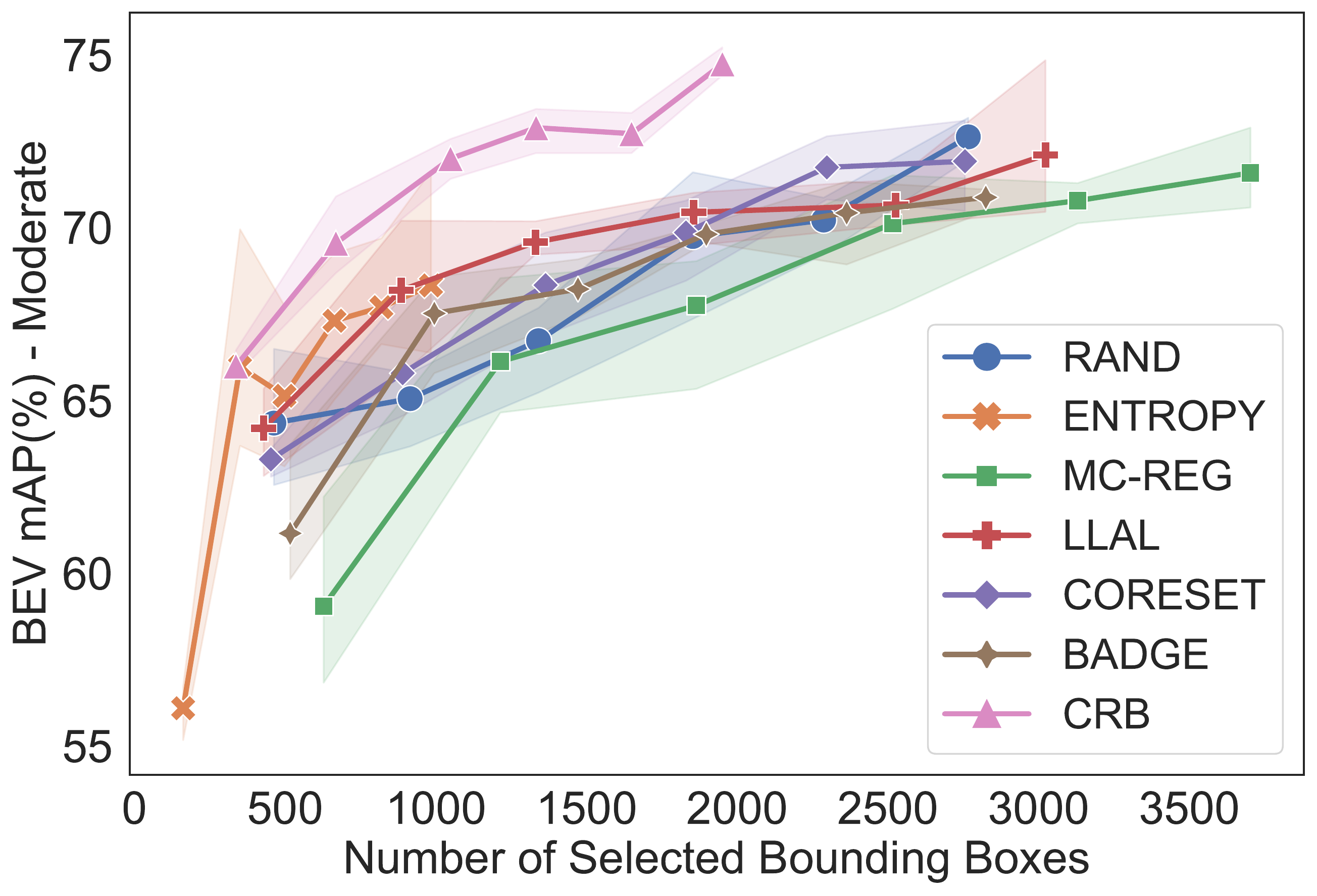} }}%
    \subfloat{{\includegraphics[width=0.33\textwidth]{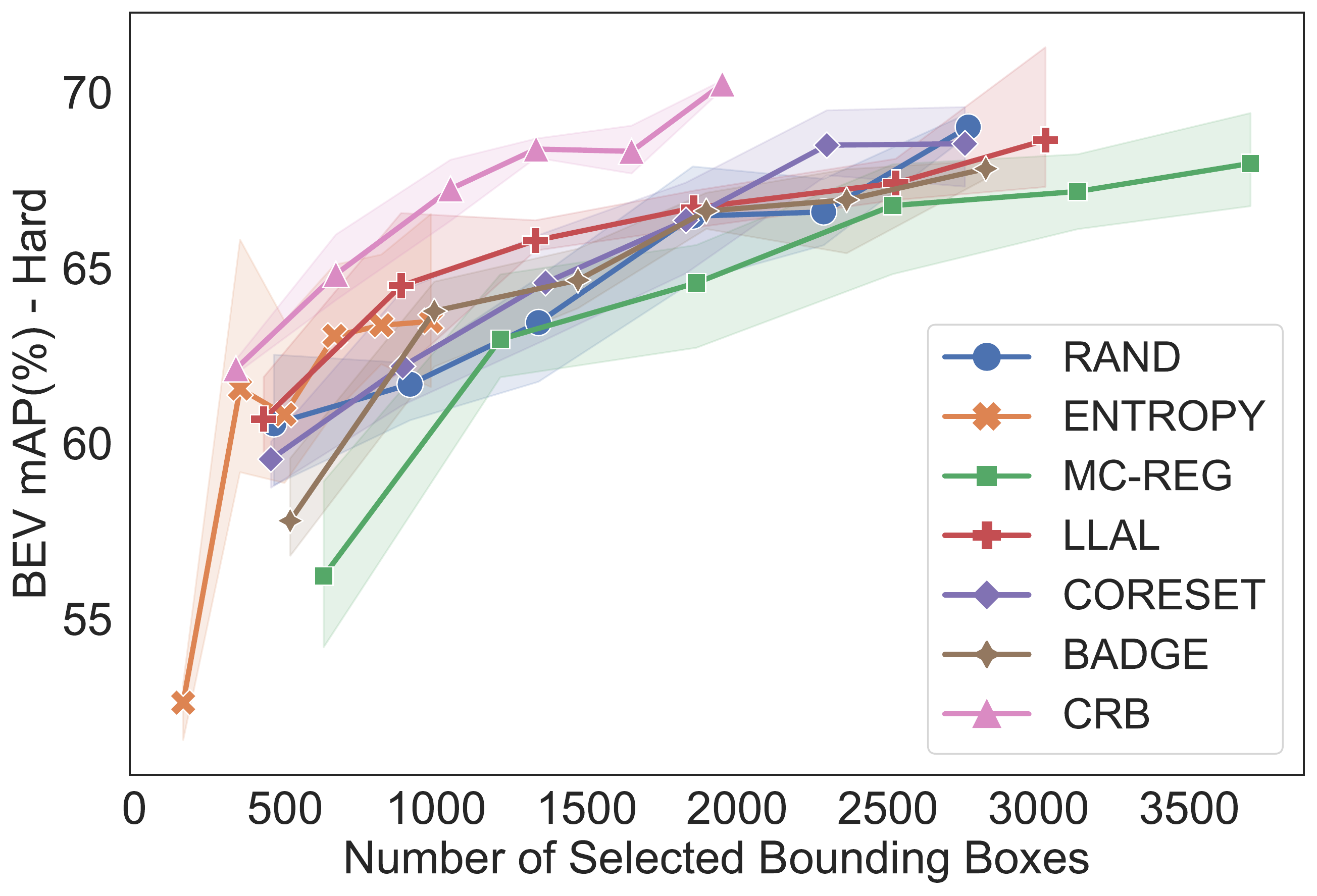} }}%
    \subfloat{{\includegraphics[width=0.33\textwidth]{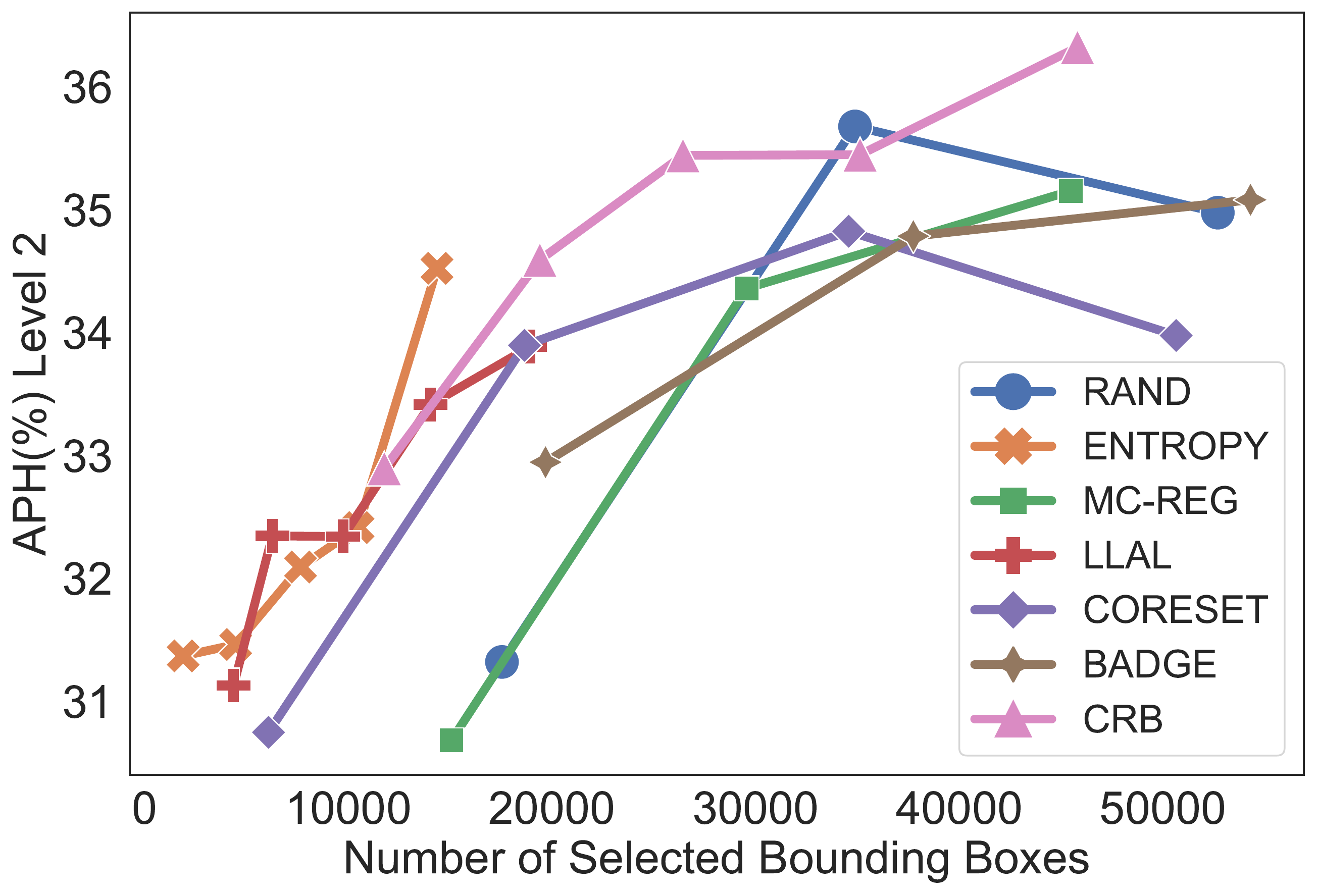} }\vspace{-5ex}}%
    \caption{3D and BEV mAP (\%) of \textsc{Crb} and AL baselines on the KITTI and Waymo \textit{val} split.}%
    \vspace{-3ex}
    \label{fig:kitti_results_boxes}%
\end{figure}

\subsection{Comparisons against Active Learning Methods} 
\vspace{-1ex}
\textbf{Quantitative Analysis}. We conducted comprehensive experiments on the KITTI and Waymo datasets to demonstrate the effectiveness of the proposed approach. The $\mathcal{K}_{1}$ and $\mathcal{K}_{2}$ are empirically set to \textcolor{black}{$300$ and $200$} for KITTI and $2,000$ and $1,200$ for Waymo. Under a fixed budget of point clouds, the performance of 3D and BEV detection achieved by different AL policies are reported in Figure \ref{fig:kitti_results_boxes}, with standard deviation of three trials shown in shaded regions. We can clearly observe that \textsc{Crb} consistently outperforms all state-of-the-art AL methods by a noticeable margin, irrespective of the number of annotated bounding boxes and difficulty settings. It is worth noting that, on the KITTI dataset, the annotation time for the proposed \textsc{Crb} is 3 times faster than \textsc{Rand}, while achieving a comparable performance. Moreover, AL baselines for regression and classification tasks (\textit{e.g.}, \textsc{Llal}) or for regression only tasks (\textit{e.g.}, \textsc{Mc-reg}) generally obtain higher scores yet leading to higher labeling costs than the classification-oriented methods (\textit{e.g.}, \textsc{Entropy}).

Table \ref{tab:generic_applied} reports the major experimental results of the state-of-the-art generic AL methods and applied AL approaches for 2D and 3D detection on the KITTI dataset. It is observed that \textsc{Llal} and \textsc{Lt/c} achieve competitive results, as the acquisition criteria adopted jointly consider the classification and regression task. Our proposed \textsc{Crb} improves the 3D mAP scores by 6.7\% which validates the effectiveness of minimizing the generalization risk.

\noindent\textbf{Qualitative Analysis}. To intuitively understand the merits of our proposed active 3D detection strategy, Figure \ref{fig:3d_vis} demonstrates that the 3D detection results yielded by \textbf{\textsc{Rand}} (bottom left) and \textbf{\textsc{Crb}} selection (bottom right) from the corresponding image (upper row). Both 3D detectors are trained under the budget of 1K annotated bounding boxes. False positives and corrected predictions are indicated with red and green boxes. It is observed that, under the same condition, \textsc{Crb} produces more accurate and more confident predictions than \textsc{Rand}. Besides, looking at the cyclist highlighted in the orange box in Figure \ref{fig:3d_vis}, the detector trained with \textsc{Rand} produces a significantly lower confidence score compared to our approach. This confirms that the samples selected by \textsc{Crb} are aligned better with the test cases. More visualizations can be found in the supplemental material.

\begin{table*}[t]\vspace{-3ex}
  \caption{Ablative study of different active learning criteria on the KITTI \textit{val} split. 3D and BEV AP scores (\%) are reported when 1,000 bounding boxes are annotated.\vspace{-1.5ex}}
  \label{tab:abla}
  \resizebox{1\linewidth}{!}{%
  \centering
  \begin{tabular}{ccc ccc ccc}
    \toprule
     & &  & \multicolumn{3}{c}{3D Detection mAP} &\multicolumn{3}{c}{BEV Detection mAP}\\
     \cmidrule(l){4-6}\cmidrule(l){7-9}
     \textsc{Cls} & \textsc{Rps} & \textsc{Gpdb} &\textsc{Easy} &\textsc{Moderate} &\textsc{Hard} &\textsc{Easy} &\textsc{Moderate} &\textsc{Hard}\\
    \midrule
     -                           &-         &-
    &$70.70_{\pm1.60}$ &$58.27_{\pm1.04}$ &$54.69_{\pm 1.30}$ &$75.37_{\pm 1.65}$ &$64.54_{\pm1.69}$ &$61.36_{\pm 1.61}$\\
    $\surd$                       &-         &-          &$77.76_{\pm1.70}$ &$64.56_{\pm1.39}$ &$59.54_{\pm 1.13}$ &$81.07_{\pm 1.67}$ &$69.76_{\pm1.45}$ &$65.01_{\pm 1.31}$\\
    -                             &$\surd$   &-         
    &$74.93_{\pm 3.11}$ &$61.65_{\pm1.95}$ &$57.70_{\pm 1.52}$ &$78.85_{\pm 2.31}$ &$67.07_{\pm1.36}$ &$63.47_{\pm 1.21}$\\
    -                             &-         &$\surd$   
    &$69.11_{\pm 13.22}$ &$56.12_{\pm12.74}$ &$52.85_{\pm 11.49}$ &$73.57_{\pm 10.45}$ &$62.49_{\pm10.62}$ &$59.45_{\pm 9.78}$\\
    $\surd$                       &$\surd$   &-         
    &$76.19_{\pm 2.13}$ &$62.81_{\pm1.31}$ &$58.03_{\pm 1.18}$ &$80.73_{\pm 0.92}$ &$68.67_{\pm0.21}$ &$64.42_{\pm 0.22}$\\
    $\surd$                       &-         &$\surd$   
    &$76.72_{\pm 0.78}$ &$64.70_{\pm1.07}$ &$59.68_{\pm 0.93}$ &$80.71_{\pm 0.26}$ &$70.01_{\pm0.40}$ &$65.47_{\pm 0.56}$\\
    $\surd$                       &$\surd$   &$\surd$   &$\textbf{79.03}_{\pm 1.39}$ &$\textbf{65.86}_{\pm1.21}$ &$\textbf{61.06}_{\pm 1.43}$ &$\textbf{82.60}_{\pm 1.34}$ &$\textbf{70.74}_{\pm0.57}$ &$\textbf{66.41}_{\pm 1.22}$\\
  \bottomrule
\end{tabular}\vspace{-2ex}
}
\end{table*}
\begin{figure}[t]
    \centering\vspace{-2ex}
    \includegraphics[width=1\linewidth]{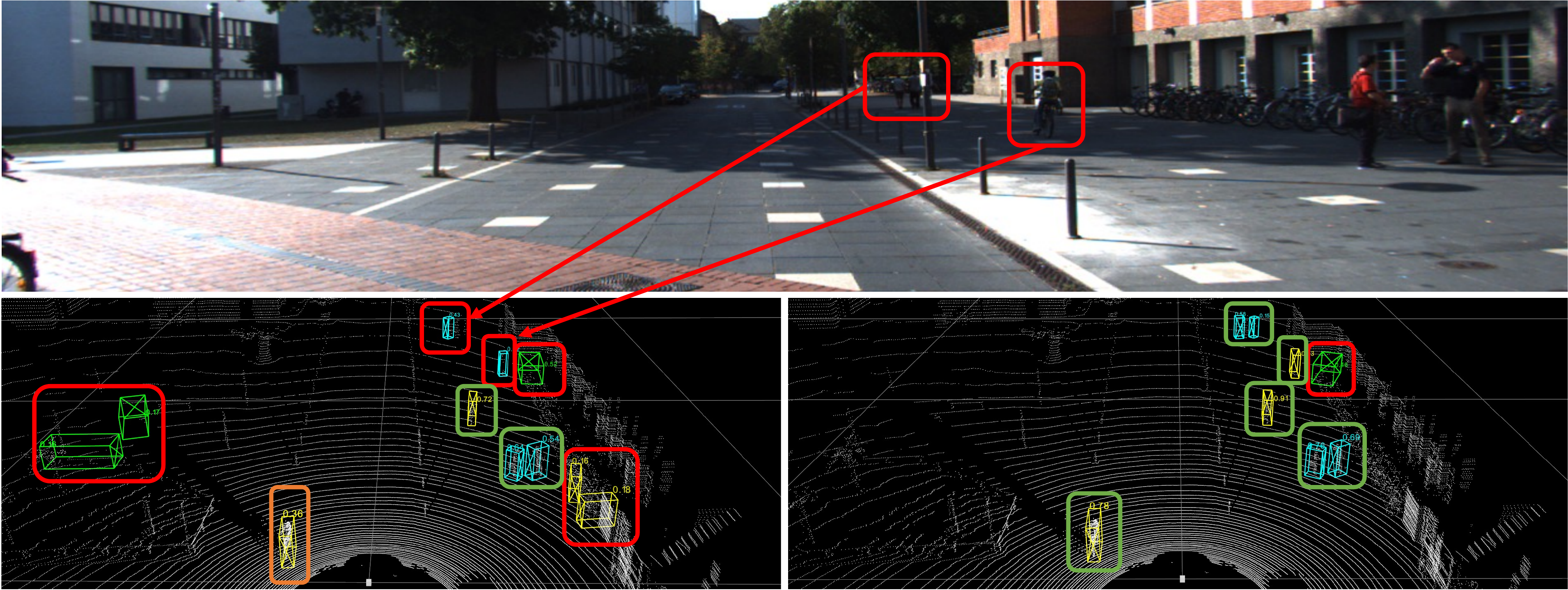}\vspace{-1.5ex}
    \caption{A case study of active 3D detection performance of \textbf{\textsc{Rand}} (bottom left) and \textbf{\textsc{Crb}} (bottom right) under the budge of 1,000 annotated bounding boxes. False positive (corrected predictions) are highlighted in red (green) boxes. The orange box denotes the detection with low confidence.}\vspace{-2ex}
    \label{fig:3d_vis}
\end{figure}
 
\vspace{-2ex}
\subsection{Ablation Study}\label{sec:ablaton}
\vspace{-2ex}
\textbf{Study of Active Selection Criteria}. Table \ref{tab:abla} reports the performance comparisons of six variants of the proposed \textsc{Crb} method and the basic random selection baseline (1-st row) on the KITTI dataset. We report the 3D and BEV mAP metrics at all difficulty levels with 1,000 bounding boxes annotated. We observe that only applying \textsc{Gpdb} (4-th row) produces 12.5\% lower scores and greater variance than the full model (the last row). However, with \textsc{Cls} (6-th row), the performance increases by approximately 10\% with the minimum variance. This phenomenon evidences the importance of optimizing the discrepancy for both classification and regression tasks. It's further shown that removing any selection criteria from the proposed \textsc{Crb} triggers a drop on mAP scores, confirming the importance of each in a sample-efficient AL strategy.

\begin{figure}[h]
\vspace{-3ex}
  \begin{minipage}[c]{0.71\textwidth}
    \subfloat{{\includegraphics[width=0.495\textwidth]{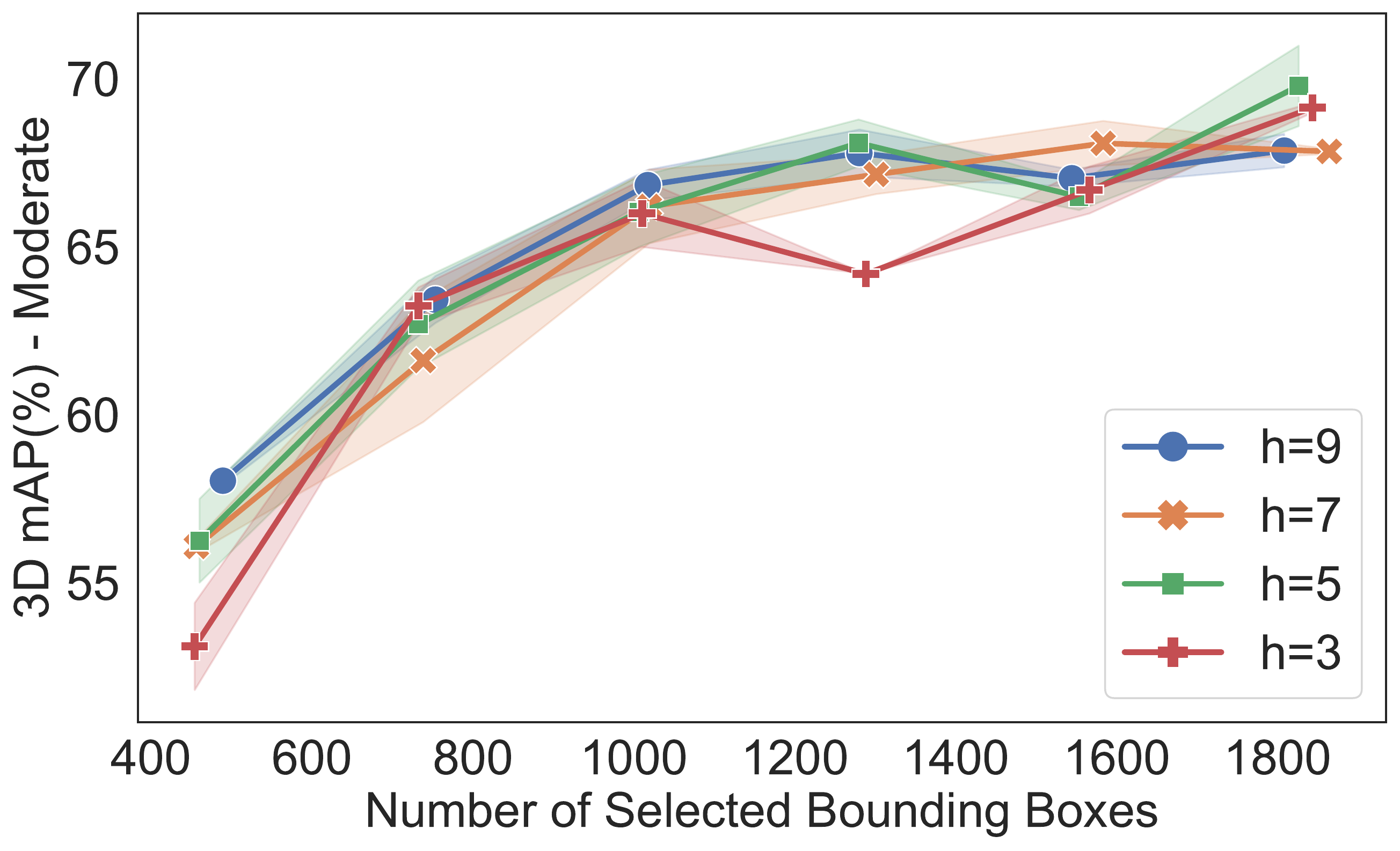} }}%
    \subfloat{{\includegraphics[width=0.495\textwidth]{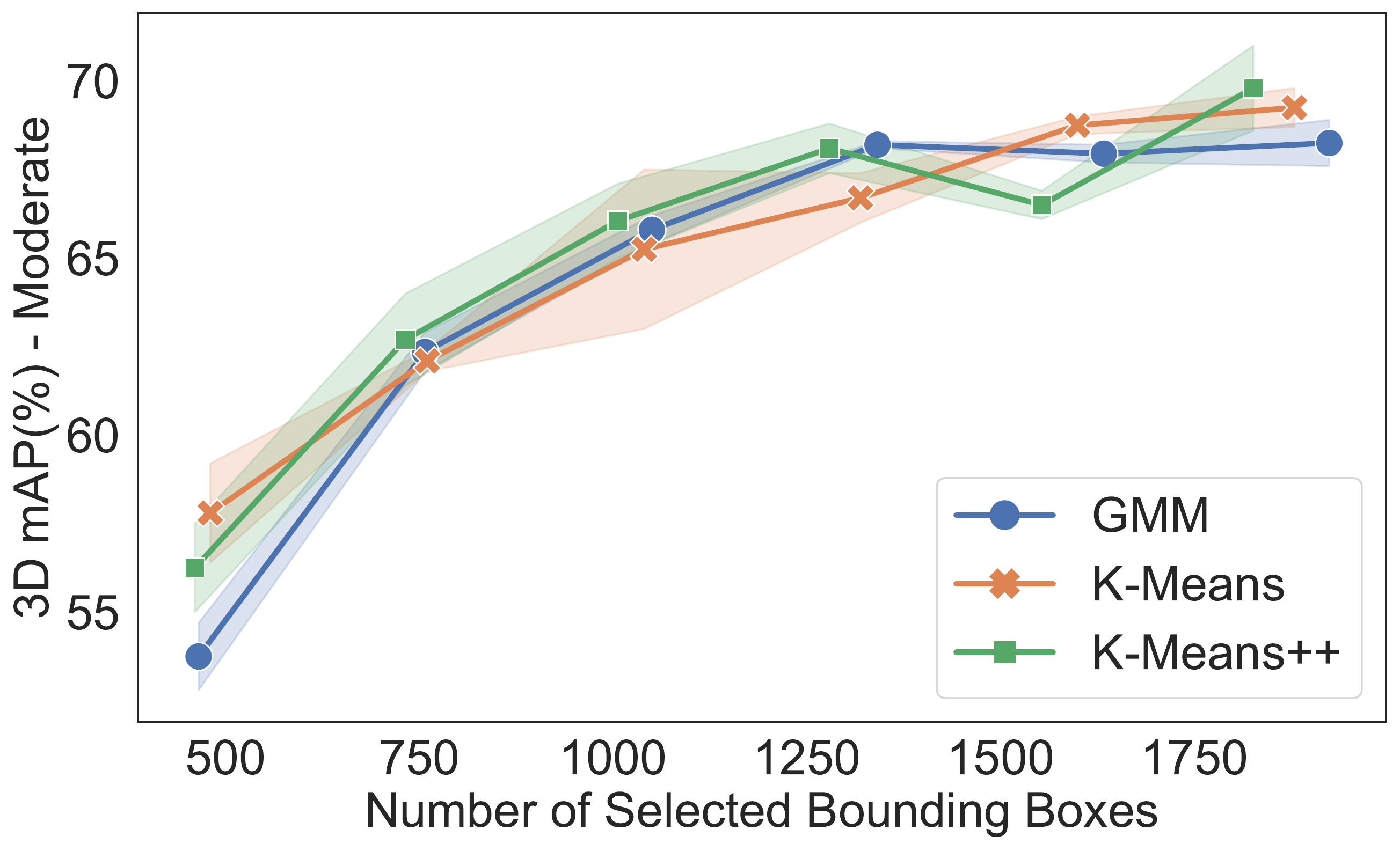} }}%
    \vspace{-2ex}
  \end{minipage}\hfill
  \begin{minipage}[c]{0.28\textwidth}
    \caption{
      Results on KITTI \textit{val} set with varying KDE bandwidth $h$ (left) and prototype selection approaches (right) with increasing queried bounding boxes.
    } \label{fig:03-03}
  \end{minipage}
\end{figure}

\noindent\textbf{Sensitivity to Prototype Selection.} We examine the sensitivity of performance to different prototype selection methods used in the \textsc{Rps} module on the KITTI dataset (moderate difficulty level). Particularly, In Figure \ref{fig:03-03} (right), we show the performance of our approach using different prototype selection methods of the Gaussian mixture model (\textsc{Gmm}), \textsc{K-means}, and \textsc{K-means}++. To fairly reflect the trend in the performance curves, we run two trials for each prototype selection approach and plot the mean and the variance bars. \textsc{K-means} is slightly more stable than the other two, with higher time complexity and better representation learning. It is observed that there is very little variation ($\sim 1.5\%$) in the performance of our approach when using different prototype selection methods. This confirms that the \textsc{Crb}'s superiority over existing baselines is not coming from the prototype selection method.

\noindent\textbf{Sensitivity to Bandwidth $h$.} Figure \ref{fig:03-03} depicts the results of \textsc{Crb} with the bandwidth $h$ varying in $\{3, 5, 7, 9\}$. Choosing the optimal bandwidth value $h^*$ can avoid under-smoothing ($h < h^*$) and over-smoothing ($h > h^*$) in KDE. Except $h=3$ which yields a large variation, \textsc{Crb} with the bandwidth of all other values reach similar detection results within the 2\% absolute difference on 3D mAP. This evidences that the \textsc{Crb} is robust to different values of bandwidth. 

\textbf{Sensitivity to Detector Architecture}. We validate the sensitivity of performance to choices of one-stage and two-stage detectors. Table \ref{tab:second} reports the results with the \textsc{Second} detection backbone on the KITTI dataset. With only 3\% queried 3D bounding boxes, it is observed that the proposed \textsc{Crb} approach consistently outperforms the SOTA generic active learning approaches across a range of detection difficulties, improving 4.7\% and 2.8\% on 3D mAP and BEV mAP scores.

\begin{table}[t] \vspace{-6ex}
\centering 
\caption{Performance comparisons on KITTI \textit{val} set \textit{w.r.t.} varying thresholds $\mathcal{K}_1$ and $\mathcal{K}_2$ after two rounds of active selection (8\% point clouds, ~1\% bounding boxes). Results are reported with 3D AP with 40 recall positions. $^\dag$ indicates the reported performance of the backbone trained with the full labeled set (100\%).}
\resizebox{1\linewidth}{!}{%
\begin{tabular}{l l ccc ccc c c c c c c c c}
\toprule 
& &\multicolumn{3}{c}{\textsc{Car}}&\multicolumn{3}{c}{\textsc{Pedestrian}}&\multicolumn{3}{c}{\textsc{Cyclist}} & \multicolumn{3}{c}{\textsc{Average}} \\ 
\cmidrule(l){3-5}\cmidrule(l){6-8} \cmidrule(l){9-11} \cmidrule(l){12-14}  
$\mathcal{K}_1$ & $\mathcal{K}_2$ &\textsc{Easy} &\textsc{Mod.} &\textsc{Hard} &\textsc{Easy} &\textsc{Mod.} &\textsc{Hard} &\textsc{Easy} &\textsc{Mod.} &\textsc{Hard} &\textsc{Easy} &\textsc{Mod.} &\textsc{Hard}\\ 
\midrule
$500$  &$400$ &$90.04$ &$79.08$ &$\textbf{74.66}$ &$57.11$ &$51.10$ &$\textbf{51.12}$ &$81.97$ &$63.40$ &$59.62$  &$76.50$ &$64.53$ &$60.10$\\
$500$  &$300$ &$90.98$ &$79.02$ &$74.04$ &$64.17$ &$54.80$ &$50.82$ &$\textbf{86.96}$ &$\textbf{67.45}$ &$\textbf{63.56}$  &$\textbf{80.70}$ &$\textbf{67.81}$ &$\textbf{62.81}$\\
$400$  &$300$ &$\textbf{91.30}$ &$\textbf{79.21}$ &$74.00$ &$62.93$ &$55.67$ &$49.27$ &$79.02$ &$60.50$ &$56.74$ &$77.75$ &$65.13$ &$60.00$\\
$300$  &$200$ &$90.45$ &$78.81$ &$73.44$ &$\textbf{65.00}$ &$\textbf{55.91}$ &$\textbf{51.12}$ &$84.82$ &$65.77$ &$61.53$ &$80.09$ &$67.32$ &$62.05$\\
\midrule
\multicolumn{2}{c}{\textsc{Pv-rcnn}$^\dag$} &92.56 &84.36 &82.48 &64.26 &56.67 &51.91 &88.88 &71.95 &66.78 &81.75 &70.99 &67.06\\
\bottomrule \vspace{-0.7cm}
\end{tabular}
}
\label{tab:param_k1k2} 
\end{table}   

\noindent\textbf{Sensitivity Analysis of Thresholds $\mathcal{K}_1$ and $\mathcal{K}_2$.} 
We examine the sensitivity of our approach to different values of threshold parameters $\mathcal{K}_1$ and $\mathcal{K}_2$. We report the mean average precision (mAP) on the KITTI dataset, including both 3D and BEV views at all difficulty levels. We check four possible combinations of $\mathcal{K}_1$ and $\mathcal{K}_2$ and show the results in Table \ref{tab:param_k1k2}. We can observe that at \textsc{Moderate} and \textsc{Hard} levels, there is only 3.28\% and 2.81\% fluctuation on average mAP. In the last row, we further report the accuracy achieved by the backbone detector trained with all labeled training data and a larger batch size. With only 8\% point clouds and 1\% annotated bounding boxes, \textsc{Crb} achieves a comparable performance to the full model.

\vspace{-2ex}
\subsection{Complexity Analysis}\label{sec:complexity}
\vspace{-1ex}
Table \ref{tab:complexity} shows the time complexity of training and active selection for different active learning approaches. $n$ indicates the total number of unlabeled point clouds, $N_r$ is the quantity selected, and $E$ is the training epochs, with $N_r\ll n$. We can clearly observe that, at training stage, the complexity of all AL strategies is $\mathcal{O}(E n)$, except \textsc{Llal} that needs extra epochs $E_{l}$ to train the loss prediction module. At the active selection stage, \textsc{Rand} randomly generates $N_r$ indices to retrieve samples from the pool. 
 \textsc{Coreset} computes pairwise distances between the embedding of selected samples and unlabeled samples that yields the time complexity of $\mathcal{O}(N_rn)$. \textsc{Badge} iterates through the gradients of all unlabeled samples passing gradients into \textsc{K-means}++ algorithm, with the complexity of $\mathcal{O}(N_rn)$ bounded by \textsc{K-means}++. Given $\mathcal{K}_1, \mathcal{K}_2 \approx N_r$, the time complexity of our method is $\mathcal{O}(n\log n + 2N_r^2)$, with $\mathcal{O}(n\log(n))$ being the complexity of sorting the entropy scores in \textsc{Cls}, and $\mathcal{O}(N_r^2)$ coming from $\mathcal{K}_2$-medoids and greedy search in \textsc{Rps} and \textsc{Gpdb}. Note that, in our case, $\mathcal{O}(n\log n + 2N_r^2) < \mathcal{O}(N_r n)$. The complexity of simple ranking-based baselines is $\mathcal{O}(n\log(n))$ due to sorting the sample acquisition scores. Comparing our method with recent state-of-the-arts, \textsc{Llal} has the highest training complexity, and \textsc{Badge} and \textsc{Coreset} have the highest selection complexity. Unlike the existing baseline, training and selection complexities of the proposed \textsc{Crb} are upper bounded by the reasonable asymptotic growth rates.\begin{table}[h]
 \vspace{-2ex}
  \resizebox{\linewidth}{!}{
    \begin{minipage}{.6\textwidth}
        \caption{AL Results with one-stage 3D detector \textsc{Second}.}
        \centering\label{tab:second}
        \vspace{-0.05cm}
            \resizebox{\linewidth}{!}{
            \begin{tabular}{l ccc ccc}
                \toprule
                 &\multicolumn{3}{c}{3D Detection mAP} &\multicolumn{3}{c}{BEV Detection mAP}\\
                 \cmidrule(l){2-4}\cmidrule(l){5-7}
                  &\textsc{Easy} &\textsc{Moderate} &\textsc{Hard} &\textsc{Easy} &\textsc{Moderate} &\textsc{Hard}\\
                \midrule
                \textsc{Rand}	&75.23	&60.83	&56.55	&80.20	&67.56	&63.30\\
                \textsc{Llal}	&72.02	&58.96	&54.21	&79.50	&66.82	&62.48\\
                \textsc{Coreset} &74.74	&58.86	&54.61	&79.71	&65.53	&61.39\\
                \textsc{Badge}	&75.38	&61.65	&56.72	&80.81	&68.83	&64.17\\
                \textsc{Crb}	&\textbf{78.96}	&\textbf{64.27}	&\textbf{59.60}	&\textbf{83.28}	&\textbf{70.49}	&\textbf{66.09}\\
              \bottomrule
            \end{tabular}}
    \end{minipage}%

    \begin{minipage}{.39\textwidth}
        \centering
        \caption{Complexity Analysis.}\label{tab:complexity}
        \resizebox{1\linewidth}{!}{
            \begin{tabular}{l c c}
                \toprule
                AL Strategy & Training &  Selection \\
                \midrule
                 \textsc{Rand} & $\mathcal{O}(E n)$ & $\mathcal{O}(N_r)$
                 \\
                 \textsc{Entropy} & $\mathcal{O}(E n)$ & $\mathcal{O}(n\log n)$
                 \\
                 \textsc{Mc-reg} & $\mathcal{O}(E n)$ & $\mathcal{O}(n\log n)$
                 \\
                 \textsc{Llal} & $\mathcal{O}((E + E_{l}) n)$ & $\mathcal{O}(n\log n)$
                 \\
                 \textsc{Coreset} & $\mathcal{O}(E n)$ & $\mathcal{O}(N_rn)$
                 \\
                 \textsc{Badge} & $\mathcal{O}(E n)$ & $\mathcal{O}(N_rn)$
                 \\
                 \textsc{Crb} & $\mathcal{O}(E n)$ & $\mathcal{O}(n\log n + 2N_r^2)$
                 \\
              \bottomrule
              \vspace{-2ex}
                \end{tabular}
} 
    \end{minipage}
}  
\end{table}
%

\vspace{-6ex}
\section{Discussion}
\vspace{-2ex}
This paper studies three novel criteria for sample-efficient active 3D object detection, that can effectively achieve high performance with minimum costs of 3D box annotations and runtime complexity. We theoretically analyze the relationship between finding the optimal acquired subset and mitigating the sets discrepancy. The framework is versatile and can accommodate existing AL strategies to provide in-depth insights into heuristic design. The limitation of this work lies in a set of assumptions made on the prior distribution of the test data, which could be violated in practice. For more discussions, please refer to Sec. A.1 in Appendix. In contrast, it opens an opportunity of adopting our framework for active domain adaptation, where the target distribution is accessible for alignment. Addressing these two avenues is left for future work.

\section*{Reproducibility Statement}
The source code of the developed active 3D detection toolbox is available in the supplementary material, which accommodates various AL approaches and one-stage and two-stage 3D detectors. We specify the settings of hyper-parameters, the training scheme and the implementation details of our model and AL baselines in Sec. B of the supplementary material. We show the proofs of Theorem \ref{theo:crb} in Sec. C followed by the overview of the algorithm in Sec. D in the supplementary material. We repeat the experiments on the KITTI dataset 3 times with different initial labeled sets and show the standard deviation in plots and tables.

\section*{Ethics Statement}
Our work may have a positive impact on communities to reduce the costs of annotation, computation, and carbon footprint. The high-performing AL strategy greatly enhances the feasibility and practicability of 3D detection in critical yet data-scarce fields such as medical imaging. We did not use crowdsourcing and did not conduct research with human subjects in our experiments. We cited the creators when using existing assets (\textit{e.g.}, code, data, models).

\appendix
In this appendix, we discuss the prior distribution selection, \textcolor{black}{motivation of the Stage 2,} and evaluation division of difficulty in Sec \ref{sec:prior} and Sec \ref{sec:difficulty}, respectively. In the rest of the supplementary material, we provide the implementation details of all baselines and the proposed approach in Sec \ref{sec:imple} followed by the proof of Theorem \ref{theo:crb}. In Sec \ref{sec:alg}, the overall algorithm is summarized. Additional experimental results on KITTI (Sec \ref{sec:kitti}) and Waymo (Sec \ref{sec:waymo}) datasets are reported and analyzed. We further conducted supplemental experiments on parameter sensitivity (Sec \ref{sec:param}) and visualizations (Sec \ref{sec:vis}). In the end, we leave the related work and the associated discussion in Sec \ref{sec:rw}.
\section{Appendix}

\subsection{More Discussions on Prior Distribution}\label{sec:prior}
In mainstream 3D detection datasets, the curated test set is commonly long-tailed distributed, with a few head classes (\textit{e.g.}, car) possessing a large number of samples and all the rest of the tail classes possessing only a few samples. As such, the trained detector can be easily biased towards head classes with massive training data, resulting in high accuracy on head classes and low accuracy on tail classes. This suggests that for 3D detection tasks, \textbf{mean average precision (mAP)} can be a \textbf{fairer} metric of evaluation, by taking an average of all AP values per class. When the test label is uniformly distributed, mAP scores will be equal to the AP scores for all samples. This motivates us to choose the uniform distribution as the prior distribution, rather than estimating the test label distribution from the initial labeled set $\mathcal{D}_L$. In this case, the trained model tends to be more robust and resilient to the imbalanced training data, achieving higher mAP scores.\\

To justify the effectiveness of choosing the uniform distribution, we provide more comparisons with the SOTA active learning methods in Table \ref{tab:uniform} and Table \ref{tab:uniform-gap}, which do not take the uniform distribution as an assumption. We clearly observe that such AL methods perform poorly on \textbf{tail classes} (\textit{e.g.}, pedestrian and cyclist), confirming that the yielded models are biased towards learning car samples.

\begin{table*}[h]
  \caption{Performance gap (\%) between different AL methods and fully supervised backbone when acquiring approximately 1\% queried bounding boxes on KITTI. Gaps are calculated by subtracting the performance of a fully supervised backbone from the performance of AL methods. \vspace{-1.5ex}}
  \label{tab:uniform-gap}
  \resizebox{1\linewidth}{!}{%
  \centering
  \begin{tabular}{lcc ccc ccc ccc ccc}
    \toprule
       & \multicolumn{3}{c}{Car ($\downarrow$)} &\multicolumn{3}{c}{Pedestrian ($\downarrow$)}
       &\multicolumn{3}{c}{Cyclist ($\downarrow$)}
        &\multicolumn{3}{c}{Average ($\downarrow$)} \\
     \cmidrule(l){2-4}\cmidrule(l){5-7}\cmidrule(l){8-10}\cmidrule(l){11-13} 
     Method &\textsc{Easy} &\textsc{Mod.} &\textsc{Hard} &\textsc{Easy} &\textsc{Mod.} &\textsc{Hard} &\textsc{Easy} &\textsc{Mod.} &\textsc{Hard} & \textsc{Easy} &\textsc{Mod.} &\textsc{Hard}\\
    \midrule
    \textsc{Llal}
    &$2.61$ &$5.71$ &$\textbf{7.16}$ &$7.92$ &$6.80$ &$5.94$ &$13.33$ &$11.60$ &$11.42$ &$7.81$ &$8.04$ &$8.18$\\
    \textsc{Coreset}         
    &$4.79$ &$6.63$ &$9.53$ &$16.99$ &$14.70$ &$13.72$ &$7.15$ &$12.23$ &$11.14$ &$9.49$ &$11.18$ &$11.47$\\
    \textsc{Badge}         
    &$2.60$ &$8.58$ &$11.94$ &$12.32$ &$10.43$ &$10.93$ &$4.77$ &$9.66$ &$8.66$ &$6.41$ &$9.55$ &$10.51$\\
    \textsc{CRB}   
    &$\textbf{1.58}$ &$\textbf{5.34}$ &$8.44$ &$\textbf{0.09}$ &$\textbf{1.87}$ &$\textbf{1.09}$ &$\textbf{1.92}$ &$\textbf{4.50}$ &$\textbf{3.22}$ &$\textbf{1.05}$ &$\textbf{3.18}$ &$\textbf{4.25}$ \\
  \bottomrule \vspace{-4ex}
\end{tabular}
}
\end{table*}

\subsection{More Discussions on Evaluation Division of Difficulty}\label{sec:difficulty}

\textcolor{black}{On the KITTI dataset, the evaluation difficulty is set based on the visual look\footnote{\url{http://www.cvlibs.net/datasets/kitti/eval_object.php}} of the images, which is supposed to be unavailable for our LiDAR-based detection task. On the other hand, the Waymo dataset leverages a more reasonable and general setting of difficulty evaluation, with LEVEL 1 and LEVEL 2 difficulties indicating “more than five points” and “at least one point” inside labeled bounding boxes, respectively. This aligns with the design of the balance criterion (Stage 3), as the sparse point clouds or dense point clouds can be equally learned. In Table \ref{tab:crb-fully-waymo}, we report the performance of the proposed approach with a small portion of point clouds and the fully supervised baseline reported in \citep{IA-SSD}, on the Waymo dataset. From Table \ref{tab:crb-fully-waymo}, we can observe that the performance gap between the detectors trained with active learning (approx. 50K bounding box annotations) and fully supervised learning (approx. 8 million bounding box annotations) is smaller in LEVEL 2 ($7.18\%$ in LEVEL 2 vs $8.08\%$ in LEVEL 1), which aligns with the balance criteria in the proposed CRB framework.}

\begin{table*}[t]
  \caption{Performance comparisons on KITTI \textit{val} set with different SOTA AL methods when acquiring approximately 1\% queried bounding boxes. Results are reported with 3D AP with 40 recall positions. $^\dag$ indicates the reported performance of the backbone trained with the full labeled set (100\%).\vspace{-1.5ex}}
  \label{tab:uniform}
  \resizebox{1\linewidth}{!}{%
  \centering
  \begin{tabular}{lcc ccc ccc ccc ccc}
    \toprule
       & \multicolumn{3}{c}{Car} &\multicolumn{3}{c}{Pedestrian}
       &\multicolumn{3}{c}{Cyclist}
        &\multicolumn{3}{c}{Average} \\
     \cmidrule(l){2-4}\cmidrule(l){5-7}\cmidrule(l){8-10}\cmidrule(l){11-13} 
     Method &\textsc{Easy} &\textsc{Mod.} &\textsc{Hard} &\textsc{Easy} &\textsc{Mod.} &\textsc{Hard} &\textsc{Easy} &\textsc{Mod.} &\textsc{Hard} & \textsc{Easy} &\textsc{Mod.} &\textsc{Hard}\\
    \midrule
    \textsc{Llal}
    &$89.95$ &$78.65$ &$\textbf{75.32}$ &$56.34$ &$49.87$ &$45.97$ &$75.55$ &$60.35$ &$55.36$ &$73.94$ &$62.95$ &$58.88$\\
    \textsc{Coreset}         
    &$87.77$ &$77.73$ &$72.95$ &$47.27$ &$41.97$ &$38.19$ &$81.73$ &$59.72$ &$55.64$ &$72.26$ &$59.81$ &$55.59$\\
    \textsc{Badge}         
    &$89.96$ &$75.78$ &$70.54$ &$51.94$ &$46.24$ &$40.98$ &$84.11$ &$62.29$ &$58.12$ &$75.34$ &$61.44$ &$56.55$\\
    \textsc{Crb}   
    &$\textbf{90.98}$ &$\textbf{79.02}$ &$74.04$ &$\textbf{64.17}$ &$\textbf{54.80}$ &$\textbf{50.82}$ &$\textbf{86.96}$ &$\textbf{67.45}$ &$\textbf{63.56}$ &$\textbf{80.70}$ &$\textbf{67.81}$ &$\textbf{62.81}$\\
    \midrule
    \textsc{Pv-rcnn}$^\dag$       
    &$92.56$ &$84.36$ &$82.48$ &$64.26$ &$56.67$ &$51.91$ &$88.88$ &$71.95$ &$66.78$ &$81.75$ &$70.99$ &$67.06$\\
  \bottomrule \vspace{-4ex}
\end{tabular}
}
\end{table*}

\textcolor{black}{
\subsection{More Discussions on the Motivation of the Stage 2}\label{sec:motivation}
Our main objective of Stage 2, \textit{i.e.}, Representative Prototype Selection is to determine a subset $\mathcal{D}_{S_2}^*$ from the pre-selected set $S_1$ in the last stage, by minimizing the set discrepancy in the latent feature space. However, the test features are not observable during the training phase, and it is hard to guarantee that the feature distribution can be comprehensively captured. As stated in Remark section, we focus on the features that are not learned well from the training set  due to the zero training error assumption and reconsider the feature matching problem from a gradient perspective. In particular, we split the test set into two group In the gradient space: (1) seen test samples that can be easily recognized will cluster near the origin, (2) while the novel test samples will diversely distribute in the subspace. As the first group of samples have been sufficiently covered by the initiated, in this stage, we focus on finding matching with the latter group. By assuming the prior distribution of gradients follows a Gaussian distribution, finding the K-metroids is naturally a choice to mitigate the gap between mean and variance. K-metroids algorithm breaks the dataset up into groups and attempts to minimize the distance between points labeled to be in a cluster and a point designated as the center of that cluster (\textit{i.e.}, prototype). By selecting the prototypes in the second stage, we implicitly bridge the gap between the selected set and the test set at a latent feature level. }

\section{Implementation Details}\label{sec:imple}
\begin{wraptable}{r}{0.4\textwidth}
\vspace{-3ex}
  \caption{Comparing the performance of detectors with active learning (AL) by \textsc{Crb} and fully supervised learning (FSL) on Waymo \textit{val} set. Results (mAP $\%$) are calculated by Waymo official evaluation metric.}
  \label{tab:crb-fully-waymo}
\resizebox{1\linewidth}{!}{
\begin{tabular}{l c c}
    \toprule
    Method & mAP Level 1 &  mAP Level 2 \\
    \midrule
     \textsc{Crb} & $58.60$ & $52.65$
     \\
     FSL & $66.68$ & $59.83$
     \\
    \midrule
     Gap ($\downarrow$)  & $-8.08$ & $-7.18$\\
  \bottomrule
  \vspace{-2ex}
    \end{tabular}
}    
\end{wraptable}

\subsection{Evaluation Metrics.} 
To fairly evaluate baselines and the proposed method on KITTI dataset \citep{DBLP:conf/cvpr/GeigerLU12}, we follow the work of~\citep{DBLP:conf/cvpr/ShiGJ0SWL20}: we utilize Average Precision (AP) for 3D and bird eye view (BEV) detection, and the task difficulty is categorized to \textsc{Easy}, \textsc{Moderate}, and \textsc{Hard}, with a rotated IoU threshold of $0.7$ for cars and $0.5$ for pedestrian and cyclists. The results evaluated on the validation split are calculated with $40$ recall positions. To evaluate on Waymo dataset \citep{DBLP:conf/cvpr/SunKDCPTGZCCVHN20}, we adopt the officially published evaluation tool for performance comparisons, which utilizes AP and the average precision weighted by heading (APH). The respective IoU thresholds for vehicles, pedestrians, and cyclists are set to 0.7, 0.5, and 0.5. Regarding detection difficulty, the Waymo test set is further divided into two levels. \textsc{Level 1} (and \textsc{Level 2}) indicates there are more than five inside points (at least one point) in the ground-truth objects.

\subsection{Implementation Details of Training}

To ensure the reproducibility of the baselines and the proposed approach, we develop a PyTorch-based active 3D detection toolbox (attached in the supplemental material) that implements mainstream AL approaches and can accommodate most of the public benchmark datasets. For fair comparison, all active learning methods are constructed from the \textsc{Pv-rcnn}~\citep{DBLP:conf/cvpr/ShiGJ0SWL20} backbone. All experiments are conducted on a GPU cluster with three V100 GPUs. The runtime for an experiment on KITTI and Waymo is around 11 hours and 100 hours, respectively. Note that, training \textsc{Pv-rcnn}on the full set typically requires 40 GPU hours for KITTI and 800 GPU hours for Waymo.

\noindent\textbf{Parameter Settings}. The batch sizes for training and evaluation are fixed to 6 and 16 on both datasets. The Adam optimizer is adopted with a learning rate initiated as 0.01, and scheduled by one cycle scheduler. The number of \textsc{Mc-dropout} stochastic passes is set to 5 for all methods. 

\noindent\textbf{Active Learning Protocols}. \textcolor{black}{As our work is the first comprehensive study on active 3D detection task, the active training protocol for all AL baselines and the proposed method is empirically defined.} For all experiments, we first randomly select $m$ fully labeled point clouds from the training set as the initial $\mathcal{D}_L$. With the annotated data, the 3D detector is trained with $E$ epochs, which is then freezed to select $N_r$ candidates from $\mathcal{D}_U$ for label acquisition. \textcolor{black}{We set the $m$ and $N_r$ to 2.5~3\%  point clouds (\textit{i.e.}, $N_r=m=$100 for KITTI, $N_r=m=$400 for Waymo) to trade-off between reliable model training and high computational costs.}The aforementioned training and selection steps will alternate for $R$ rounds. Empirically, we set $E=30$, $R=6$ for KITTI, and fix $E=40$, $R=5$ for Waymo.
\subsection{Implementation Details of Baselines and \textsc{Crb}}\label{sec:implementation}
In this section, we introduce more implementation details of both baselines and the proposed \textsc{Crb}. 

\noindent \textbf{\textsc{Crb}}. In comparison with baselines as reported in Figure 2, the $\mathcal{K}_{1}$ and $\mathcal{K}_{2}$ are empirically set to $300$, $200$ for KITTI and $2,000$ and $1,200$ for Waymo. The gradient maps used for $\textsc{Rps}$ are extracted from the second convolutional layer in the shared block of \textsc{Pv-rcnn}. Three dropout layers in \textsc{Pv-rcnn} are enabled during the \textsc{Mc-dropout} and the dropout rate is fixed to 0.3 for both datasets. The number of \textsc{Mc-dropout} stochastic passes are set to 5 for all methods. In the \textsc{Gpcb} stage, we measure the KL-divergence between the KDE PDF of the selected set and the uniform prior distribution of the point cloud density for each class. The goal of conducting a greedy search is to find the optimal subset that can achieve the minimum sum of KL divergence for all classes. Considering the high variance of KL divergence across different classes, we unify the scale of KL-divergence to $\bar{d}_c$ by applying the following function,
\begin{align*}
    \bar{d}_c = \frac{2}{\pi} \arctan{\frac{\pi}{2} d_c},
\end{align*}
where $d_c$ denotes the KL-divergence for the $c$-th class. To this end, the ultimate objective for greedy search is $\argmin_{\mathcal{D}_S\subset\mathcal{D}_{S_2}} \sum_{c\in[C]}\bar{d}_c$. The normalized measurement can avoid dominance by any single class.

\noindent \textbf{\textsc{Coreset}} \citep{DBLP:conf/iclr/SenerS18}. The embeddings extracted for both labeled and unlabeled data are the output from the shared block, with the dimension of $128$ by $256$. The \textsc{Coreset} adopts the furthest-first traversal for k-Center clustering strategy, which computes the Euclidean distance between each embedding pair.

\noindent \textbf{\textsc{Llal}} \citep{DBLP:conf/cvpr/YooK19}. For implementing the loss prediction module in \textsc{LLAL}, we construct a two-block module that connects to two layers of the \textsc{Pv-rcnn}, which takes multi-level knowledge into consideration for loss prediction. Particularly, each block consists of a convolutional layer with a channel size of $265$ and a kernel size of $1$, a batchnorm layer, and a relu activation layer. The outputs are then concatenated and fed to a fully connected layer and map to a loss score. All real loss for each training data point is saved and serves as the ground-truth to train the loss prediction module.

\noindent \textbf{\textsc{Badge}}. According to \citep{DBLP:conf/iclr/AshZK0A20}, hypothetical labels for the classifier are determined by the classes with the highest predicted probabilities. The gradient matrix with the dimension $256$ by $256$ for each unlabeled point cloud is extracted from the last convolutional layer of the \textsc{Pv-rcnn}'s classification head and then fed into the \textsc{Badge} algorithm.

\section{Proof of Theorem 2.1}

\begin{theorem}
\label{theo:crb}
Let $\mathcal{H}$ be a hypothesis space of Vapnik-Chervonenkis (VC) dimension $d$, with $f$ and $g$ being the classification and regression branches, respectively. The $\widehat{\mathcal{D}}_{S}$ and $\widehat{\mathcal{D}}_T$ represent the empirical distribution induced by samples drawn from the acquired subset $\mathcal{D}_S$ and the test set $\mathcal{D}_T$, and $\ell$ the loss function bounded by $\mathcal{J}$. It is proven that $\forall$ $\delta \in (0, 1)$, and $\forall f, g\in \mathcal{H}$, with probability at least $1-\delta$ the following inequality holds, 
\vspace{-1ex}
\begin{align*}
        \mathfrak{R}_T[\ell(f, g;\bm{w})]\leq \mathfrak{R}_S[\ell(f, g;\bm{w})] +  \frac{1}{2}disc(\widehat{\mathcal{D}}_{S}, \widehat{\mathcal{D}}_T) +\lambda^* + \text{const},
\end{align*}
where $\text{const} = 3 \mathcal{J} (\sqrt{\frac{\log \frac{4}{\delta}}{2 N_r}} + \sqrt{\frac{\log \frac{4}{\delta}}{2 N_t}}) + \sqrt{\frac{2d \log(e N_r/d)}{N_r}} + \sqrt{\frac{2d \log(e N_t/d)}{N_t}}$.

Notably, $\lambda^* = \mathfrak{R}_T[\ell(f^*, g^*; \bm{w}^*)] +  \mathfrak{R}_S[\ell(f^*, g^*; \bm{w}^*)]$ denotes the joint risk of the optimal hypothesis $f^*$ and $g^*$, with $\bm{w}^*$ being the model weights. $N_r$ and $N_t$ indicate the number of samples in the $\mathcal{D}_S$ and $\mathcal{D}_T$. The proof can be found in the supplementary material.
\end{theorem}

\begin{proof}
        For brevity, we omit the model weights $\bm{w}$ in the following proof. Based on the triangle inequality of $\ell$ and the definition of the discrepancy distance $disc(\cdot, \cdot)$, the following inequality holds,
        \begin{align*}
            \mathfrak{R}_T[\ell(f, g)]&\leq \mathfrak{R}_T[\ell(f^*, g^*)] + \frac{1}{2} \mathfrak{R}_T[\ell(f, f^*)] + \frac{1}{2} \mathfrak{R}_T[\ell(g, g^*)]\\
            &\leq \mathfrak{R}_T[\ell(f^*, g^*)] + \mathfrak{R}_S[\ell(f^*, g^*)] + \frac{1}{2} |\mathfrak{R}_T[\ell(f, f^*)]  - \mathfrak{R}_S[\ell(f, f^*)]| \\&+ \frac{1}{2} |\mathfrak{R}_T[\ell(g, g^*)] - \mathfrak{R}_S[\ell(g, g^*)]|\\
            &\leq \mathfrak{R}_T[\ell(f^*, g^*)] + \mathfrak{R}_S[\ell(f^*, g^*)]  + \frac{1}{2} disc(\mathcal{D}_S, \mathcal{D}_T)\\
            &\leq \mathfrak{R}_T[\ell(f^*, g^*)] + \mathfrak{R}_S[\ell(f, g)] + \mathfrak{R}_S[\ell(f^*, g^*)] + \frac{1}{2} disc(\mathcal{D}_S, \mathcal{D}_T).
        \end{align*}
        By defining the joint risk of the optimal hypothesis $\lambda^* = \mathfrak{R}_T[\ell(f^*, g^*)] +  \mathfrak{R}_S[\ell(f^*, g^*)]$ and the Corollary 6 in \citep{DBLP:conf/colt/MansourMR09}, we have,
        \begin{align*}
            \mathfrak{R}_T[\ell(f, g)]&\leq  \mathfrak{R}_S[\ell(f, g)] + \frac{1}{2}disc(\mathcal{D}_S, \mathcal{D}_T) + \lambda^*\\
            &\leq  \mathfrak{R}_S[\ell(f, g)]  + \frac{1}{2}disc(\widehat{\mathcal{D}}_S, \widehat{\mathcal{D}}_T) + \lambda^* + 4q (\text{Rad}_S(\mathcal{H}) + \text{Rad}_{T}(\mathcal{H})) \\&+ 3 \mathcal{J} (\sqrt{\frac{\log \frac{4}{\delta}}{2 N_r}} + \sqrt{\frac{\log \frac{4}{\delta}}{2 N_t}}),
        \end{align*}
        where $N_r$ and $N_t$ indicate the sample size of the selected set and the test set, respectively. $q$ stands for the function is $q$-Lipschitz. As our regression loss, $\ell^{reg}$ is the smooth-L1 loss function and bounded by $\mathcal{J}$, $q$ equals $1$ in our case. $\text{Rad}_S(\mathcal{H})$ and $\text{Rad}_T(\mathcal{H})$ indicates the empirical Rademacher complexity of a hypothesis set $\mathcal{H}$ whose VC dimension is $d$ over the selected set and the test set.
        
        Considering the Rademacher complexity is bounded by:
        \begin{align*}
            \text{Rad}_S(\mathcal{H}) \leq \sqrt{\frac{2d \log(e N_r/d)}{N_r}}, \quad \text{Rad}_T(\mathcal{H}) \leq \sqrt{\frac{2d \log(e N_t/d)}{N_t}},
        \end{align*}
        then we can rewrite the inequality as,
        \begin{align*}
            \mathfrak{R}_T[\ell(f, g)]\leq \mathfrak{R}_S[\ell(f, g)]  + \frac{1}{2}disc(\widehat{\mathcal{D}}_S, \widehat{\mathcal{D}}_T) + \lambda^* + const,
        \end{align*}
        where $const = 3 \mathcal{J} (\sqrt{\frac{\log \frac{4}{\delta}}{2 N_r}} + \sqrt{\frac{\log \frac{4}{\delta}}{2 N_t}}) + \sqrt{\frac{2d \log(e N_r/d)}{N_r}} + \sqrt{\frac{2d \log(e N_t/d)}{N_t}}$.
    \end{proof}

\begin{algorithm}[t]
\caption{The algorithm of CRB for active 3D object detection}\label{alg:CRB}
\begin{algorithmic}
\Inputs{
 $ \mathcal{D}_L$: initially labeled point clouds \\
$\mathcal{D}_U$: unlabeled pool of point clouds \\
$\bm{\Omega}$: oracle \\
$B$: total budget of active selection\\ 
$e(\cdot): \mathcal{P}\rightarrow \bm{x}$: point cloud encoder of 3D detector \\
$f(\cdot)$: classifier of 3D detector\\
$g(\cdot)$: regression head of 3D detector \\
$R$: total active learning rounds
}
\\
\State $\mathcal{D}_S \gets \emptyset$
\State Pre-train 3D detector $\{e(\cdot)$, $f(\cdot)$, $g(\cdot)\}$ with $ \mathcal{D}_L$ until converge
\State $\mathcal{D}_S \gets \mathcal{D}_S \cup \mathcal{D}_L$

\For{$r \in [R]$} \Comment{For each round of active selection}

\State $\widehat{Y} \gets f\circ e(\mathcal{D}_U)$ \Comment{Get the predicted labels} 
\State $\widehat{\mathcal{B}}, \phi \gets g\circ e(\mathcal{D}_U)$ \Comment{Get the predicted boxes $\widehat{\mathcal{B}}$ and box point densities $\phi$}

\State $\bar{B} \gets$ Hypothetical labels computed by Equation (5)
\State $\mathcal{D}^{*}_{S_1} \gets \textsc{Cls}(\mathcal{D}_U, \widehat{Y})$ selects $\mathcal{K}_1$ samples via Equation (2),(3),(4) \State\Comment{Stage 1: Concise Label Sampling}
\State $\mathcal{D}^{*}_{S_2} \gets \textsc{Rps}(\mathcal{D}^{*}_{S_1}, e(\mathcal{D}^{*}_{S_1}), \bar{B})$ selects $\mathcal{K}_2$ samples via Equation (6),(7) 
\State \Comment{Stage 2:  Representative Prototype Selection}
\State $\mathcal{D}^{*}_{S} \gets \textsc{Gpdb}(\mathcal{D}^{*}_{S_2}, \widehat{\mathcal{B}}, \phi)$ selects $N_r$ samples via Equation (8),(9) \State\Comment{Stage 3: Greedy Point Cloud Density Balancing}

\State $\mathcal{D}^{*}_{S} \gets  \bm{\Omega}(\mathcal{D}^{*}_{S})$ \Comment{Query labels from oracles}
\State $\mathcal{D}_S \gets \mathcal{D}_S \cup \mathcal{D}^{*}_{S}$
\State $\mathcal{D}_U \gets \mathcal{D}_U \backslash \mathcal{D}^{*}_{S}$
\State Train 3D detector $\{e(\cdot)$, $f(\cdot)$, $g(\cdot)\}$ with $ \mathcal{D}_S$ until converge

\EndFor
\end{algorithmic}
\end{algorithm}

\section{Algorithm Description}\label{sec:alg}
To thoroughly describe the procedure of active 3D object detection by the proposed CRB, we present the Algorithm \ref{alg:CRB} in detail. Firstly, the 3D detector consisting of an encoder $\{e(\cdot)$, a classifier $f(\cdot)$, and regression heads $g(\cdot)\}$ is pre-trained with a small set $\mathcal{D}_L$ of labeled point clouds. During the stage 1: \textsc{Cls}, the pre-trained 3D detector infers all samples from the unlabeled pool $\mathcal{D}_U$ and obtains the predicted bounding boxes $\widehat{\mathcal{B}}$, predicted box labels $\widehat{Y}$, and calculated box point densities $\phi$ for each point cloud. Also, the hypothetical labels $\bar{B}$ through stochastic monte-carlo sampling are computed by Equation (5) during inference. Based on the criterion of maximizing the label entropy,  the set $\mathcal{D}^{*}_{S_1}$ containing $\mathcal{K}_1$ candidates are formed via Equations (2), (3), (4). In stage 2, we set the model to the training mode and allow the gradient back-propagation to retrieve gradients for each point cloud. Yet, the model weights will be fixed and not updated. \textsc{Rps} selects the set $\mathcal{D}^{*}_{S_2}$ of size $\mathcal{K}_2$ from the previous candidate set $\mathcal{D}^{*}_{S_1}$ based on the Equation (6) and (7). In stage 3, \textsc{Gbps} selects the set $\mathcal{D}^{*}_{S}$ of size $N_r$ from set $\mathcal{D}^{*}_{S_2}$, predicted boxes $\widehat{\mathcal{B}}$ and box point densities $\phi$ via Equation (8), (9). The final set $\mathcal{D}^{*}_{S}$ at this round is then annotated by an oracle $\bm{\Omega}$ and merged with the selected set in the previous round as the training data. Notably, the selected set at the 0-th round is $\mathcal{D}_L$. When the training data is determined, we re-train the 3D detector with the merged selected set until the model is converged. We iterate the above process starting with model inference for $R$ rounds and add  $N_r$ queried samples to the selected set $\mathcal{D}_S$ for each round.


\section{More Experimental Results on KITTI}\label{sec:kitti}

\subsection{AL performance comparisons on \textsc{Easy} difficulty level} 
\begin{figure}[!htp]%
    \subfloat{{\includegraphics[width=0.49\textwidth]{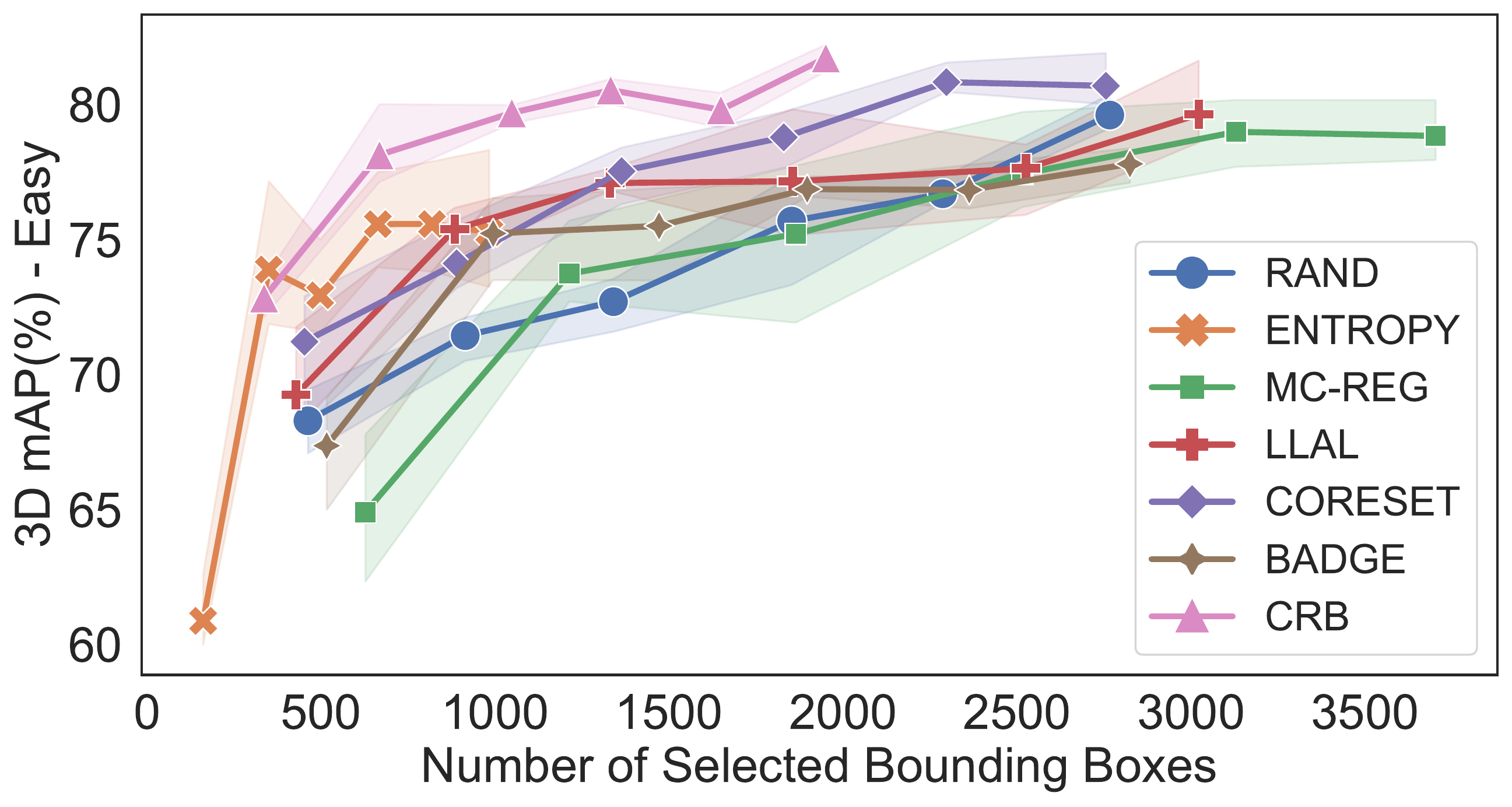} }}%
    \subfloat{{\includegraphics[width=0.49\textwidth]{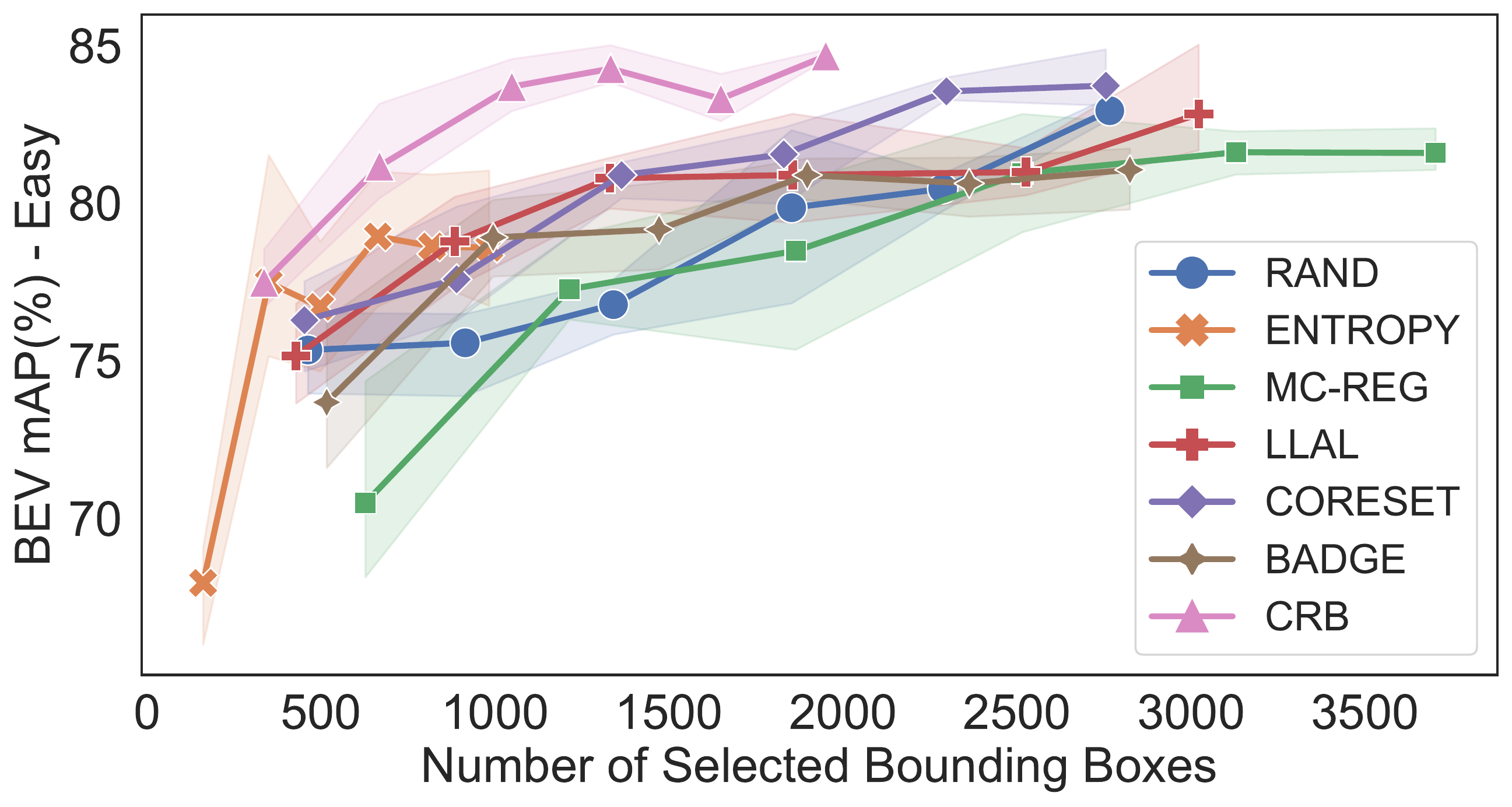} }}%
    \caption{3D and BEV mAP (\%) of \textsc{Crb} and AL baselines on the KITTI \textit{val} split at the \textsc{Easy} level.}
    \label{fig:kitti_results_box_easy}%
\end{figure}

In addition to the \textsc{Moderate} and \textsc{Hard} difficulties reported in the body text, we provide the additional quantitative analysis \textit{w.r.t.} the \textsc{Easy} mode. Figure \ref{fig:kitti_results_box_easy} depicts the mAP(\%) variation of the baselines against the proposed \textsc{Crb} with an increasing number of selected bounding boxes. The solid lines indicate the mean value from three running trials and the standard deviation are shown in the shaded area. The results indicate that with increasing annotation cost, \textsc{Crb} consistently achieves the highest mAP and outperforms the state-of-the-art active learning approaches on both 3D and BEV views. Note that \textsc{Crb} with only $1$k boxes selected for annotation reaches the comparable performance of \textsc{Rand} that selects around $3$k boxes. Other AL baselines share the same trend as the ones under the difficulties of \textsc{Moderate} and \textsc{Hard} (reported in Figure 2 of the main paper).


\subsection{AL performance comparisons for each class}
\begin{figure}[t]%
\vspace{-4ex}
    \subfloat{{\includegraphics[width=0.33\textwidth]{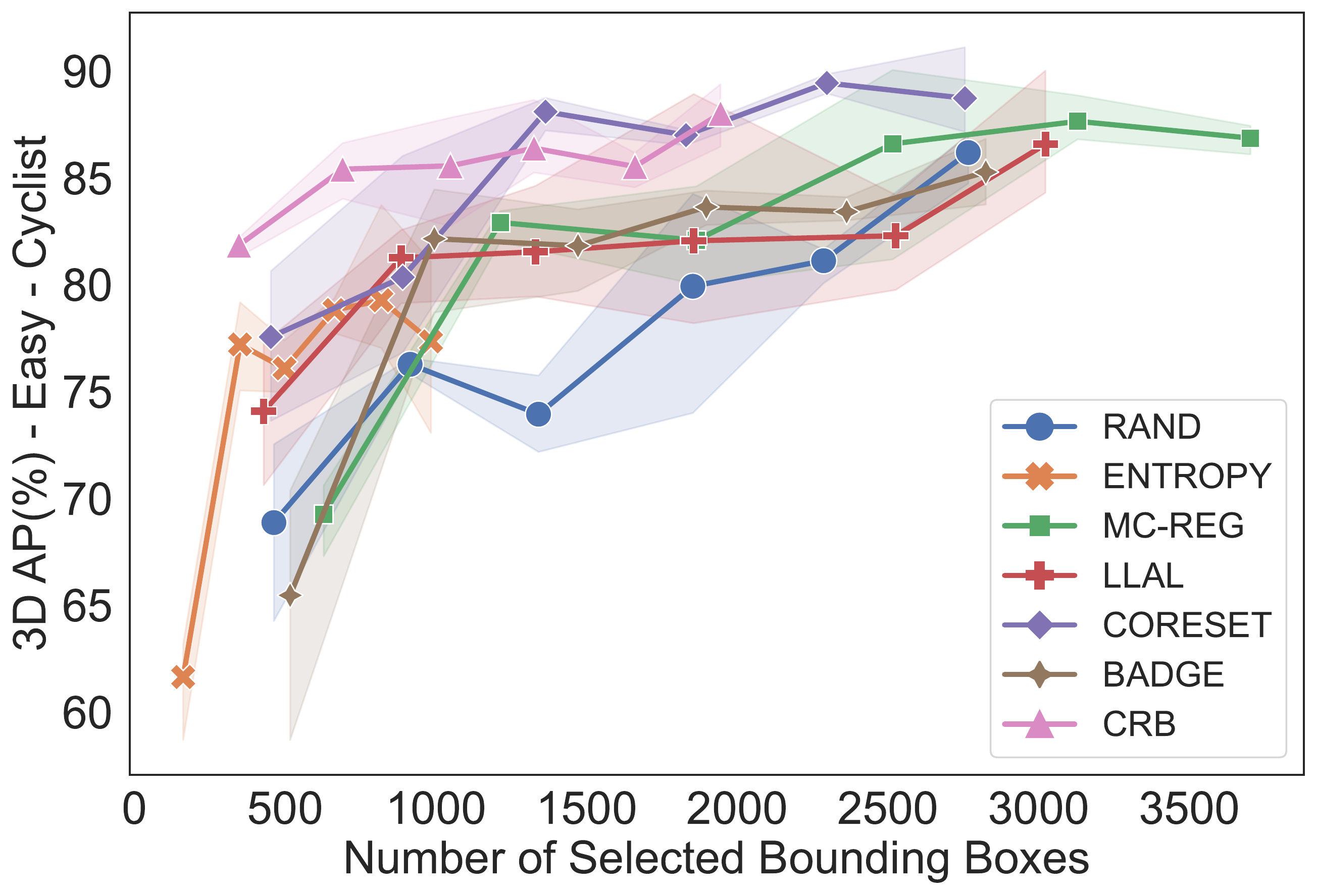} }}%
    \subfloat{{\includegraphics[width=0.33\textwidth]{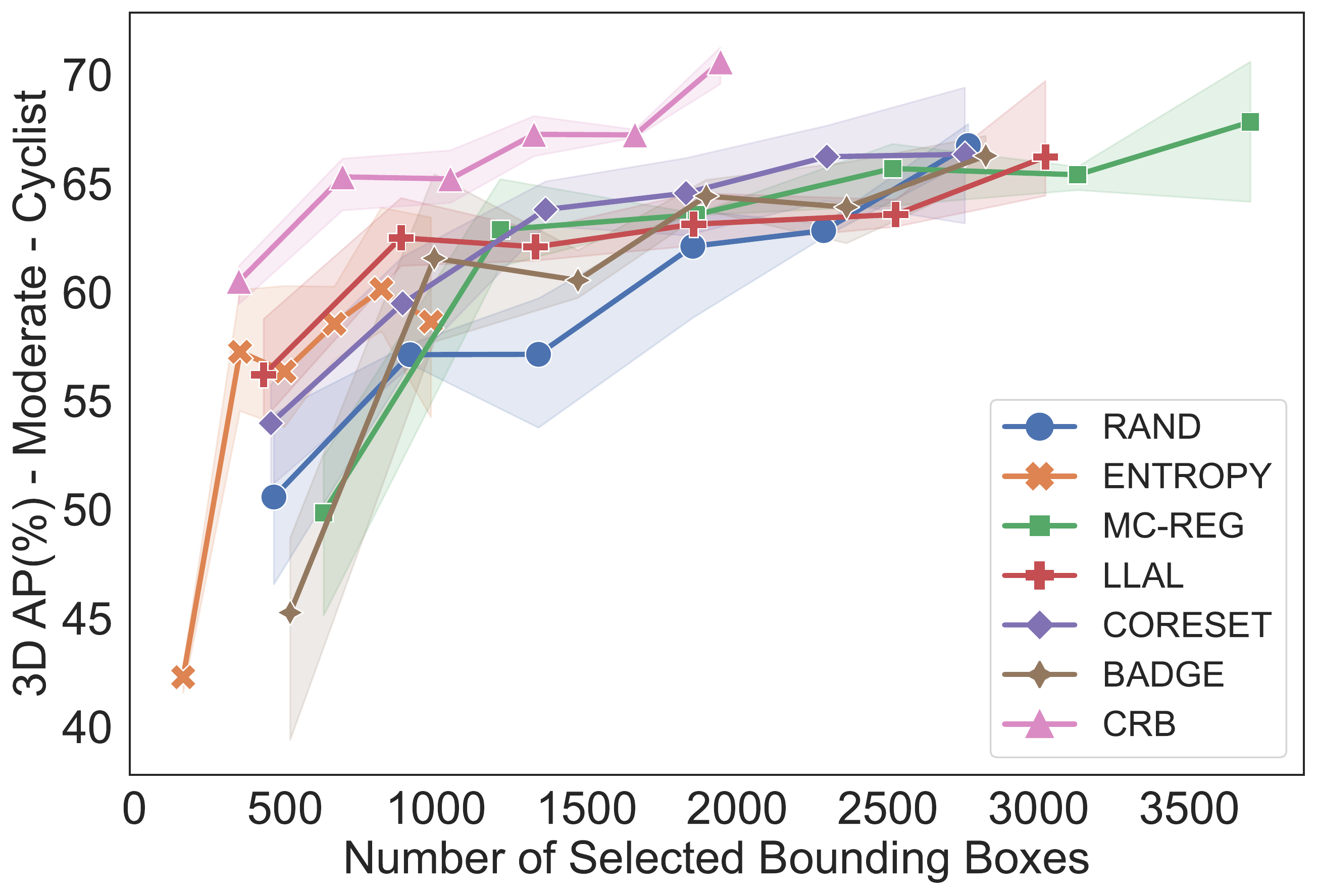} }}%
    \subfloat{{\includegraphics[width=0.33\textwidth]{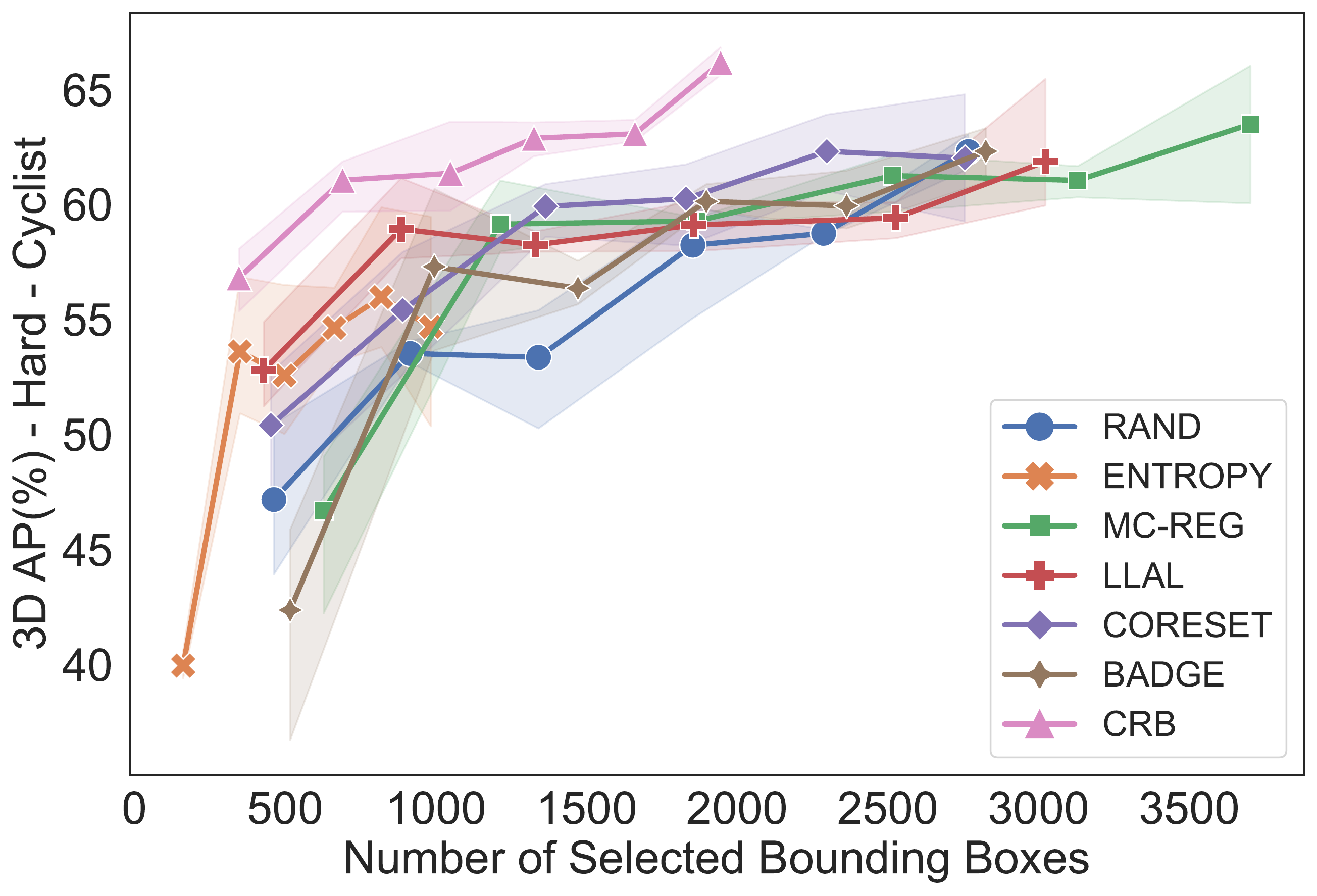} }}%
     \hfill
    \subfloat{{\includegraphics[width=0.33\textwidth]{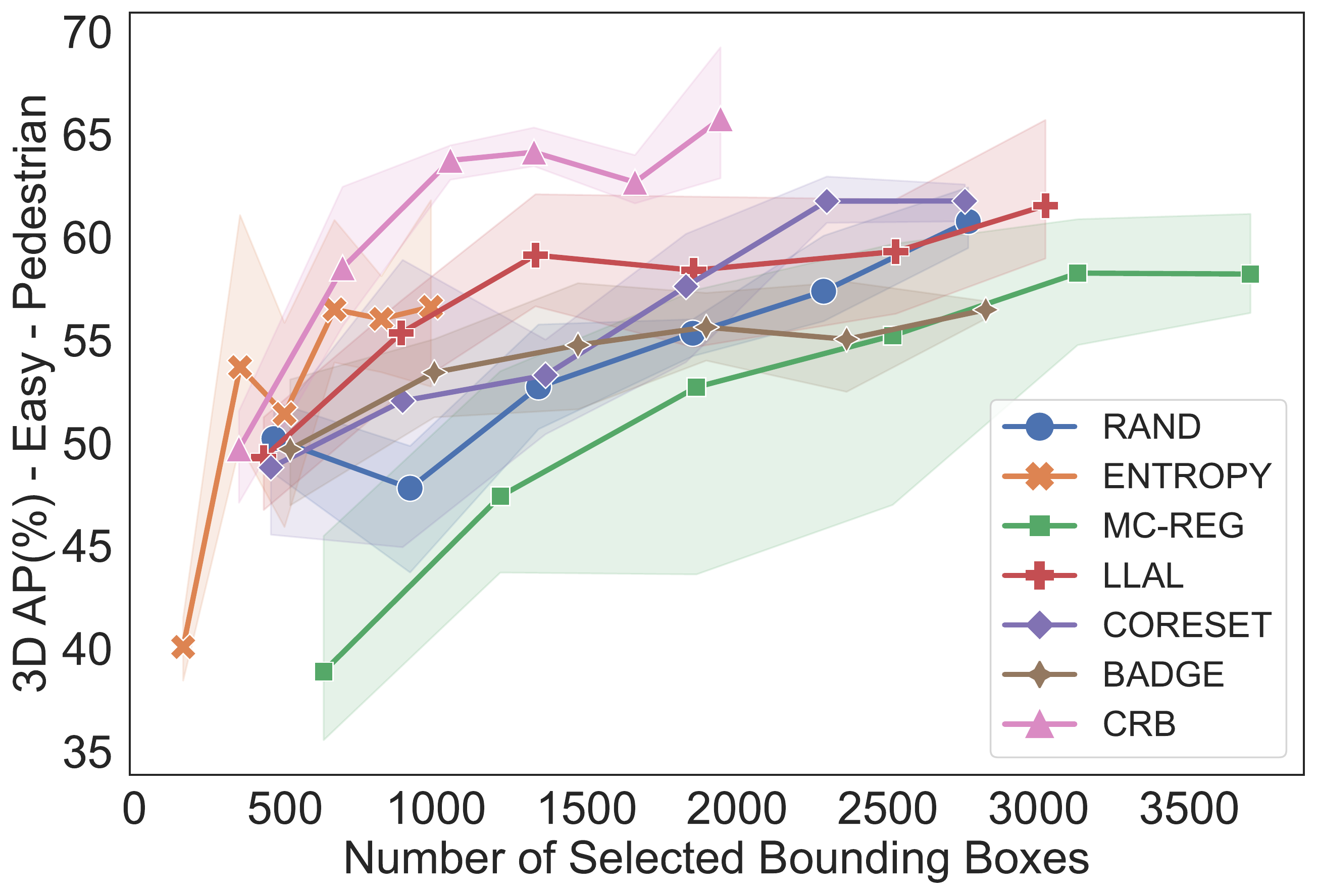} }}%
    \subfloat{{\includegraphics[width=0.33\textwidth]{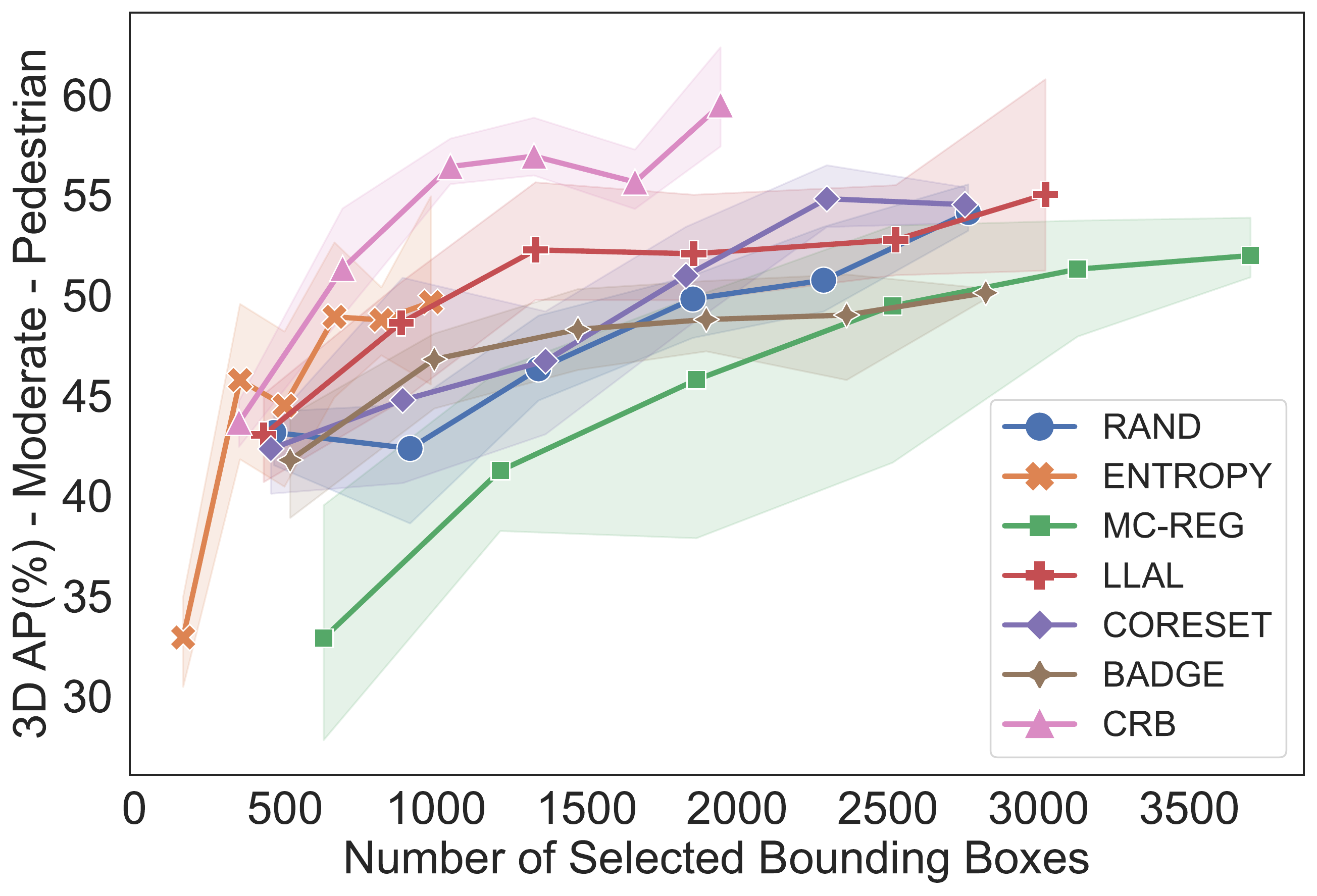} }}%
    \subfloat{{\includegraphics[width=0.33\textwidth]{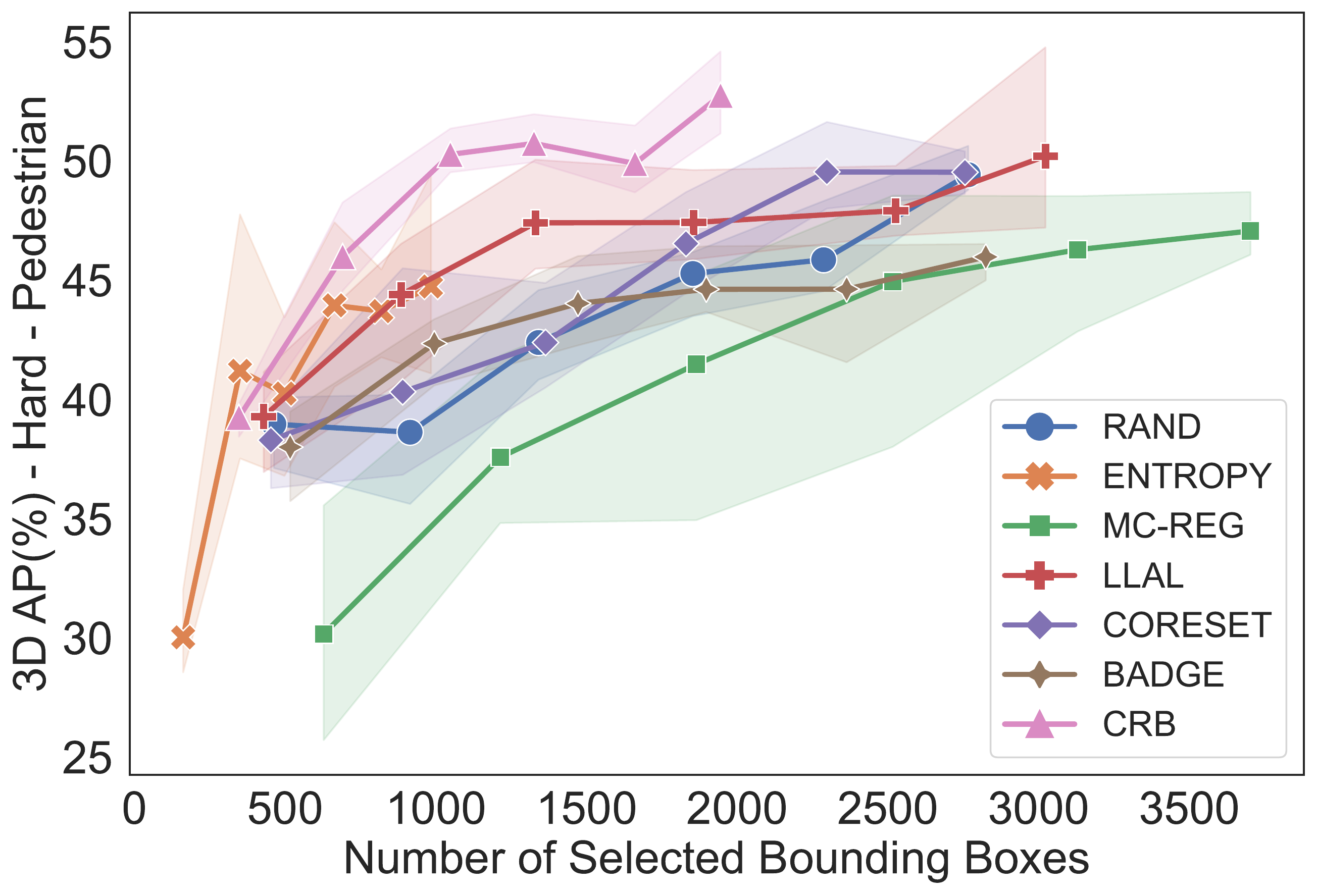} }}%
    \caption{Detection results of different classes on the KITTI \textit{val} set (3D view) with an increasing number of queried bounding boxes.}%
    \vspace{-3ex}
    \label{fig:kitti_cls_results_boxes_3D}%
\end{figure}

\begin{figure}[h]%
    \subfloat{{\includegraphics[width=0.33\textwidth]{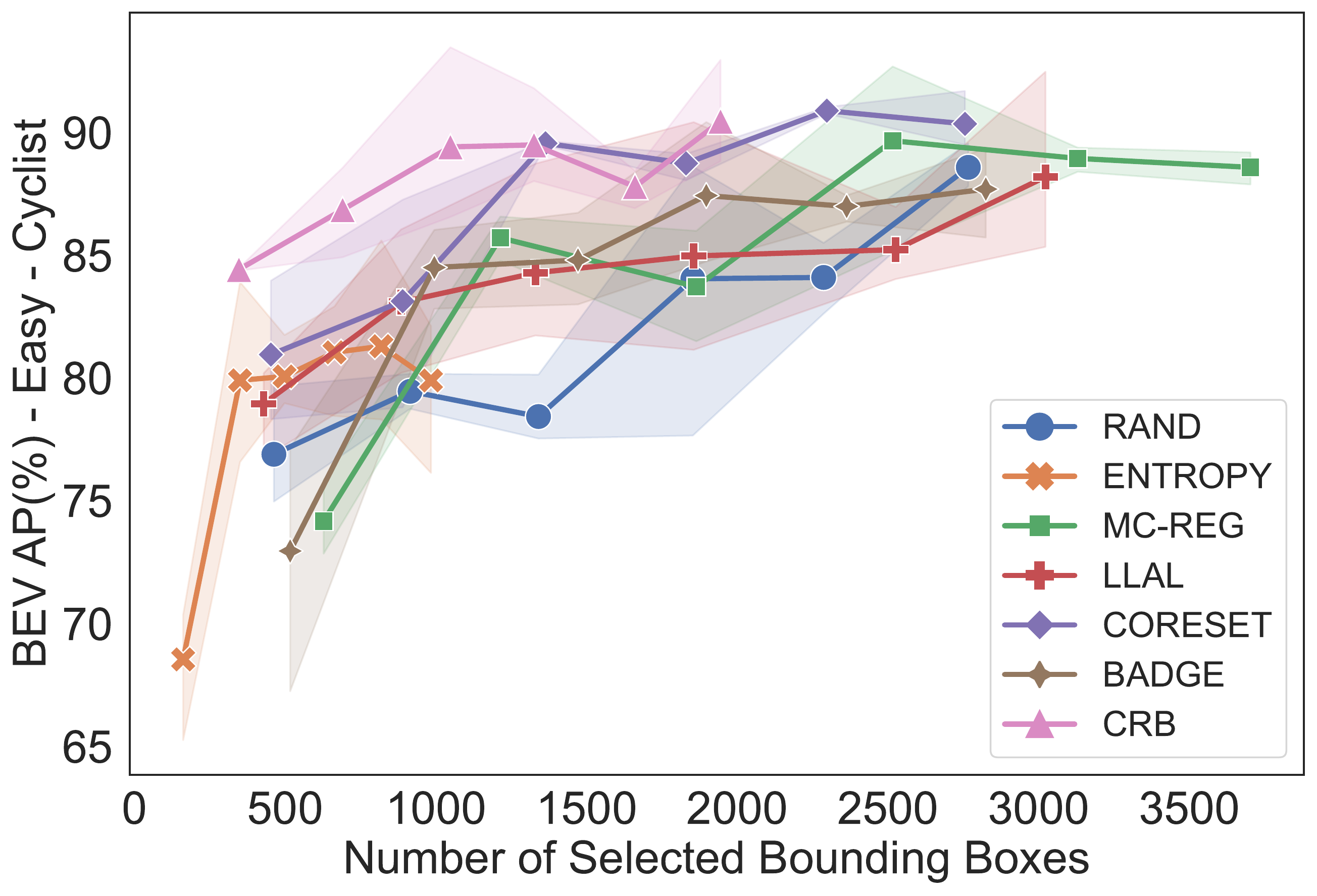} }}%
    \subfloat{{\includegraphics[width=0.33\textwidth]{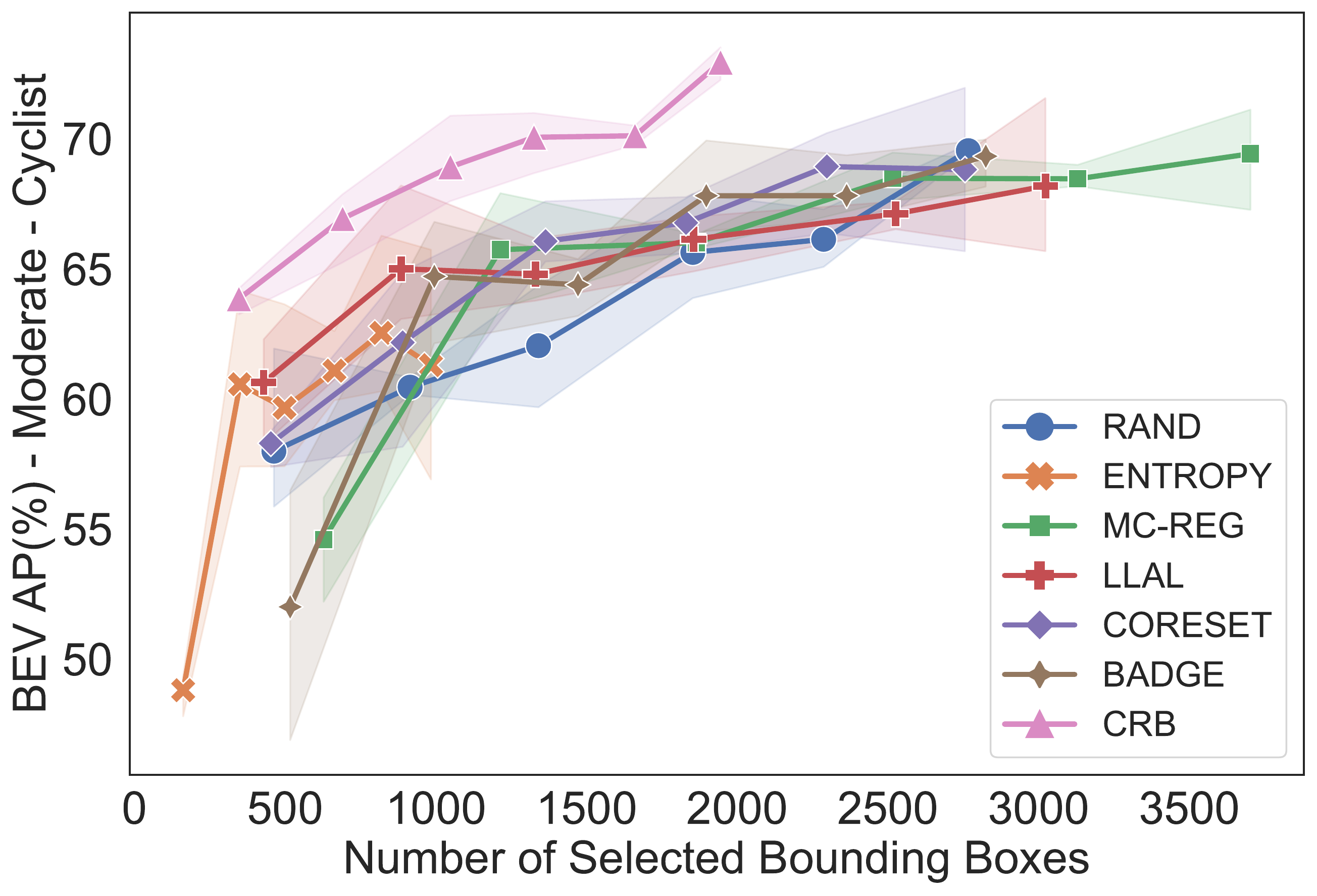} }}%
    \subfloat{{\includegraphics[width=0.33\textwidth]{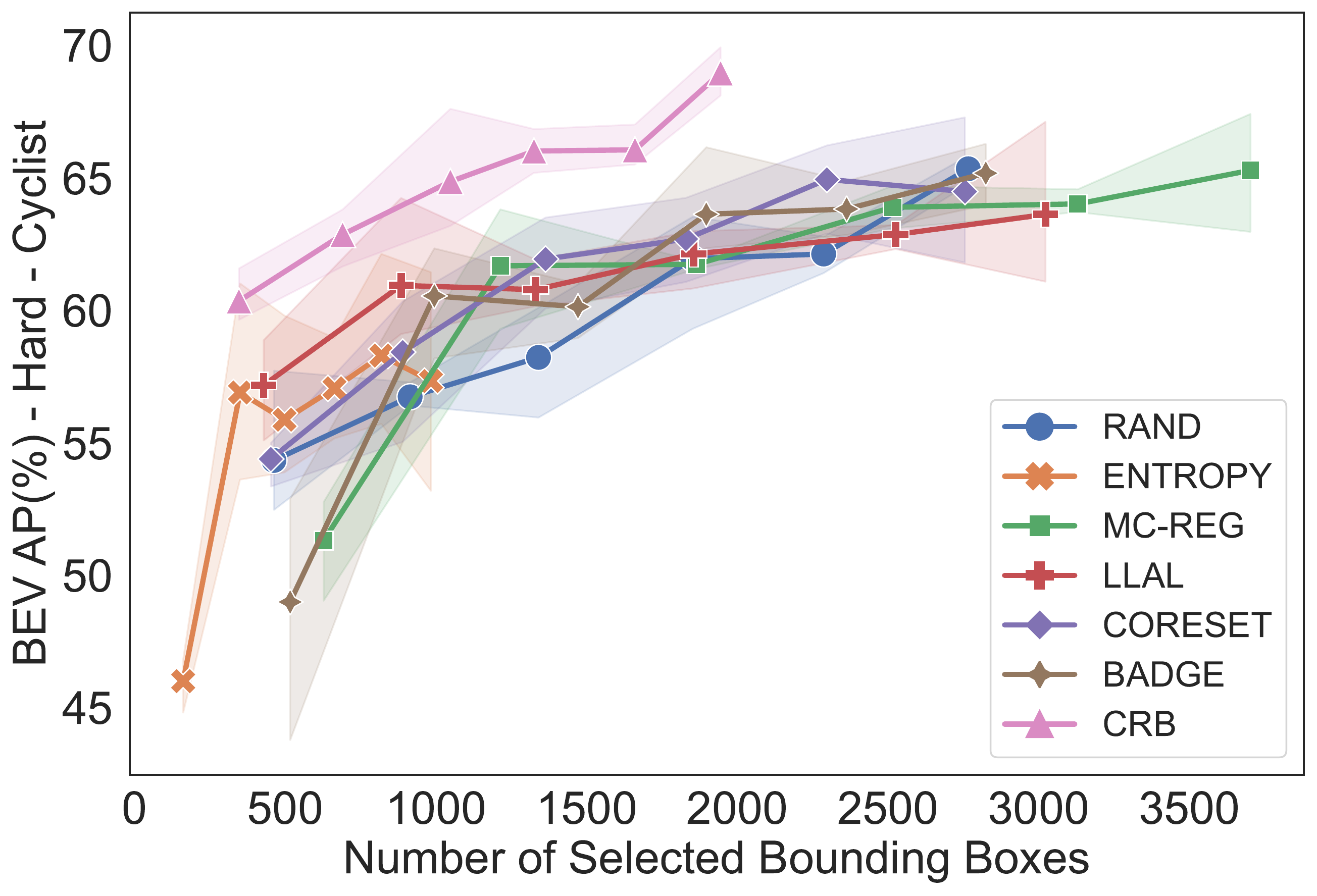} }}%
     \hfill
    \subfloat{{\includegraphics[width=0.33\textwidth]{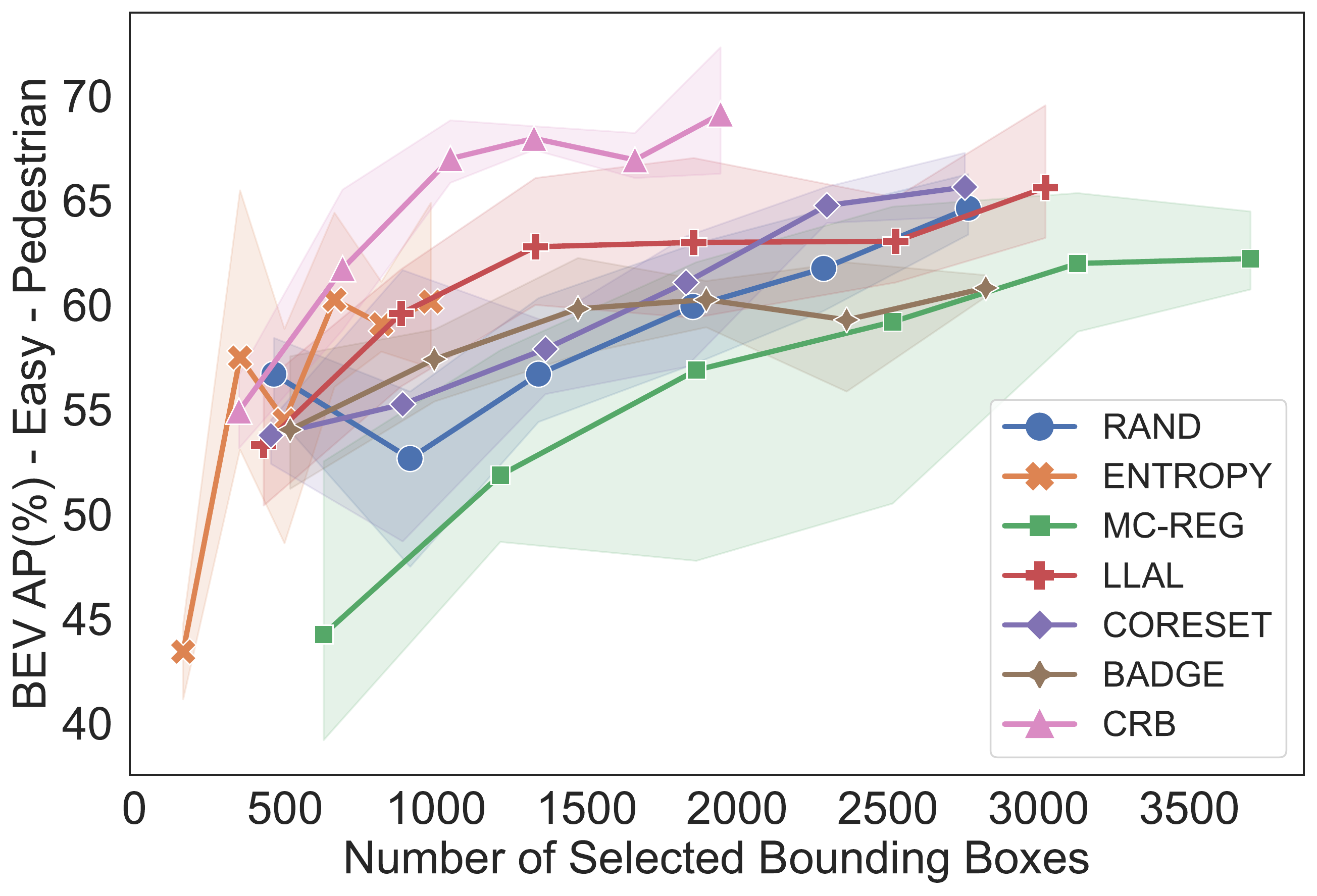} }}%
    \subfloat{{\includegraphics[width=0.33\textwidth]{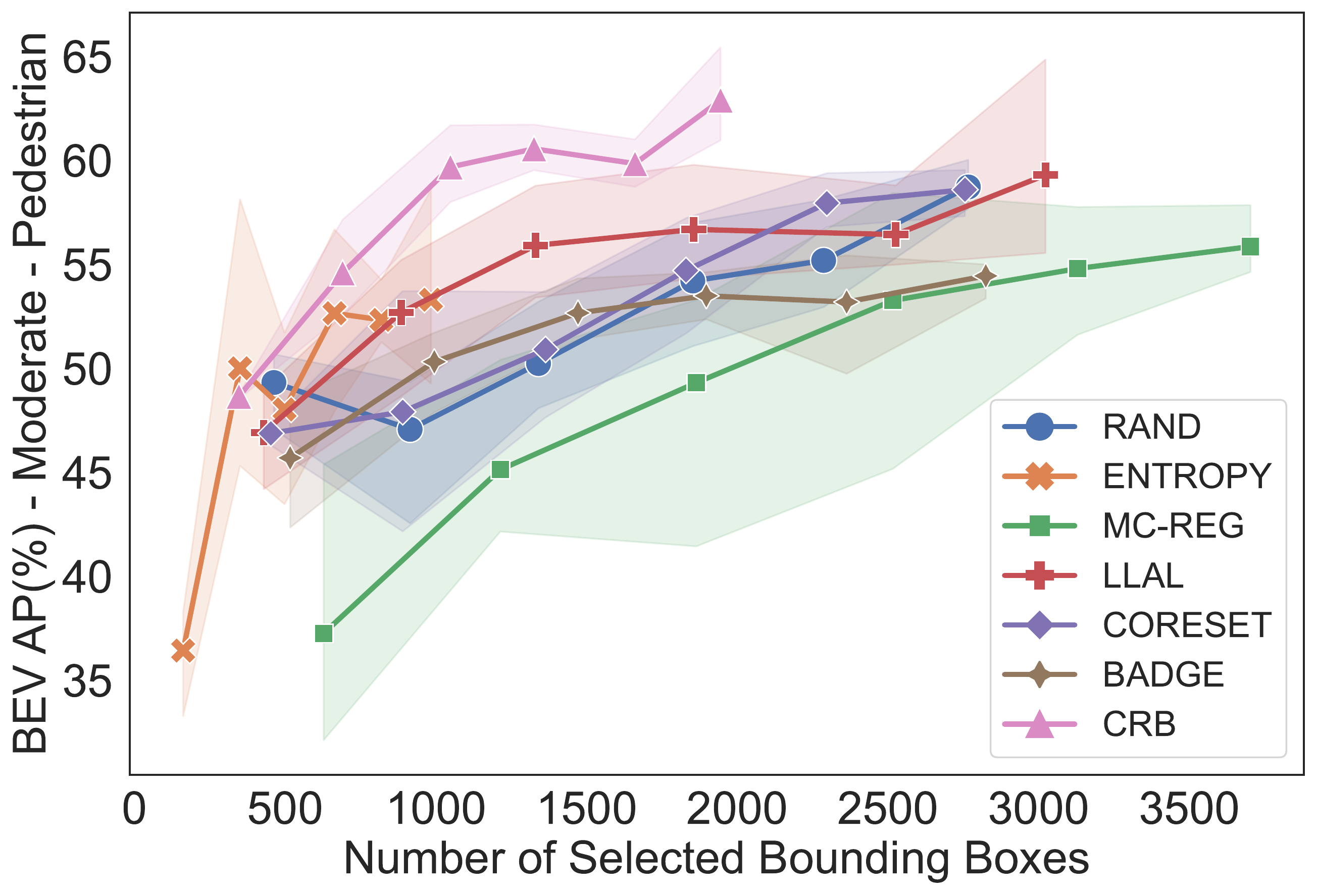} }}%
    \subfloat{{\includegraphics[width=0.33\textwidth]{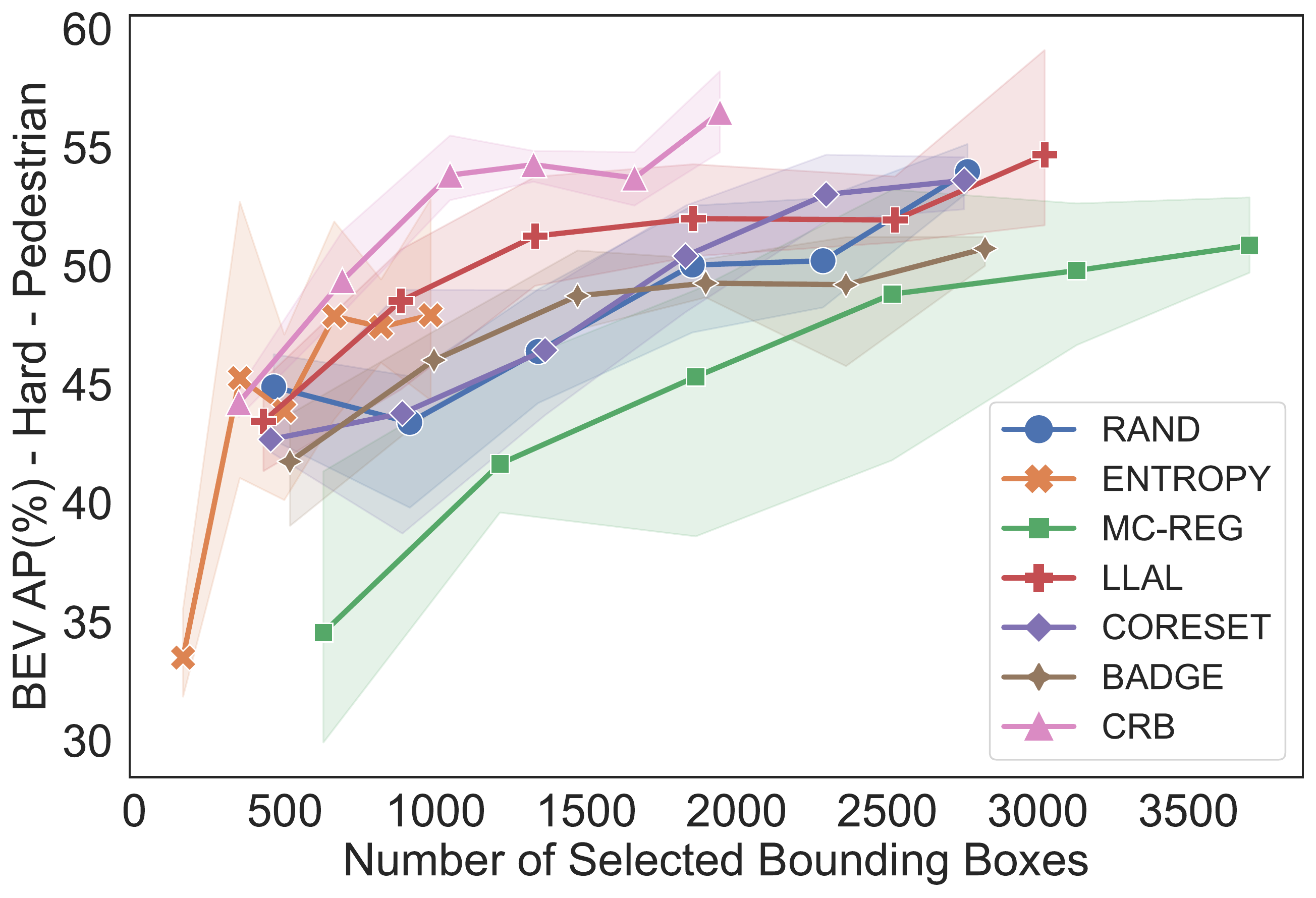} }}%
    \caption{Detection results of different classes on the KITTI \textit{val} set (BEV view) with increasing number of queried bounding boxes.}%
    \vspace{-3ex}
    \label{fig:kitti_cls_results_boxes_BEV}%
\end{figure}

To investigate the effectiveness of AL strategies on detecting specific classes, we plot the results of Cyclist and Pedestrian at all difficulty levels in Figure \ref{fig:kitti_cls_results_boxes_3D} (3D AP) and Figure \ref{fig:kitti_cls_results_boxes_BEV} (BEV AP).  We mainly compare three aspects: performance, annotation cost and error variance. 1)Performance: the plots in Figure \ref{fig:kitti_cls_results_boxes_3D} and Figure \ref{fig:kitti_cls_results_boxes_BEV} show that the proposed \textsc{Crb} outperforms all state-of-the-art AL methods by a noticeable margin, for all settings of difficulty, classes and views, except at easy cyclist. This evidences that our proposed AL approach explores samples with more conceptual semantics covering test sets so that the detector tends to perform better on more challenging samples. 2) Annotation cost: all the plots consistently demonstrate that the proposed \textsc{Crb} reaches comparable performance while requiring very few ($\sim$1/3) annotation costs as baselines, except \textsc{Entropy}. \textsc{Entropy} takes the minimal annotation cost, yet 
its result is inferior, especially for difficult classes like Cyclist. 3) Variance: we observe that AP variance of \textsc{Crb} is lower than all baselines, which shows that our method is less sensitive to randomness and more stable to produce expected results.

\subsection{AL performance comparisons for each active selection round}
\begin{figure}[h]%
    \subfloat{{\includegraphics[width=0.33\textwidth]{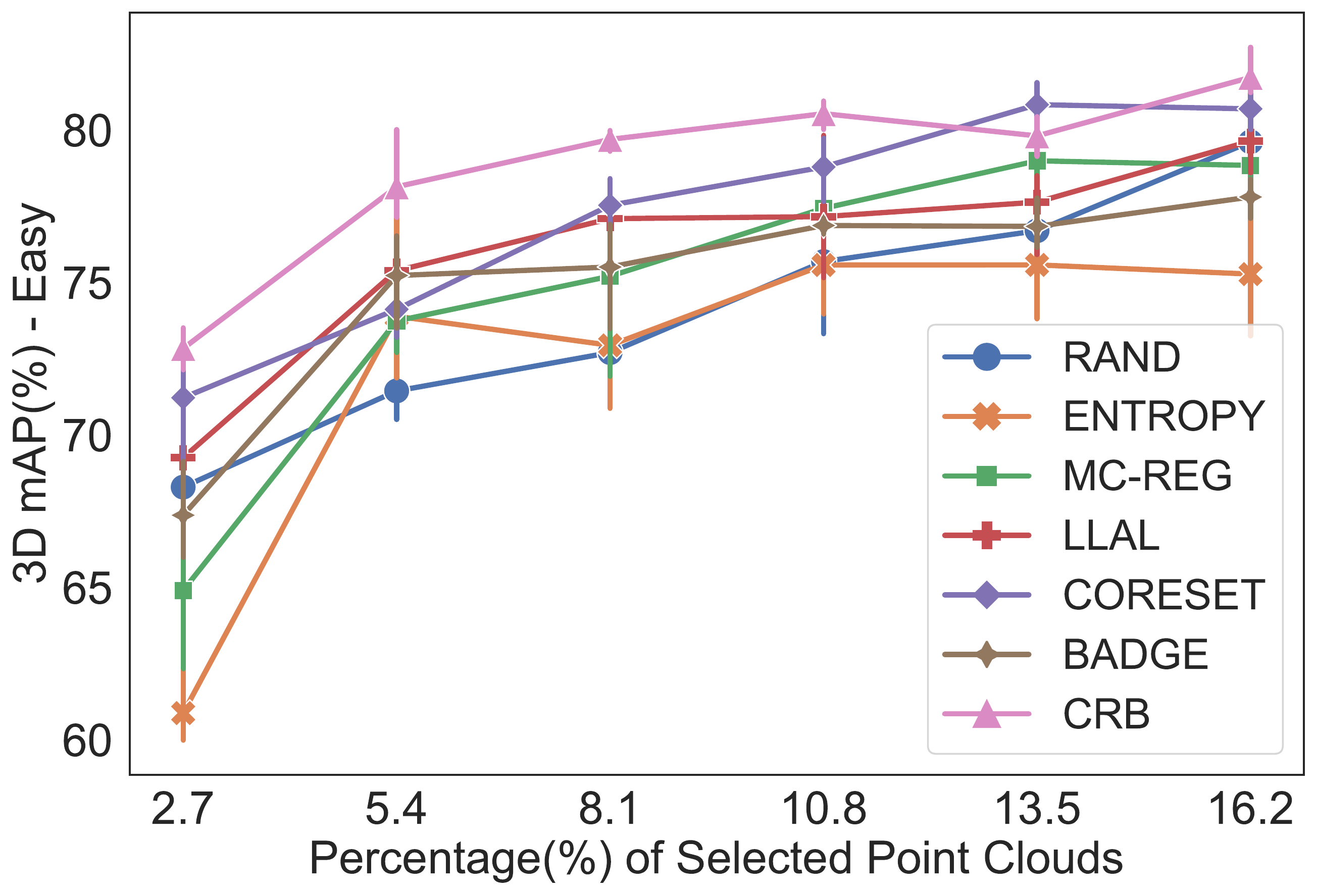} }}%
    \subfloat{{\includegraphics[width=0.33\textwidth]{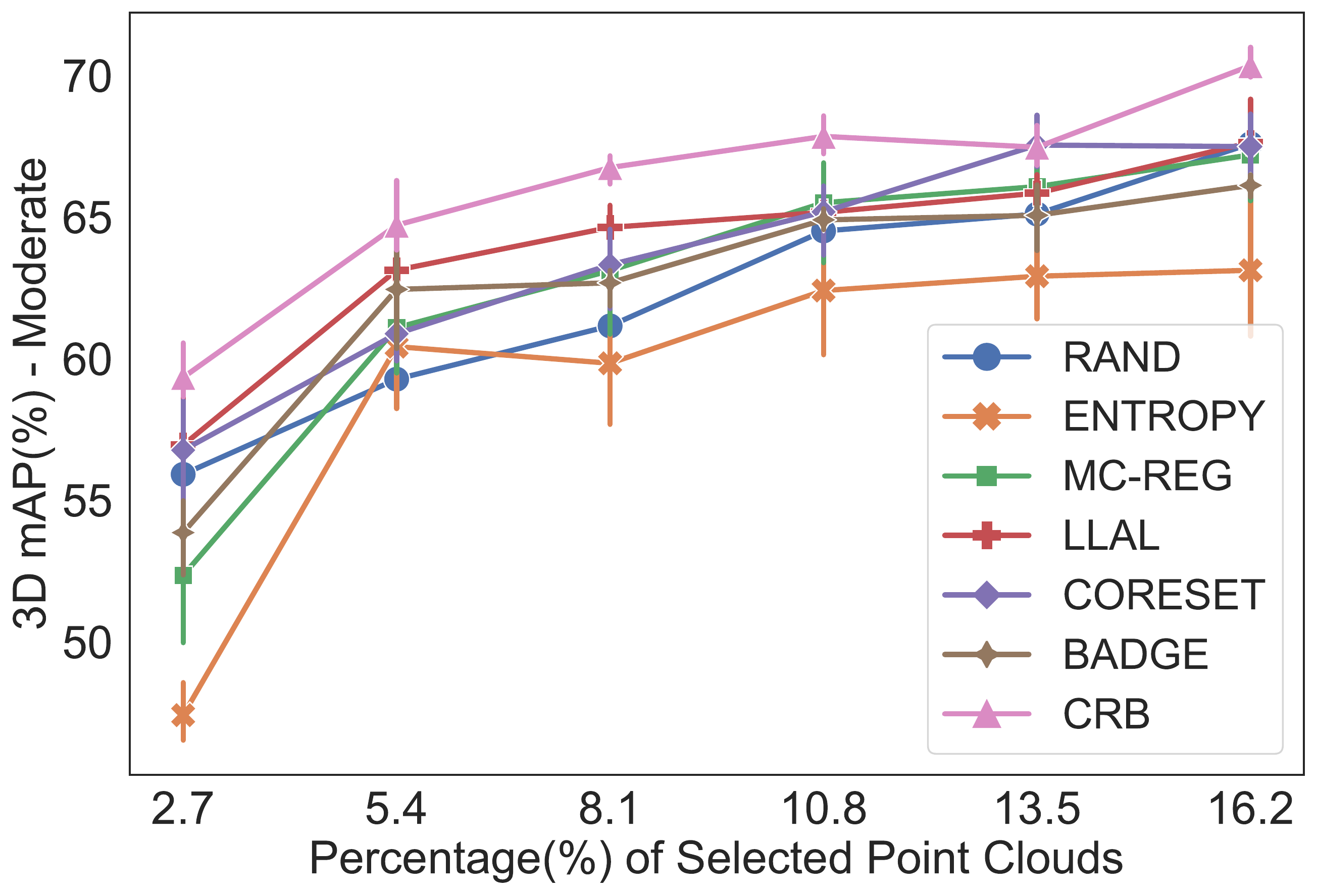} }}%
    \subfloat{{\includegraphics[width=0.33\textwidth]{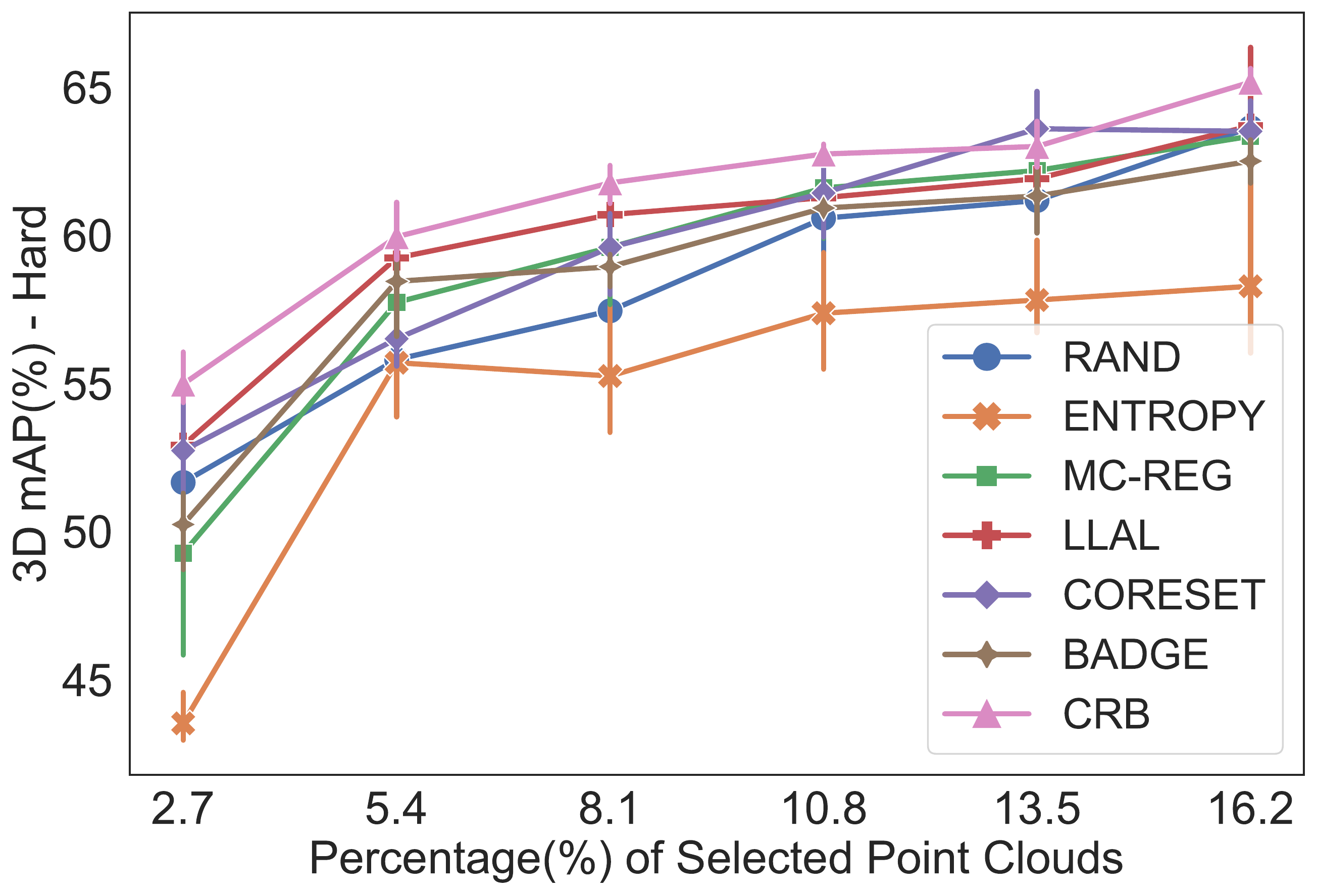} }}%
    \hfill
    \subfloat{{\includegraphics[width=0.33\textwidth]{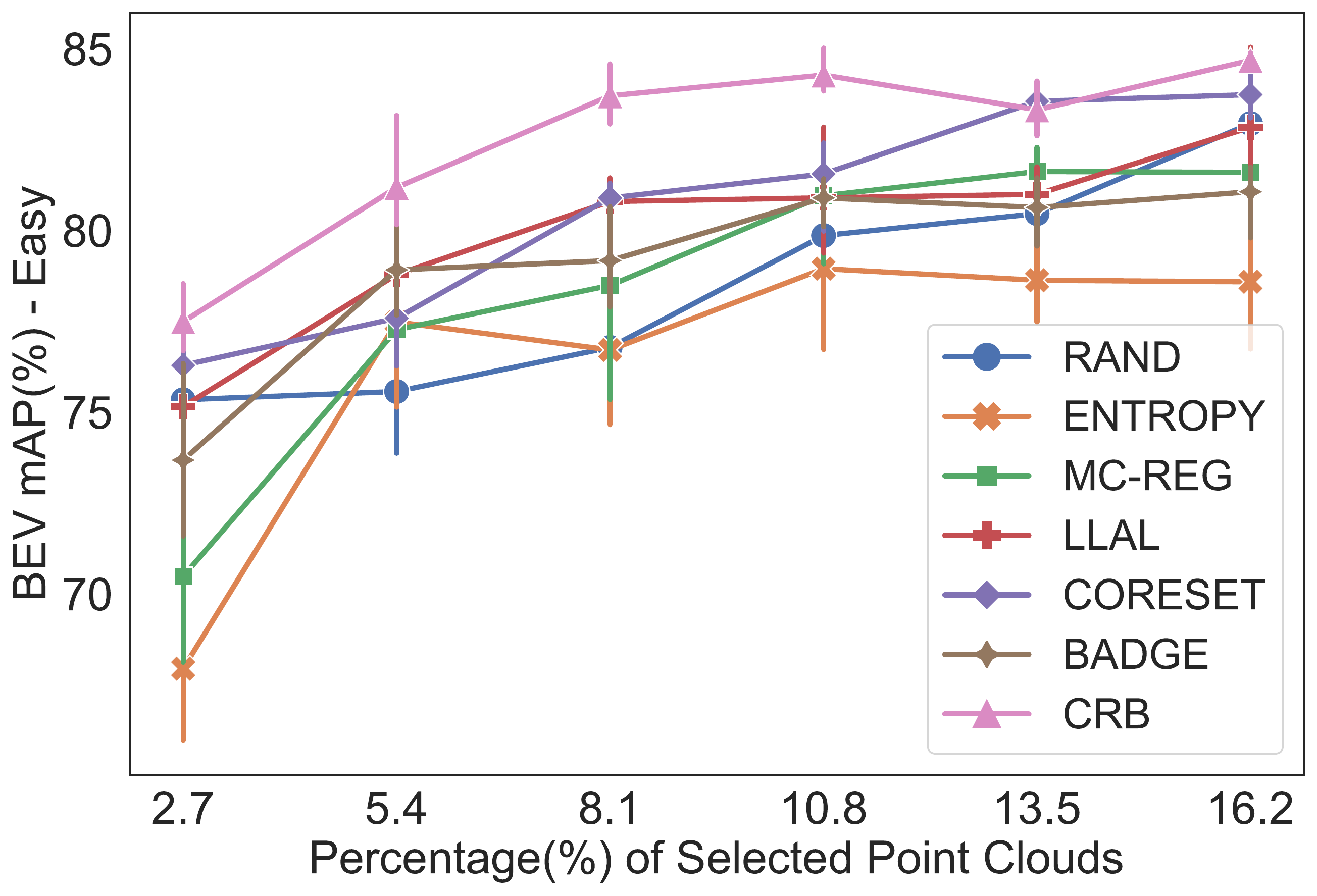} }}%
    \subfloat{{\includegraphics[width=0.33\textwidth]{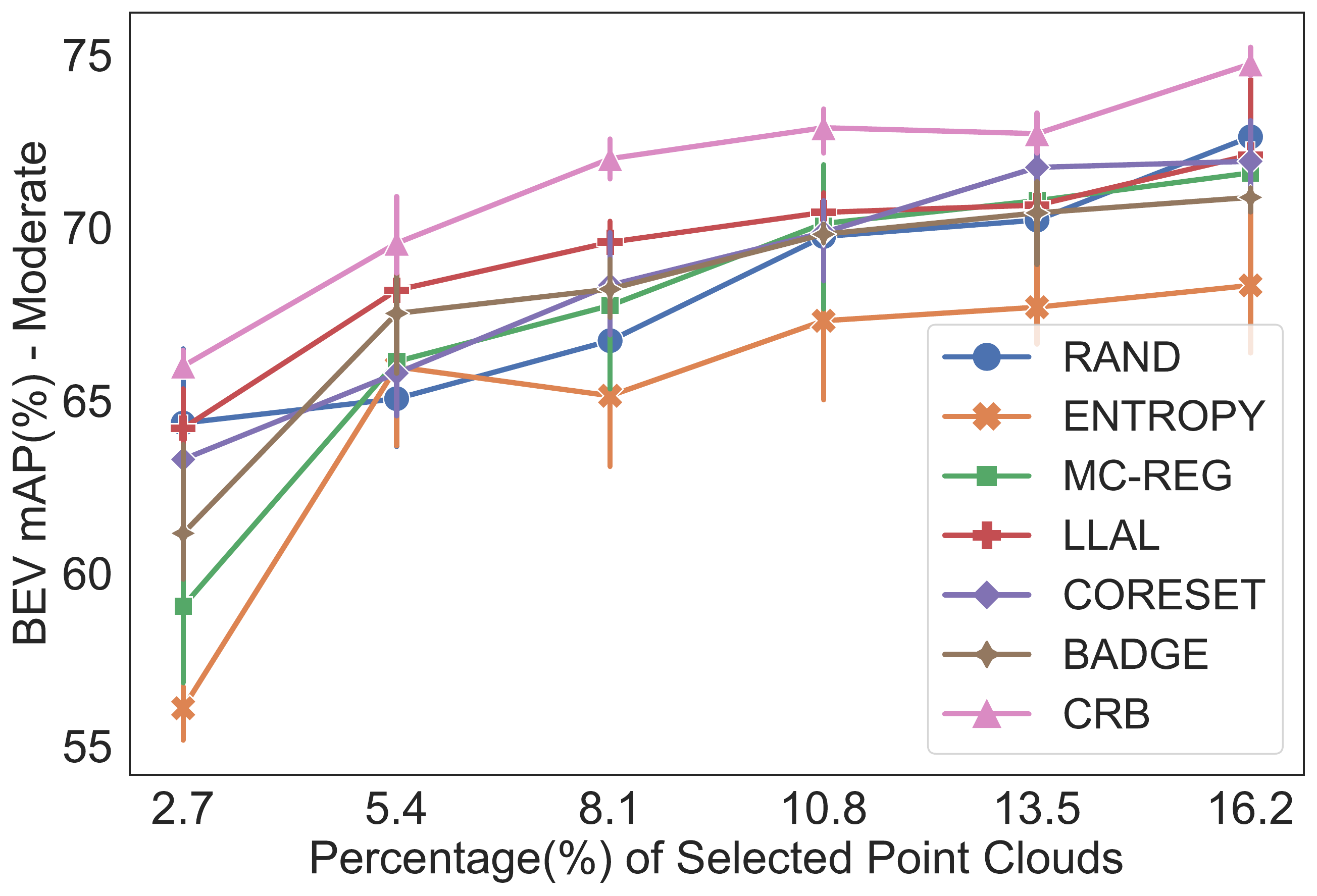} }}%
    \subfloat{{\includegraphics[width=0.33\textwidth]{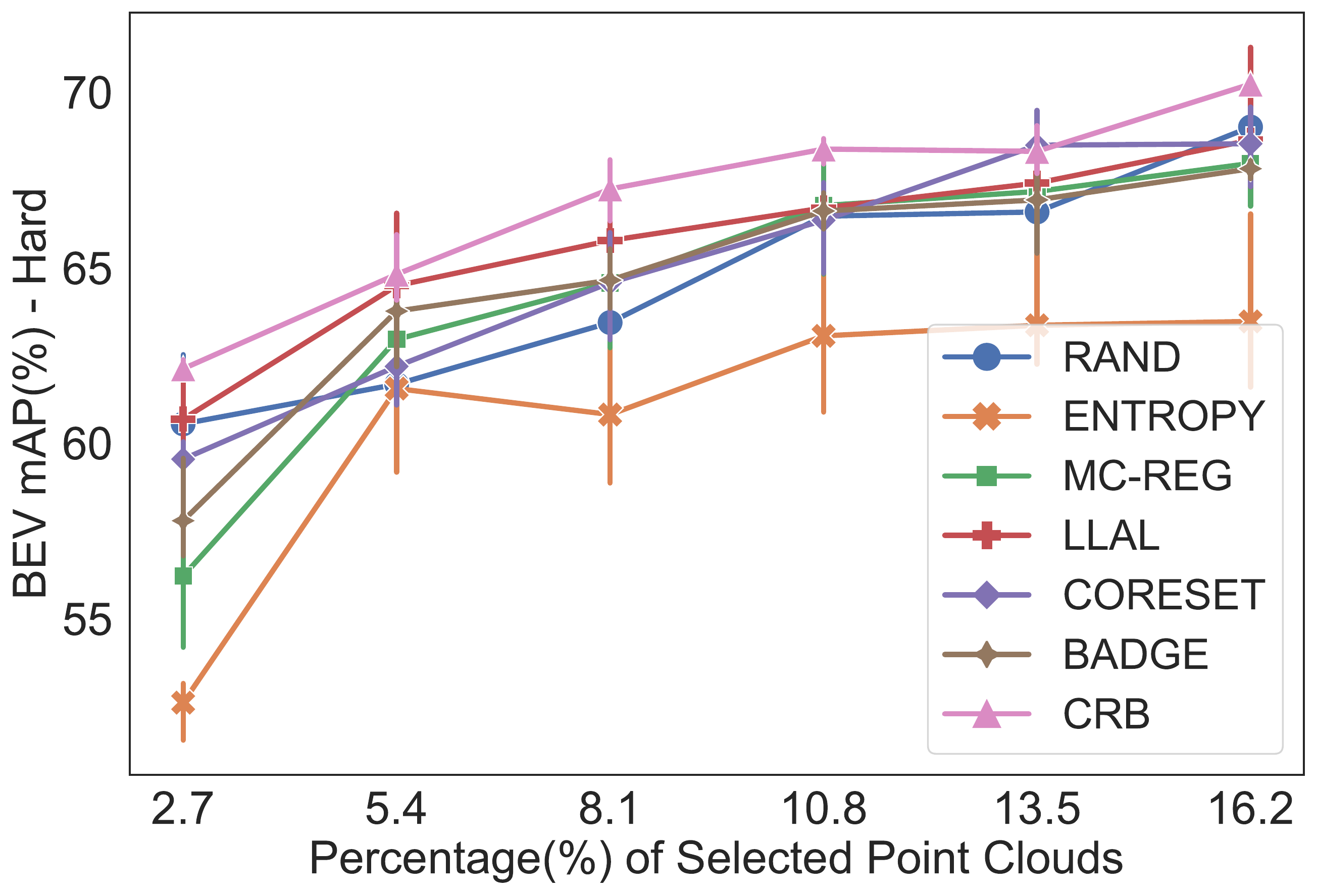} }}%
    \caption{Results on KITTI datasets with an increasing percentage of queried point clouds.}%
    \label{fig:kitti_results_pc}%
\end{figure}

Figure \ref{fig:kitti_results_pc} compares the performance variation of the AL baselines against the proposed \textsc{Crb} with the increasing percentage of queried point clouds (from 2.7\% to 16.2\%). The reported performance is mAP scores (\%) $\pm$ the standard deviation of three trials for both 3D view (top row) and BEV view (bottom row) and all difficulty levels. We clearly observe that our method \textsc{Crb} consistently outperforms the state-of-the-art results, irrespective of percentage of annotated point clouds and difficulty settings. Surprisingly, when the annotation costs reaches 16.2\%, \textsc{Rand} strategy outperforms all the baselines at the \textsc{Moderate} and \textsc{Hard} level. This implicitly evidences that existing uncertainty and diversity-based AL strategies fail to select samples that are aligned with test cases. 



\begin{figure}[t]%
    \subfloat{{\includegraphics[width=0.24\textwidth]{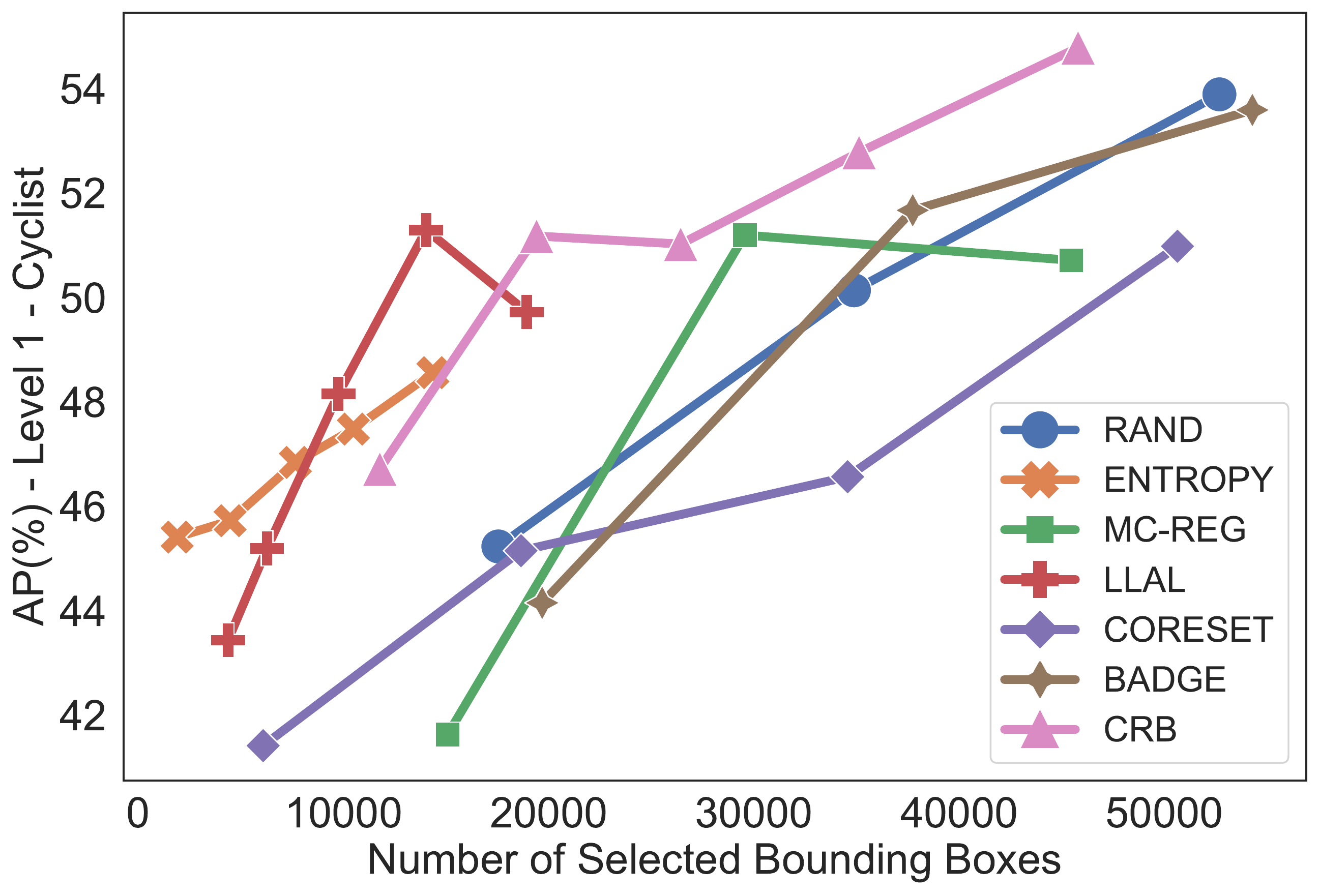} }}%
    \subfloat{{\includegraphics[width=0.24\textwidth]{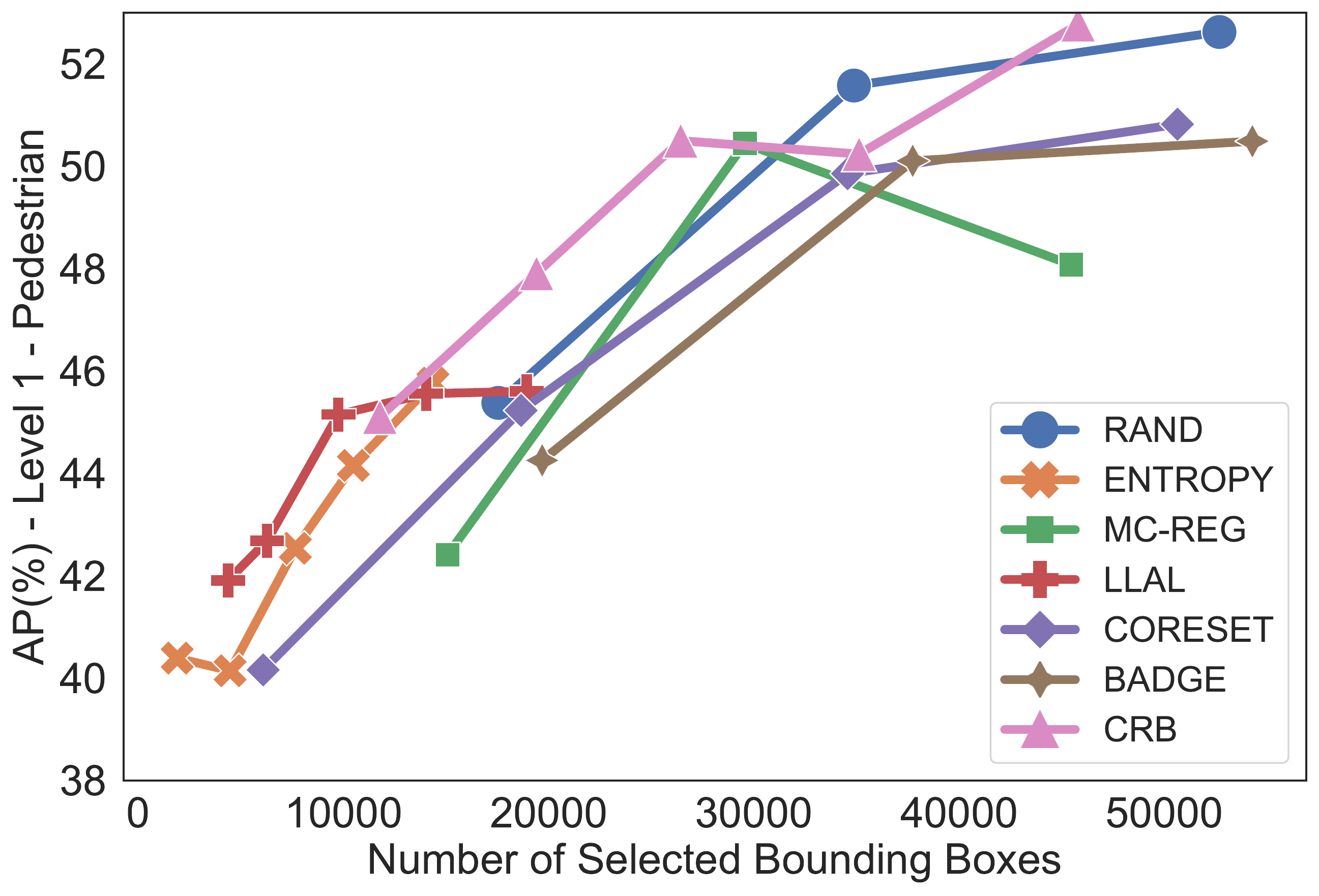} }}%
    \subfloat{{\includegraphics[width=0.24\textwidth]{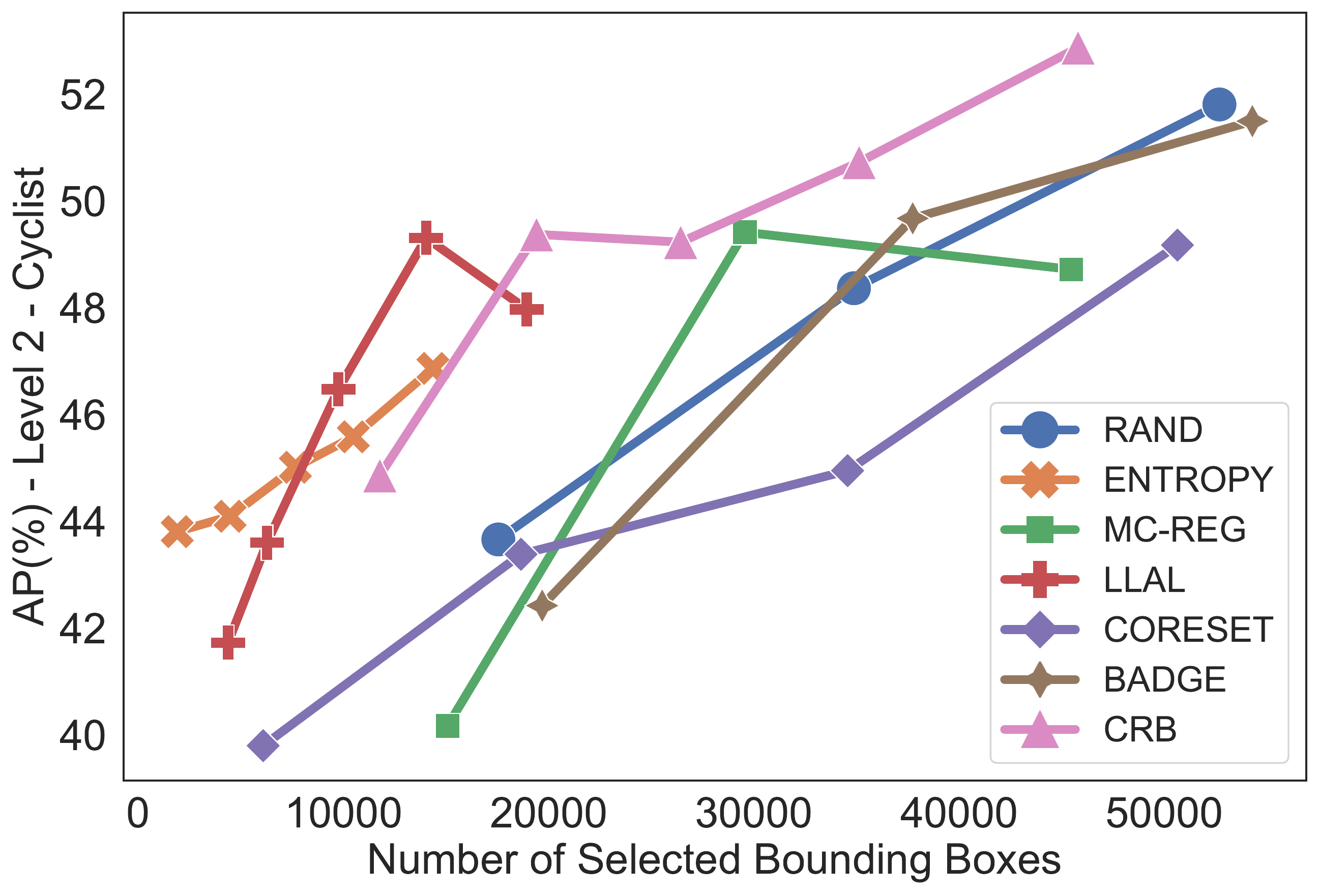} }}%
    \subfloat{{\includegraphics[width=0.24\textwidth]{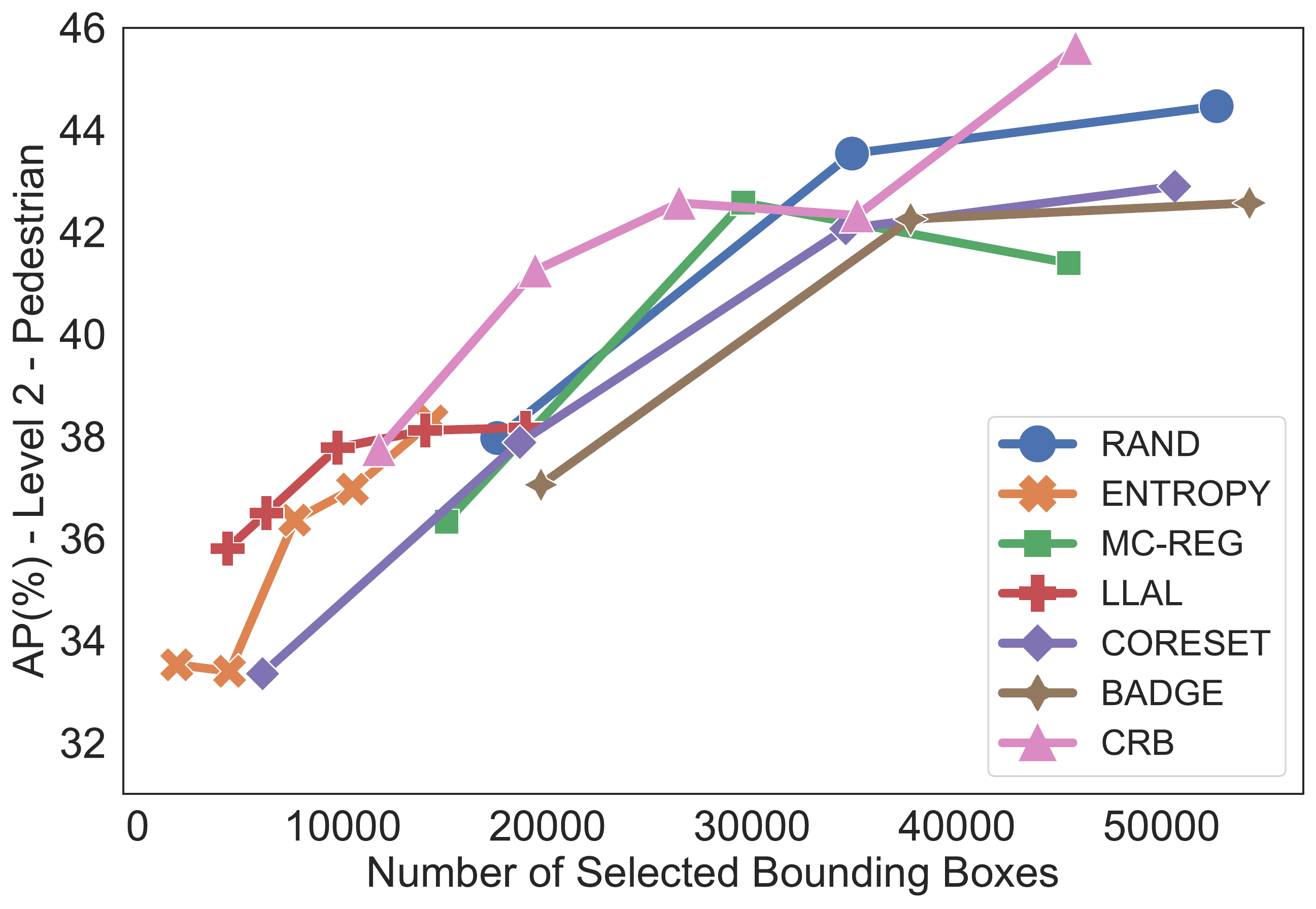} }}%
    \caption{Results of \textsc{Crb} and baselines on the Waymo \textit{val} splt for different classes at Level 2.}%
    \label{fig:waymo_results_cls_level_2}%
\end{figure}


\section{More Experimental Results on Waymo}\label{sec:waymo}
To explore the performance for different classes on the Waymo dataset, we plot the AP(\%) variation of Cyclist and Pedestrian yielded by the baselines and \textsc{Crb} with increasing annotated bounding boxes in Figure \ref{fig:waymo_results_cls_level_2}. We present the results at two levels of difficulty officially defined by Waymo. \textsc{Level 1} (and \textsc{Level 2}) indicates there are more than five inside points (at least one point) of the ground-truth objects. As can be observed by the AP curves in the plots, \textsc{Crb} achieves the superiror recognition accuracy when the annotation cost comes to $\sim$ 45k bounding boxes. Specifically, the AP values of \textsc{Crb} are boosted by the largest margin (3.1\% on \textsc{Level 2} Cyclist and 1.6\% on \textsc{Level 2} Pedestrian) over the best performing baseline (\textsc{Rand}) that takes extra cost of 5k bounding boxes than ours. Surprisingly, note the results on the class of Pedestrian, the AP curves of most baselines except \textsc{Entropy} and \textsc{Llal} are bounded by \textsc{Rand}. The AP curves of \textsc{Entropy} and \textsc{Llal} are bounded by \textsc{Crb} with the increasing cost to 15k $\sim$ 20k bounding boxes. This confirms the \textsc{Crb}’s superiority over compared AL baselines. Besides, the boosted margin achieved by \textsc{Crb} set for \textsc{Level 2} Pedestrian is larger than set for \textsc{Level 1} Pedestrian. This indicates that the samples selected by \textsc{Crb} matches well with the data at the time, covering more diverse samples that span different difficulties.

\section{Additional Qualitative Analysis}\label{sec:vis}

To intuitively demonstrate the benefits of our proposed active 3D detection strategy, Figure \ref{fig:3d_vis} visualizes that the 3D detection results produced by \textbf{\textsc{Rand}} (bottom left) and \textbf{\textsc{Crb}} selection (bottom right) from the corresponding image (upper row). Both 3D detectors are trained under the budget of 1K annotated bounding boxes. False positives and corrected predictions are indicated with red and green boxes. It is observed that, under the same condition, \textsc{Crb} produces more accurate and more confident predictions than \textsc{Rand}. Specifically, our \textsc{Crb} yields accurate predictions for multiple pedestrians on the right sidewalk, while \textsc{Rand} fails. Besides, note the car parked on the left that is highlighted in the orange box in Figure \ref{fig:3d_vis}, the detector trained with \textsc{Rand} produces a significantly lower confidence score ($0.62$) compared to our approach ($0.95$). This validates that the point clouds selected by \textsc{Crb} are aligned more tightly with the test samples.

\begin{figure}[h]
    \centering
    \includegraphics[width=1\linewidth]{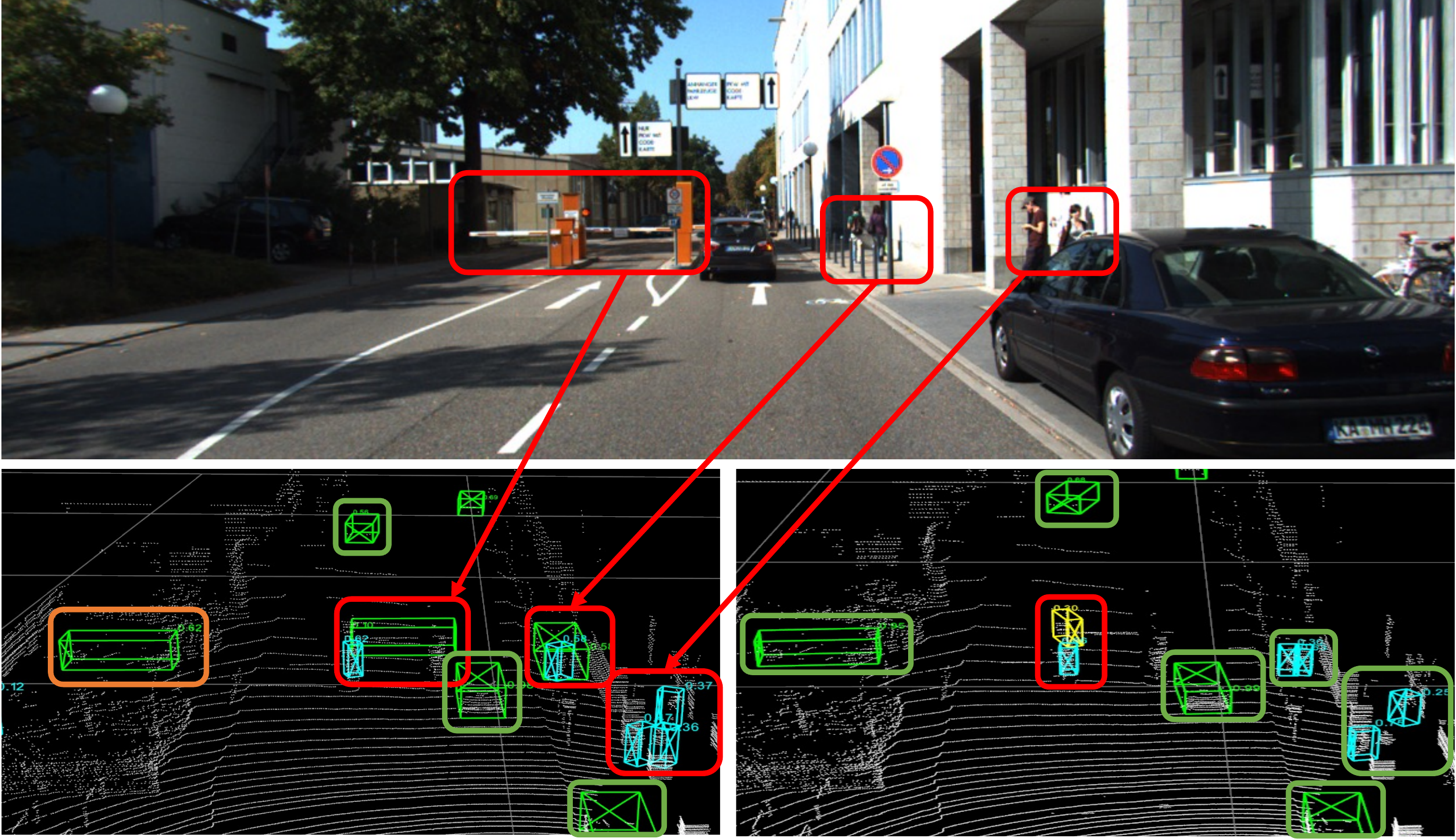}
    \caption{Another case study of active 3D detection performance of \textbf{\textsc{Rand}} (bottom left) and \textbf{\textsc{Crb}} (bottom right) under the budge of 1,000 annotated bounding boxes. False positive (corrected predictions) are highlighted in red (green) boxes. The orange box denotes the detection with low confidence.}
    \label{fig:3d_vis}
\end{figure}
\begin{figure}[h]
  \begin{minipage}[c]{0.72\textwidth}
    \subfloat{{\includegraphics[width=0.495\textwidth]{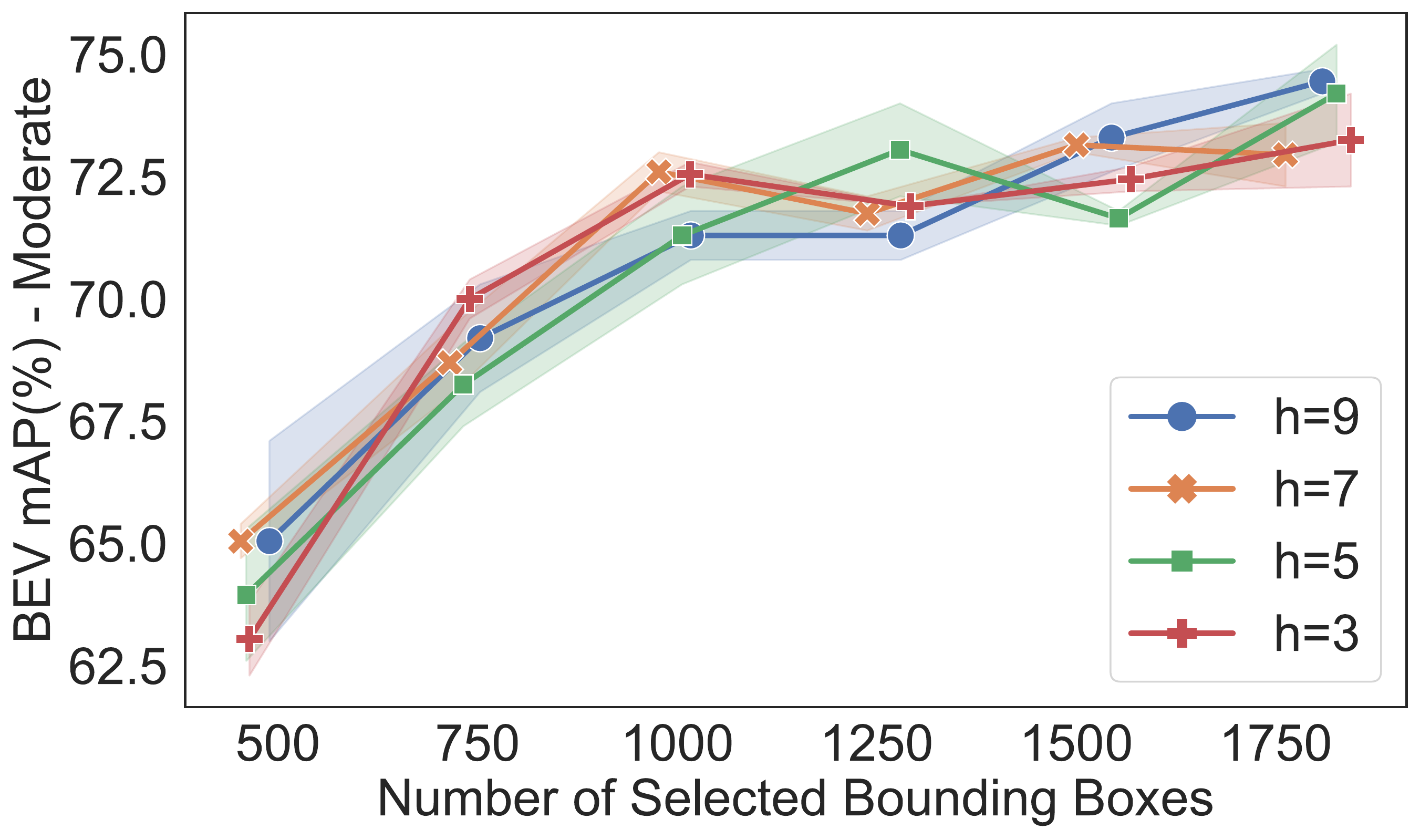} }}%
    \subfloat{{\includegraphics[width=0.495\textwidth]{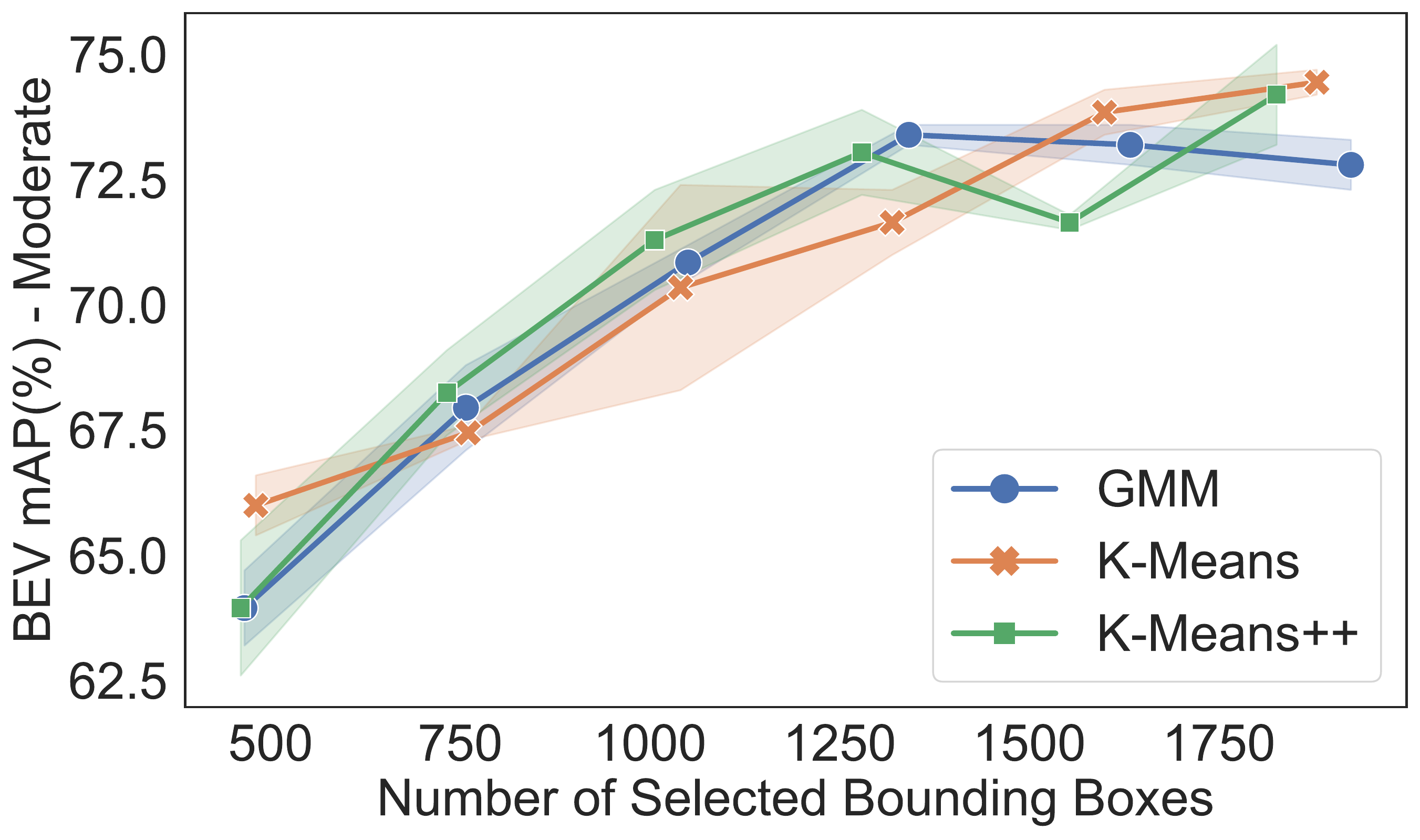} }}%
  \end{minipage}\hfill
  \begin{minipage}[c]{0.27\textwidth}
  \vspace{1ex}
    \caption{
       Performance comparison on KITTI \textit{val} set with varying KDE bandwidth $h$ (left) and prototype selection approaches (right) with increasing queried bounding boxes.
    } \label{fig:param}
  \end{minipage}\vspace{-2ex}
\end{figure}

\section{Additional Results for Parameter Sensitivity Analysis}\label{sec:param}

\noindent\textbf{Sensitivity to Prototype Selection.} To further analyze the sensitivity of performance to different prototype selection approaches, \textit{i.e.},  \textsc{Gmm}, \textsc{K-means}, and \textsc{K-means}++, we show more results on BEV views in Figure \ref{fig:param} (right). We again run two trials for each prototype selection method and plot the mean and the variance bars. Note that there is very little difference ($1.65\%$ in the last round) in the mAP(\%) of our approach when using different prototype selection methods. This evidences that the more performance gains achieved by \textsc{Crb} than existing baselines do not depend on choosing the prototype selection method.

\noindent\textbf{Sensitivity to Bandwidth $h$.} Figure \ref{fig:param} shows additional results w.r.t the BEV views of \textsc{Crb} with the bandwidth $h$ varying in $\{3, 5, 7, 9\}$. Observing the trends of four curves, \textsc{Crb} with the bandwidth of all values yields consistent results within the 1.7\% variation. This demonstrates that the \textsc{Crb} is insensitive to different values set for bandwidth and can produce similar mAP(\%) on BEV views.

\section{Related Work}\label{sec:rw}

\textbf{Generic Active Learning}. For a comprehensive review of classic active learning methods and their applications, we refer readers to \citep{10.1145/3472291}. Most active learning approaches were tailored for image classification task, where the \textit{uncertainty} \citep{DBLP:conf/ijcnn/WangS14, DBLP:conf/ijcnn/WangS14, DBLP:conf/icml/LewisC94,5206627margin, roth2006margin, Parvaneh_2022_CVPR_feature_mix, DBLP:conf/iccv/DuZCC0021, DBLP:conf/nips/KimSJM21, DBLP:conf/bmvc/BhatnagarGTS21} and \textit{diversity} \citep{DBLP:conf/iclr/SenerS18, DBLP:conf/iccv/ElhamifarSYS13, DBLP:conf/nips/Guo10, DBLP:journals/ijcv/YangMNCH15, DBLP:conf/icml/NguyenS04, DBLP:conf/iccv/0003R15, DBLP:conf/cvpr/AodhaCKB14} of samples are measured as the acquisition criteria. The hybrid works \citep{DBLP:conf/cvpr/KimPKC21, DBLP:conf/nips/CitovskyDGKRRK21, DBLP:conf/iclr/AshZK0A20, DBLP:journals/neco/MacKay92b, DBLP:conf/iccv/LiuDZLDH21, DBLP:conf/nips/KirschAG19, DBLP:journals/corr/abs-1112-5745} combine both paradigms such as by measuring uncertainty as to the gradient magnitude \citep{DBLP:conf/iclr/AshZK0A20} at the final layer of neural networks and selecting gradients that span a diverse set of directions. In addition to the above two mainstream methods, \citep{DBLP:conf/nips/SettlesCR07, roy2001toward-optimal-active, freytag2014selecting, DBLP:conf/cvpr/YooK19} estimate the expected model changes or predicted losses as the sample importance. 

\textbf{Active Learning for 2D Detection}. Lately, the attention of AL has shifted from image classification to the task of object detection \citep{DBLP:conf/cvpr/SiddiquiVN20,DBLP:conf/miccai/LiY20}. Early work \citep{DBLP:conf/bmvc/RoyUN18} exploits the detection inconsistency of outputs among different convolution layers and leverages the query by committee approach to select informative samples. Concurrent work \citep{DBLP:conf/accv/KaoLS018} introduces the notion of localization tightness as the regression uncertainty, which is calculated by the overlapping area between region proposals and the final predictions of bounding boxes. Other uncertainty-based methods attempt to aggregate pixel-level scores for each image \citep{Aghdam_2019_ICCV}, reformulate detectors by adding Bayesian inference to estimate the uncertainty \citep{DBLP:conf/icra/HarakehSW20} or replace conventional detection head with the Gaussian mixture model to compute aleatoric and epistemic uncertainty \citep{DBLP:conf/iccv/ChoiELFA21}. {\color{black} A hybrid method \citep{DBLP:conf/cvpr/WuC022} considers image-level uncertainty calculated by entropy and instance-level diversity measured by the similarity to the prototypes. Lately, AL technique is leveraged for transfer learning by selecting a few uncertain labeled source bounding boxes with high transferability to the target domain, where the transferability is defined by domain discriminators  \citep{9548667,DBLP:journals/tci/Al-SaffarBBTGA21}. Inspired by neural architecture searching, \cite{DBLP:conf/bmvc/TangJWXZ0LX21} adopted the ‘swap-expand’ strategy to seek a suitable neural architecture including depth, resolution, and receptive fields at each active selection round. Recently, some works augment the weakly-supervised object detection (WSOD) with an active learning scheme. In WSOD, only image-level category labels are available during training. Some conventional AL methods such as predicted probability, probability margin are explored in \citep{wang2022weaklySupervisedObject}, while in \citep{vo2022activeLearningStrategies}, “box-in-box" is introduced to select images where two predicted boxes belong to the same category and the small one is “contained” in the larger one.} Nevertheless, it is not trivial to adapt all existing AL approaches for 2D detection as the ensemble learning and network modification leads to more model parameters to learn, which could be hardly affordable for 3D tasks.

\textbf{Active Learning for 3D Detection}. Active learning for 3D object detection has been relatively under-explored than other tasks, potentially due to its large-scale nature. Most existing works \citep{DBLP:conf/ivs/FengWRMD19, DBLP:conf/ivs/SchmidtRTK20} simply apply the off-the-shelf generic AL strategies and use hand-crafted heuristics including Shannon entropy \citep{DBLP:conf/ijcnn/WangS14}, ensemble \citep{DBLP:conf/cvpr/BeluchGNK18}, localization tightness \citep{DBLP:conf/accv/KaoLS018} and \textsc{Mc-dropout} \citep{DBLP:conf/icml/GalG16} for 3D detection learning. However, the abovementioned solutions base on the cost of labelling point clouds rather than the number of 3D bounding boxes, which inherently being biased to the point clouds containing more objects. However, in our work, the proposed \textsc{Crb} greedily search for the unique point clouds while maintaining the same marginal distribution for generalization, which implicitly quires objects to annotate without repetition and save labeling costs.

{\color{black}
\textbf{Active Learning for 3D Semantic Segmentation}. The adoption of active learning techniques has successfully reduced the significant burden of point-by-point human labeling in large-scale point cloud datasets.
Super-point \citep{DBLP:journals/corr/abs-2101-06931} is introduced to represent a spectral clustering containing points which are most likely belonging to the same category, then only super-points with high score are labeled at each round. An improved work \cite{DBLP:conf/mm/ShaoLLCYLX22} further encoded the super-points with a graph neural network, where the edges denote distance between super-points, and then projects the super-point features into the diversity space to select the most representative super-points. Another streaming of work \citep{DBLP:conf/iccv/WuLHLSHH21} is to obtain point labels for uncertain and diverse regions to prevent the high cost of labeling the entire point cloud. Although semantic segmentation and object detection are different vision tasks, both can benefit from active learning to substantially alleviate the manual labelling cost. }

\textbf{Connections to Semi-supervised Active Learning}.
Aiming at unifying unlabeled sample selection and model training, the concept of semi-supervised active learning \citep{DBLP:conf/interspeech/DrugmanPK16,DBLP:journals/cogsr/RheeEKAJ17,DBLP:conf/iccv/SinhaED19,DBLP:conf/eccv/GaoZYADP20,DBLP:conf/iccv/LiuDZLDH21,DBLP:conf/cvpr/KimPKC21,DBLP:conf/nips/KimSJM21,DBLP:conf/emnlp/ZhangP21,DBLP:conf/iccv/0003SKKTJS0Z21,DBLP:conf/cvpr/CaramalauBK21,DBLP:conf/nips/CitovskyDGKRRK21,elezi2022not_all, DBLP:conf/cvpr/GudovskiyHYT20} has been raised. \citep{DBLP:conf/interspeech/DrugmanPK16} combines the semi-supervised learning (SSL) and active learning (AL) for speech understanding that leverages the confidence score obtained from the posterior probabilities of decoded texts. \citep{DBLP:conf/iclr/SenerS18} incorporated a Ladder network for SSL during AL cycles, while the performance gains are marginal compared to the supervised counterpart. \citep{DBLP:conf/iccv/SinhaED19} trained a variational adversarial active learning (\textsc{Vaal}) model with both labeled and unlabeled data points, where the discriminator is able to estimate how representative each sample is from the pool. \citep{elezi2022not_all} proposed a combined strategy for training 2D object detection, which queries samples of high uncertainty and low robustness for supervised learning and takes full advantage of easy samples via auto-labeling. As our work is under the umbrella of the pool-based active learning, accessible unlabeled data are not used for model training in our setting, hereby the semi-supervised active learning algorithms were not considered in experimental comparisons.

\bibliography{main}

\begin{thebibliography}{86}
\providecommand{\natexlab}[1]{#1}
\providecommand{\url}[1]{\texttt{#1}}
\expandafter\ifx\csname urlstyle\endcsname\relax
  \providecommand{\doi}[1]{doi: #1}\else
  \providecommand{\doi}{doi: \begingroup \urlstyle{rm}\Url}\fi

\bibitem[Aghdam et~al.(2019)Aghdam, Gonzalez-Garcia, Weijer, and
  Lopez]{Aghdam_2019_ICCV}
Hamed~H. Aghdam, Abel Gonzalez-Garcia, Joost van~de Weijer, and Antonio~M.
  Lopez.
\newblock Active learning for deep detection neural networks.
\newblock In \emph{Proc. International Conference on Computer Vision (ICCV)},
  pp.\  3672--3680, 2019.

\bibitem[Ahmed et~al.(2018)Ahmed, Tan, Chew, Mamun, and
  Wong]{DBLP:conf/iros/AhmedTCMW18}
Syeda~Mariam Ahmed, Yan~Zhi Tan, Chee{-}Meng Chew, Abdullah~Al Mamun, and
  Fook~Seng Wong.
\newblock Edge and corner detection for unorganized 3d point clouds with
  application to robotic welding.
\newblock In \emph{Proc. International Conference on Intelligent Robots and
  Systems (IROS)}, pp.\  7350--7355, 2018.

\bibitem[Al{-}Saffar et~al.(2021)Al{-}Saffar, Bialkowski, Baktashmotlagh,
  Trakic, Guo, and Abbosh]{DBLP:journals/tci/Al-SaffarBBTGA21}
Ahmed Al{-}Saffar, Alina Bialkowski, Mahsa Baktashmotlagh, Adnan Trakic, Lei
  Guo, and Amin~M. Abbosh.
\newblock Closing the gap of simulation to reality in electromagnetic imaging
  of brain strokes via deep neural networks.
\newblock \emph{{IEEE} Transactions on Computational Imaging}, 7:\penalty0
  13--21, 2021.

\bibitem[Aodha et~al.(2014)Aodha, Campbell, Kautz, and
  Brostow]{DBLP:conf/cvpr/AodhaCKB14}
Oisin~Mac Aodha, Neill D.~F. Campbell, Jan Kautz, and Gabriel~J. Brostow.
\newblock Hierarchical subquery evaluation for active learning on a graph.
\newblock In \emph{Proc. IEEE Conference on Computer Vision and Pattern
  Recognition (CVPR)}, pp.\  564--571, 2014.

\bibitem[Ash et~al.(2020)Ash, Zhang, Krishnamurthy, Langford, and
  Agarwal]{DBLP:conf/iclr/AshZK0A20}
Jordan~T. Ash, Chicheng Zhang, Akshay Krishnamurthy, John Langford, and Alekh
  Agarwal.
\newblock Deep batch active learning by diverse, uncertain gradient lower
  bounds.
\newblock In \emph{Proc. International Conference on Learning Representations
  (ICLR)}, 2020.

\bibitem[Beluch et~al.(2018)Beluch, Genewein, N{\"{u}}rnberger, and
  K{\"{o}}hler]{DBLP:conf/cvpr/BeluchGNK18}
William~H. Beluch, Tim Genewein, Andreas N{\"{u}}rnberger, and Jan~M.
  K{\"{o}}hler.
\newblock The power of ensembles for active learning in image classification.
\newblock In \emph{Proc. IEEE Conference on Computer Vision and Pattern
  Recognition (CVPR)}, pp.\  9368--9377, 2018.

\bibitem[Ben{-}David et~al.(2010)Ben{-}David, Blitzer, Crammer, Kulesza,
  Pereira, and Vaughan]{DBLP:journals/ml/Ben-DavidBCKPV10}
Shai Ben{-}David, John Blitzer, Koby Crammer, Alex Kulesza, Fernando Pereira,
  and Jennifer~Wortman Vaughan.
\newblock A theory of learning from different domains.
\newblock \emph{Journal of Machine Learning}, 79\penalty0 (1-2):\penalty0
  151--175, 2010.

\bibitem[Bhatnagar et~al.(2021)Bhatnagar, Goyal, Tank, and
  Sethi]{DBLP:conf/bmvc/BhatnagarGTS21}
Shubhang Bhatnagar, Sachin Goyal, Darshan Tank, and Amit Sethi.
\newblock {PAL} : Pretext-based active learning.
\newblock In \emph{Proc. British Machine Vision Conference (BMVC)}, pp.\  195.
  {BMVA} Press, 2021.

\bibitem[Caramalau et~al.(2021)Caramalau, Bhattarai, and
  Kim]{DBLP:conf/cvpr/CaramalauBK21}
Razvan Caramalau, Binod Bhattarai, and Tae{-}Kyun Kim.
\newblock Sequential graph convolutional network for active learning.
\newblock In \emph{Proc. IEEE Conference on Computer Vision and Pattern
  Recognition (CVPR)}, pp.\  9583--9592, 2021.

\bibitem[Choi et~al.(2021)Choi, Elezi, Lee, Farabet, and
  Alvarez]{DBLP:conf/iccv/ChoiELFA21}
Jiwoong Choi, Ismail Elezi, Hyuk{-}Jae Lee, Cl{\'{e}}ment Farabet, and Jose~M.
  Alvarez.
\newblock Active learning for deep object detection via probabilistic modeling.
\newblock In \emph{Proc. International Conference on Computer Vision (ICCV)},
  pp.\  10244--10253, 2021.

\bibitem[Citovsky et~al.(2021)Citovsky, DeSalvo, Gentile, Karydas, Rajagopalan,
  Rostamizadeh, and Kumar]{DBLP:conf/nips/CitovskyDGKRRK21}
Gui Citovsky, Giulia DeSalvo, Claudio Gentile, Lazaros Karydas, Anand
  Rajagopalan, Afshin Rostamizadeh, and Sanjiv Kumar.
\newblock Batch active learning at scale.
\newblock In \emph{Proc. Annual Conference on Neural Information Processing
  (NeurIPS)}, pp.\  11933--11944, 2021.

\bibitem[Deng et~al.(2021)Deng, Qi, Najibi, Funkhouser, Zhou, and
  Anguelov]{DBLP:conf/nips/DengQNFZA21}
Boyang Deng, Charles~R. Qi, Mahyar Najibi, Thomas~A. Funkhouser, Yin Zhou, and
  Dragomir Anguelov.
\newblock Revisiting 3d object detection from an egocentric perspective.
\newblock In \emph{Proc. Annual Conference on Neural Information Processing
  (NeurIPS)}, pp.\  26066--26079, 2021.

\bibitem[Drugman et~al.(2016)Drugman, Pylkk{\"{o}}nen, and
  Kneser]{DBLP:conf/interspeech/DrugmanPK16}
Thomas Drugman, Janne Pylkk{\"{o}}nen, and Reinhard Kneser.
\newblock Active and semi-supervised learning in {ASR:} benefits on the
  acoustic and language models.
\newblock In Nelson Morgan (ed.), \emph{Interspeech Annual Conference of the
  International Speech Communication Association}, pp.\  2318--2322, 2016.

\bibitem[Du et~al.(2021)Du, Zhao, Chen, Chai, Chen, and
  Li]{DBLP:conf/iccv/DuZCC0021}
Pan Du, Suyun Zhao, Hui Chen, Shuwen Chai, Hong Chen, and Cuiping Li.
\newblock Contrastive coding for active learning under class distribution
  mismatch.
\newblock In \emph{Proc. International Conference on Computer Vision (ICCV)},
  pp.\  8907--8916, 2021.

\bibitem[Elezi et~al.(2022)Elezi, Yu, Anandkumar, Leal-Taixe, and
  Alvarez]{elezi2022not_all}
Ismail Elezi, Zhiding Yu, Anima Anandkumar, Laura Leal-Taixe, and Jose~M
  Alvarez.
\newblock Not all labels are equal: Rationalizing the labeling costs for
  training object detection.
\newblock In \emph{Proceedings of the IEEE/CVF Conference on Computer Vision
  and Pattern Recognition}, pp.\  14492--14501, 2022.

\bibitem[Elhamifar et~al.(2013)Elhamifar, Sapiro, Yang, and
  Sastry]{DBLP:conf/iccv/ElhamifarSYS13}
Ehsan Elhamifar, Guillermo Sapiro, Allen~Y. Yang, and S.~Shankar Sastry.
\newblock A convex optimization framework for active learning.
\newblock In \emph{Proc. International Conference on Computer Vision (ICCV)},
  pp.\  209--216, 2013.

\bibitem[Feng et~al.(2019)Feng, Wei, Rosenbaum, Maki, and
  Dietmayer]{DBLP:conf/ivs/FengWRMD19}
Di~Feng, Xiao Wei, Lars Rosenbaum, Atsuto Maki, and Klaus Dietmayer.
\newblock Deep active learning for efficient training of a lidar 3d object
  detector.
\newblock In \emph{Proc. Intelligent Vehicles Symposium, (IV)}, pp.\  667--674,
  2019.

\bibitem[Freund et~al.(1992)Freund, Seung, Shamir, and
  Tishby]{DBLP:conf/nips/FreundSST92}
Yoav Freund, H.~Sebastian Seung, Eli Shamir, and Naftali Tishby.
\newblock Information, prediction, and query by committee.
\newblock In \emph{Proc. Annual Conference on Neural Information Processing
  (NeurIPS)}, pp.\  483--490, 1992.

\bibitem[Freytag et~al.(2014)Freytag, Rodner, and
  Denzler]{freytag2014selecting}
Alexander Freytag, Erik Rodner, and Joachim Denzler.
\newblock Selecting influential examples: Active learning with expected model
  output changes.
\newblock In \emph{Proc. European Conference on Computer Vision (ECCV)}, pp.\
  562--577, 2014.

\bibitem[Gal \& Ghahramani(2016)Gal and Ghahramani]{DBLP:conf/icml/GalG16}
Yarin Gal and Zoubin Ghahramani.
\newblock Dropout as a bayesian approximation: Representing model uncertainty
  in deep learning.
\newblock In \emph{Proc. International Conference on Machine Learning (ICML)},
  volume~48, pp.\  1050--1059, 2016.

\bibitem[Gal et~al.(2017)Gal, Islam, and Ghahramani]{DBLP:conf/icml/GalIG17}
Yarin Gal, Riashat Islam, and Zoubin Ghahramani.
\newblock Deep bayesian active learning with image data.
\newblock In \emph{Proc. International Conference on Machine Learning (ICML)},
  volume~70, pp.\  1183--1192, 2017.

\bibitem[Gao et~al.(2020)Gao, Zhang, Yu, Arik, Davis, and
  Pfister]{DBLP:conf/eccv/GaoZYADP20}
Mingfei Gao, Zizhao Zhang, Guo Yu, Sercan~{\"{O}}mer Arik, Larry~S. Davis, and
  Tomas Pfister.
\newblock Consistency-based semi-supervised active learning: Towards minimizing
  labeling cost.
\newblock In \emph{Proc. European Conference on Computer Vision (ECCV)}, volume
  12355, pp.\  510--526, 2020.

\bibitem[Geiger et~al.(2012)Geiger, Lenz, and
  Urtasun]{DBLP:conf/cvpr/GeigerLU12}
Andreas Geiger, Philip Lenz, and Raquel Urtasun.
\newblock Are we ready for autonomous driving? the {KITTI} vision benchmark
  suite.
\newblock In \emph{Proc. IEEE Conference on Computer Vision and Pattern
  Recognition (CVPR)}, pp.\  3354--3361, 2012.

\bibitem[Gudovskiy et~al.(2020)Gudovskiy, Hodgkinson, Yamaguchi, and
  Tsukizawa]{DBLP:conf/cvpr/GudovskiyHYT20}
Denis~A. Gudovskiy, Alec Hodgkinson, Takuya Yamaguchi, and Sotaro Tsukizawa.
\newblock Deep active learning for biased datasets via fisher kernel
  self-supervision.
\newblock In \emph{Proc. IEEE Conference on Computer Vision and Pattern
  Recognition (CVPR)}, pp.\  9038--9046, 2020.

\bibitem[Guo et~al.(2021)Guo, Shi, Kang, Kuang, Tang, Jiang, Sun, Wu, and
  Zhuang]{DBLP:conf/iccv/0003SKKTJS0Z21}
Jiannan Guo, Haochen Shi, Yangyang Kang, Kun Kuang, Siliang Tang, Zhuoren
  Jiang, Changlong Sun, Fei Wu, and Yueting Zhuang.
\newblock Semi-supervised active learning for semi-supervised models: Exploit
  adversarial examples with graph-based virtual labels.
\newblock In \emph{Proc. International Conference on Computer Vision (ICCV)},
  pp.\  2876--2885. {IEEE}, 2021.

\bibitem[Guo(2010)]{DBLP:conf/nips/Guo10}
Yuhong Guo.
\newblock Active instance sampling via matrix partition.
\newblock In \emph{Proc. Annual Conference on Neural Information Processing
  (NeurIPS)}, pp.\  802--810, 2010.

\bibitem[Harakeh et~al.(2020)Harakeh, Smart, and
  Waslander]{DBLP:conf/icra/HarakehSW20}
Ali Harakeh, Michael Smart, and Steven~L. Waslander.
\newblock Bayesod: {A} bayesian approach for uncertainty estimation in deep
  object detectors.
\newblock In \emph{Proc. International Conference on Robotics and Automation
  (ICRA)}, pp.\  87--93, 2020.

\bibitem[Hasan \& Roy{-}Chowdhury(2015)Hasan and
  Roy{-}Chowdhury]{DBLP:conf/iccv/0003R15}
Mahmudul Hasan and Amit~K. Roy{-}Chowdhury.
\newblock Context aware active learning of activity recognition models.
\newblock In \emph{Proc. International Conference on Computer Vision (ICCV)},
  pp.\  4543--4551, 2015.

\bibitem[Houlsby et~al.(2011)Houlsby, Huszar, Ghahramani, and
  Lengyel]{DBLP:journals/corr/abs-1112-5745}
Neil Houlsby, Ferenc Huszar, Zoubin Ghahramani, and M{\'{a}}t{\'{e}} Lengyel.
\newblock Bayesian active learning for classification and preference learning.
\newblock \emph{CoRR}, abs/1112.5745, 2011.

\bibitem[Joshi et~al.(2009)Joshi, Porikli, and Papanikolopoulos]{5206627margin}
Ajay~J. Joshi, Fatih Porikli, and Nikolaos Papanikolopoulos.
\newblock Multi-class active learning for image classification.
\newblock In \emph{Proc. IEEE Conference on Computer Vision and Pattern
  Recognition (CVPR)}, pp.\  2372--2379, 2009.

\bibitem[Kao et~al.(2018)Kao, Lee, Sen, and Liu]{DBLP:conf/accv/KaoLS018}
Chieh{-}Chi Kao, Teng{-}Yok Lee, Pradeep Sen, and Ming{-}Yu Liu.
\newblock Localization-aware active learning for object detection.
\newblock In \emph{Proc. Asian Conference on Computer (ACCV)}, pp.\  506--522,
  2018.

\bibitem[Kifer et~al.(2004)Kifer, Ben{-}David, and
  Gehrke]{DBLP:conf/vldb/KiferBG04}
Daniel Kifer, Shai Ben{-}David, and Johannes Gehrke.
\newblock Detecting change in data streams.
\newblock In \emph{(e)Proceedings of International Conference on Very Large
  Data Bases ((VLDB)}, pp.\  180--191. Morgan Kaufmann, 2004.

\bibitem[Kim et~al.(2021{\natexlab{a}})Kim, Park, Kim, and
  Chun]{DBLP:conf/cvpr/KimPKC21}
Kwanyoung Kim, Dongwon Park, Kwang~In Kim, and Se~Young Chun.
\newblock Task-aware variational adversarial active learning.
\newblock In \emph{Proc. IEEE Conference on Computer Vision and Pattern
  Recognition (CVPR)}, pp.\  8166--8175, 2021{\natexlab{a}}.

\bibitem[Kim et~al.(2021{\natexlab{b}})Kim, Song, Jang, and
  Moon]{DBLP:conf/nips/KimSJM21}
Yoon{-}Yeong Kim, Kyungwoo Song, JoonHo Jang, and Il{-}Chul Moon.
\newblock {LADA:} look-ahead data acquisition via augmentation for deep active
  learning.
\newblock In \emph{Proc. Annual Conference on Neural Information Processing
  (NeurIPS)}, pp.\  22919--22930, 2021{\natexlab{b}}.

\bibitem[Kirsch et~al.(2019)Kirsch, van Amersfoort, and
  Gal]{DBLP:conf/nips/KirschAG19}
Andreas Kirsch, Joost van Amersfoort, and Yarin Gal.
\newblock Batchbald: Efficient and diverse batch acquisition for deep bayesian
  active learning.
\newblock In \emph{Proc. Annual Conference on Neural Information Processing
  (NeurIPS)}, pp.\  7024--7035, 2019.

\bibitem[Lewis \& Catlett(1994)Lewis and Catlett]{DBLP:conf/icml/LewisC94}
David~D. Lewis and Jason Catlett.
\newblock Heterogeneous uncertainty sampling for supervised learning.
\newblock In \emph{Proc. International Conference on Machine Learning (ICML)},
  pp.\  148--156, 1994.

\bibitem[Li \& Yin(2020)Li and Yin]{DBLP:conf/miccai/LiY20}
Haohan Li and Zhaozheng Yin.
\newblock Attention, suggestion and annotation: {A} deep active learning
  framework for biomedical image segmentation.
\newblock In Anne~L. Martel, Purang Abolmaesumi, Danail Stoyanov, Diana Mateus,
  Maria~A. Zuluaga, S.~Kevin Zhou, Daniel Racoceanu, and Leo Joskowicz (eds.),
  \emph{Proc. Medical Image Computing and Computer Assisted Intervention
  (MICCAI)}, volume 12261, pp.\  3--13, 2020.

\bibitem[Liu et~al.(2021)Liu, Ding, Zhong, Li, Dai, and
  He]{DBLP:conf/iccv/LiuDZLDH21}
Zhuoming Liu, Hao Ding, Huaping Zhong, Weijia Li, Jifeng Dai, and Conghui He.
\newblock Influence selection for active learning.
\newblock In \emph{Proc. International Conference on Computer Vision (ICCV)},
  pp.\  9254--9263, 2021.

\bibitem[Ma et~al.(2021)Ma, Zeng, McDuff, and Song]{DBLP:conf/iclr/MaZMS21}
Shuang Ma, Zhaoyang Zeng, Daniel McDuff, and Yale Song.
\newblock Active contrastive learning of audio-visual video representations.
\newblock In \emph{Proc. International Conference on Learning Representations
  (ICLR)}, 2021.

\bibitem[MacKay(1992)]{DBLP:journals/neco/MacKay92b}
David J.~C. MacKay.
\newblock Information-based objective functions for active data selection.
\newblock \emph{Journal of Neural Computation}, 4\penalty0 (4):\penalty0
  590--604, 1992.

\bibitem[Mansour et~al.(2009)Mansour, Mohri, and
  Rostamizadeh]{DBLP:conf/colt/MansourMR09}
Yishay Mansour, Mehryar Mohri, and Afshin Rostamizadeh.
\newblock Domain adaptation: Learning bounds and algorithms.
\newblock In \emph{Proc. Conference on Learning Theory (COLT)}, 2009.

\bibitem[Montes et~al.(2020)Montes, Louedec, Cielniak, and
  Duckett]{DBLP:conf/iros/MontesLCD20}
Hector~A. Montes, Justin~Le Louedec, Grzegorz Cielniak, and Tom Duckett.
\newblock Real-time detection of broccoli crops in 3d point clouds for
  autonomous robotic harvesting.
\newblock In \emph{Proc. International Conference on Intelligent Robots and
  Systems (IROS)}, pp.\  10483--10488, 2020.

\bibitem[Nguyen \& Smeulders(2004)Nguyen and
  Smeulders]{DBLP:conf/icml/NguyenS04}
Hieu~Tat Nguyen and Arnold W.~M. Smeulders.
\newblock Active learning using pre-clustering.
\newblock In Carla~E. Brodley (ed.), \emph{Proc. International Conference on
  Machine Learning (ICML)}, 2004.

\bibitem[Oberman \& Calder(2018)Oberman and
  Calder]{DBLP:journals/corr/abs-1808-09540}
Adam~M. Oberman and Jeff Calder.
\newblock Lipschitz regularized deep neural networks converge and generalize.
\newblock \emph{CoRR}, abs/1808.09540, 2018.

\bibitem[Parvaneh et~al.(2022)Parvaneh, Abbasnejad, Teney, Haffari, van~den
  Hengel, and Shi]{Parvaneh_2022_CVPR_feature_mix}
Amin Parvaneh, Ehsan Abbasnejad, Damien Teney, Gholamreza~(Reza) Haffari, Anton
  van~den Hengel, and Javen~Qinfeng Shi.
\newblock Active learning by feature mixing.
\newblock In \emph{Proc. IEEE Conference on Computer Vision and Pattern
  Recognition (CVPR)}, pp.\  12237--12246, 2022.

\bibitem[Pinsler et~al.(2019)Pinsler, Gordon, Nalisnick, and
  Hern{\'{a}}ndez{-}Lobato]{DBLP:conf/nips/Pinsler0NH19}
Robert Pinsler, Jonathan Gordon, Eric~T. Nalisnick, and Jos{\'{e}}~Miguel
  Hern{\'{a}}ndez{-}Lobato.
\newblock Bayesian batch active learning as sparse subset approximation.
\newblock In \emph{Proc. Annual Conference on Neural Information Processing
  (NeurIPS)}, pp.\  6356--6367, 2019.

\bibitem[Qi et~al.(2020)Qi, Du, Siniscalchi, Ma, and
  Lee]{DBLP:journals/spl/QiDSML20}
Jun Qi, Jun Du, Sabato~Marco Siniscalchi, Xiaoli Ma, and Chin{-}Hui Lee.
\newblock On mean absolute error for deep neural network based vector-to-vector
  regression.
\newblock \emph{{IEEE} Signal Processing Letters}, 27:\penalty0 1485--1489,
  2020.

\bibitem[Ren et~al.(2021)Ren, Xiao, Chang, Huang, Li, Gupta, Chen, and
  Wang]{10.1145/3472291}
Pengzhen Ren, Yun Xiao, Xiaojun Chang, Po-Yao Huang, Zhihui Li, Brij~B. Gupta,
  Xiaojiang Chen, and Xin Wang.
\newblock A survey of deep active learning.
\newblock \emph{ACM Computing Survey}, 54\penalty0 (9):\penalty0 40, 2021.

\bibitem[Rhee et~al.(2017)Rhee, Erdenee, Kyun, Ahmed, and
  Jin]{DBLP:journals/cogsr/RheeEKAJ17}
Phill{-}Kyu Rhee, Enkhbayar Erdenee, Shin~Dong Kyun, Minhaz~Uddin Ahmed, and
  SongGuo Jin.
\newblock Active and semi-supervised learning for object detection with
  imperfect data.
\newblock \emph{Cognition System Research}, 45:\penalty0 109--123, 2017.

\bibitem[Roth \& Small(2006)Roth and Small]{roth2006margin}
Dan Roth and Kevin Small.
\newblock Margin-based active learning for structured output spaces.
\newblock In \emph{Proc. European Conference on Machine Learning (ECML)}, pp.\
  413--424, 2006.

\bibitem[Roy \& McCallum(2001{\natexlab{a}})Roy and
  McCallum]{DBLP:conf/icml/RoyM01}
Nicholas Roy and Andrew McCallum.
\newblock Toward optimal active learning through sampling estimation of error
  reduction.
\newblock In \emph{Proc. International Conference on Machine Learning (ICML)},
  pp.\  441--448, 2001{\natexlab{a}}.

\bibitem[Roy \& McCallum(2001{\natexlab{b}})Roy and
  McCallum]{roy2001toward-optimal-active}
Nicholas Roy and Andrew McCallum.
\newblock Toward optimal active learning through monte carlo estimation of
  error reduction.
\newblock In \emph{Proc. International Conference on Machine Learning (ICML)},
  pp.\  441--448, 2001{\natexlab{b}}.

\bibitem[Roy et~al.(2018)Roy, Unmesh, and Namboodiri]{DBLP:conf/bmvc/RoyUN18}
Soumya Roy, Asim Unmesh, and Vinay~P. Namboodiri.
\newblock Deep active learning for object detection.
\newblock In \emph{Proc. British Machine Vision Conference (BMVC)}, pp.\ ~91,
  2018.

\bibitem[Schmidt et~al.(2020)Schmidt, Rao, Tatsch, and
  Knoll]{DBLP:conf/ivs/SchmidtRTK20}
Sebastian Schmidt, Qing Rao, Julian Tatsch, and Alois~C. Knoll.
\newblock Advanced active learning strategies for object detection.
\newblock In \emph{Proc. Intelligent Vehicles Symposium, (IV)}, pp.\  871--876,
  2020.

\bibitem[Sener \& Savarese(2018)Sener and Savarese]{DBLP:conf/iclr/SenerS18}
Ozan Sener and Silvio Savarese.
\newblock Active learning for convolutional neural networks: {A} core-set
  approach.
\newblock In \emph{Proc. International Conference on Learning Representations
  (ICLR)}, 2018.

\bibitem[Settles et~al.(2007)Settles, Craven, and
  Ray]{DBLP:conf/nips/SettlesCR07}
Burr Settles, Mark Craven, and Soumya Ray.
\newblock Multiple-instance active learning.
\newblock In \emph{Proc. Annual Conference on Neural Information Processing
  (NeurIPS)}, pp.\  1289--1296, 2007.

\bibitem[Shannon(1948)]{DBLP:journals/bstj/Shannon48}
Claude~E. Shannon.
\newblock A mathematical theory of communication.
\newblock \emph{The Bell System Technical Journal}, 27\penalty0 (3):\penalty0
  379--423, 1948.

\bibitem[Shao et~al.(2022)Shao, Luo, Liu, Chen, Yang, Lu, and
  Xiao]{DBLP:conf/mm/ShaoLLCYLX22}
Feifei Shao, Yawei Luo, Ping Liu, Jie Chen, Yi~Yang, Yulei Lu, and Jun Xiao.
\newblock Active learning for point cloud semantic segmentation via
  spatial-structural diversity reasoning.
\newblock In \emph{Proc. International Conference on Multimedia (MM)}, pp.\
  2575--2585, 2022.

\bibitem[Shi \& Li(2019)Shi and Li]{DBLP:conf/ijcai/ShiL19}
Feng Shi and Yu{-}Feng Li.
\newblock Rapid performance gain through active model reuse.
\newblock In \emph{Proc. International Joint Conference on Artificial
  Intelligence (IJCAI)}, pp.\  3404--3410, 2019.

\bibitem[Shi et~al.(2019)Shi, Wang, and Li]{DBLP:conf/cvpr/ShiWL19}
Shaoshuai Shi, Xiaogang Wang, and Hongsheng Li.
\newblock Pointrcnn: 3d object proposal generation and detection from point
  cloud.
\newblock In \emph{Proc. IEEE Conference on Computer Vision and Pattern
  Recognition (CVPR)}, pp.\  770--779. Computer Vision Foundation / {IEEE},
  2019.

\bibitem[Shi et~al.(2020)Shi, Guo, Jiang, Wang, Shi, Wang, and
  Li]{DBLP:conf/cvpr/ShiGJ0SWL20}
Shaoshuai Shi, Chaoxu Guo, Li~Jiang, Zhe Wang, Jianping Shi, Xiaogang Wang, and
  Hongsheng Li.
\newblock {PV-RCNN:} point-voxel feature set abstraction for 3d object
  detection.
\newblock In \emph{Proc. IEEE Conference on Computer Vision and Pattern
  Recognition (CVPR)}, pp.\  10526--10535, 2020.

\bibitem[Shi \& Yu(2019)Shi and Yu]{DBLP:conf/nips/Shi019}
Weishi Shi and Qi~Yu.
\newblock Integrating bayesian and discriminative sparse kernel machines for
  multi-class active learning.
\newblock In \emph{Proc. Annual Conference on Neural Information Processing
  (NeurIPS)}, pp.\  2282--2291, 2019.

\bibitem[Shi et~al.(2021)Shi, Xu, Chen, Cai, Foo, and
  Jia]{DBLP:journals/corr/abs-2101-06931}
Xian Shi, Xun Xu, Ke~Chen, Lile Cai, Chuan~Sheng Foo, and Kui Jia.
\newblock Label-efficient point cloud semantic segmentation: An active learning
  approach.
\newblock \emph{CoRR}, abs/2101.06931, 2021.

\bibitem[Siddiqui et~al.(2020)Siddiqui, Valentin, and
  Nie{\ss}ner]{DBLP:conf/cvpr/SiddiquiVN20}
Yawar Siddiqui, Julien Valentin, and Matthias Nie{\ss}ner.
\newblock Viewal: Active learning with viewpoint entropy for semantic
  segmentation.
\newblock In \emph{Proc. IEEE Conference on Computer Vision and Pattern
  Recognition (CVPR)}, pp.\  9430--9440, 2020.

\bibitem[Sinha et~al.(2019)Sinha, Ebrahimi, and
  Darrell]{DBLP:conf/iccv/SinhaED19}
Samarth Sinha, Sayna Ebrahimi, and Trevor Darrell.
\newblock Variational adversarial active learning.
\newblock In \emph{Proc. International Conference on Computer Vision (ICCV)},
  pp.\  5971--5980, 2019.

\bibitem[Song et~al.(2015)Song, Lichtenberg, and Xiao]{DBLP:conf/cvpr/SongLX15}
Shuran Song, Samuel~P. Lichtenberg, and Jianxiong Xiao.
\newblock {SUN} {RGB-D:} {A} {RGB-D} scene understanding benchmark suite.
\newblock In \emph{Proc. IEEE Conference on Computer Vision and Pattern
  Recognition (CVPR)}, pp.\  567--576, 2015.

\bibitem[Sun et~al.(2020)Sun, Kretzschmar, Dotiwalla, Chouard, Patnaik, Tsui,
  Guo, Zhou, Chai, Caine, Vasudevan, Han, Ngiam, Zhao, Timofeev, Ettinger,
  Krivokon, Gao, Joshi, Zhang, Shlens, Chen, and
  Anguelov]{DBLP:conf/cvpr/SunKDCPTGZCCVHN20}
Pei Sun, Henrik Kretzschmar, Xerxes Dotiwalla, Aurelien Chouard, Vijaysai
  Patnaik, Paul Tsui, James Guo, Yin Zhou, Yuning Chai, Benjamin Caine, Vijay
  Vasudevan, Wei Han, Jiquan Ngiam, Hang Zhao, Aleksei Timofeev, Scott
  Ettinger, Maxim Krivokon, Amy Gao, Aditya Joshi, Yu~Zhang, Jonathon Shlens,
  Zhifeng Chen, and Dragomir Anguelov.
\newblock Scalability in perception for autonomous driving: Waymo open dataset.
\newblock In \emph{Proc. IEEE Conference on Computer Vision and Pattern
  Recognition (CVPR)}, pp.\  2443--2451, 2020.

\bibitem[Tang et~al.(2021{\natexlab{a}})Tang, Jiang, Wei, Xu, Zhang, Zhang, Lu,
  and Xu]{DBLP:conf/bmvc/TangJWXZ0LX21}
Fuhui Tang, Chenhan Jiang, Dafeng Wei, Hang Xu, Andi Zhang, Wei Zhang, Hongtao
  Lu, and Chunjing Xu.
\newblock Towards dynamic and scalable active learning with neural architecture
  adaption for object detection.
\newblock In \emph{Proc. British Machine Vision Conference (BMVC)},
  2021{\natexlab{a}}.

\bibitem[Tang et~al.(2021{\natexlab{b}})Tang, Wei, Zhao, and Huang]{9548667}
Ying-Peng Tang, Xiu-Shen Wei, Borui Zhao, and Sheng-Jun Huang.
\newblock Qbox: Partial transfer learning with active querying for object
  detection.
\newblock \emph{Journal of IEEE Transactions on Neural Networks and Learning
  Systems}, pp.\  1--13, 2021{\natexlab{b}}.
\newblock \doi{10.1109/TNNLS.2021.3111621}.

\bibitem[Tran et~al.(2019)Tran, Do, Reid, and
  Carneiro]{DBLP:conf/icml/TranDRC19}
Toan Tran, Thanh{-}Toan Do, Ian~D. Reid, and Gustavo Carneiro.
\newblock Bayesian generative active deep learning.
\newblock In \emph{Proc. International Conference on Machine Learning (ICML)},
  volume~97, pp.\  6295--6304, 2019.

\bibitem[Vo et~al.(2022)Vo, Sim{\'e}oni, Gidaris, Bursuc, P{\'e}rez, and
  Ponce]{vo2022activeLearningStrategies}
Huy~V Vo, Oriane Sim{\'e}oni, Spyros Gidaris, Andrei Bursuc, Patrick P{\'e}rez,
  and Jean Ponce.
\newblock Active learning strategies for weakly-supervised object detection.
\newblock In \emph{Proc. European Conference on Computer Vision (ECCV)}, pp.\
  211--230, 2022.

\bibitem[Wang \& Shang(2014)Wang and Shang]{DBLP:conf/ijcnn/WangS14}
Dan Wang and Yi~Shang.
\newblock A new active labeling method for deep learning.
\newblock In \emph{Proc. International Joint Conference on Neural Networks
  (IJCNN)}, pp.\  112--119, 2014.

\bibitem[Wang et~al.(2020)Wang, Lan, Gao, and Davis]{DBLP:conf/eccv/WangLGD20}
Jun Wang, Shiyi Lan, Mingfei Gao, and Larry~S. Davis.
\newblock Infofocus: 3d object detection for autonomous driving with dynamic
  information modeling.
\newblock In \emph{Proc. European Conference on Computer Vision (ECCV)}, volume
  12355, pp.\  405--420, 2020.

\bibitem[Wang et~al.(2017)Wang, Zhang, Li, Zhang, and
  Lin]{DBLP:journals/tcsv/WangZLZL17}
Keze Wang, Dongyu Zhang, Ya~Li, Ruimao Zhang, and Liang Lin.
\newblock Cost-effective active learning for deep image classification.
\newblock \emph{Journal of IEEE Transactions on Circuits and Systems for Video
  Technology}, 27\penalty0 (12):\penalty0 2591--2600, 2017.

\bibitem[Wang et~al.(2019)Wang, Li, Sun, Liu, Zhao, Seah, Quah, and
  Tandianus]{DBLP:journals/sensors/WangLSLZSQT19}
Li~Wang, Ruifeng Li, Jingwen Sun, Xingxing Liu, Lijun Zhao, Hock~Soon Seah,
  Chee~Kwang Quah, and Budianto Tandianus.
\newblock Multi-view fusion-based 3d object detection for robot indoor scene
  perception.
\newblock \emph{Sensors}, 19\penalty0 (19):\penalty0 4092, 2019.

\bibitem[Wang et~al.(2022)Wang, Xiang, Zhang, Liu, Zheng, and
  Hu]{wang2022weaklySupervisedObject}
Xiao Wang, Xiang Xiang, Baochang Zhang, Xuhui Liu, Jianying Zheng, and QingLei
  Hu.
\newblock Weakly supervised object detection based on active learning.
\newblock \emph{Journal of Neural Processing Letters}, pp.\  1--15, 2022.

\bibitem[Wu et~al.(2022)Wu, Chen, and Huang]{DBLP:conf/cvpr/WuC022}
Jiaxi Wu, Jiaxin Chen, and Di~Huang.
\newblock Entropy-based active learning for object detection with progressive
  diversity constraint.
\newblock In \emph{Proc. IEEE Conference on Computer Vision and Pattern
  Recognition (CVPR)}, pp.\  9387--9396. {IEEE}, 2022.

\bibitem[Wu et~al.(2021)Wu, Liu, Huang, Lee, Su, Huang, and
  Hsu]{DBLP:conf/iccv/WuLHLSHH21}
Tsung{-}Han Wu, Yueh{-}Cheng Liu, Yu{-}Kai Huang, Hsin{-}Ying Lee, Hung{-}Ting
  Su, Ping{-}Chia Huang, and Winston~H. Hsu.
\newblock Redal: Region-based and diversity-aware active learning for point
  cloud semantic segmentation.
\newblock In \emph{Proc. International Conference on Computer Vision (ICCV)},
  pp.\  15490--15499, 2021.

\bibitem[Yang et~al.(2015)Yang, Ma, Nie, Chang, and
  Hauptmann]{DBLP:journals/ijcv/YangMNCH15}
Yi~Yang, Zhigang Ma, Feiping Nie, Xiaojun Chang, and Alexander~G. Hauptmann.
\newblock Multi-class active learning by uncertainty sampling with diversity
  maximization.
\newblock \emph{International Journal of Computer Vision}, 113:\penalty0
  113--127, 2015.

\bibitem[Yang et~al.(2019)Yang, Sun, Liu, Shen, and
  Jia]{DBLP:conf/iccv/YangS0SJ19}
Zetong Yang, Yanan Sun, Shu Liu, Xiaoyong Shen, and Jiaya Jia.
\newblock {STD:} sparse-to-dense 3d object detector for point cloud.
\newblock In \emph{Proc. IEEE Conference on Computer Vision and Pattern
  Recognition (CVPR)}, pp.\  1951--1960. {IEEE}, 2019.

\bibitem[Yang et~al.(2020)Yang, Sun, Liu, and Jia]{DBLP:conf/cvpr/YangS0J20}
Zetong Yang, Yanan Sun, Shu Liu, and Jiaya Jia.
\newblock 3dssd: Point-based 3d single stage object detector.
\newblock In \emph{Proc. IEEE Conference on Computer Vision and Pattern
  Recognition (CVPR)}, pp.\  11037--11045. Computer Vision Foundation / {IEEE},
  2020.

\bibitem[Yoo \& Kweon(2019)Yoo and Kweon]{DBLP:conf/cvpr/YooK19}
Donggeun Yoo and In~So Kweon.
\newblock Learning loss for active learning.
\newblock In \emph{Proc. IEEE Conference on Computer Vision and Pattern
  Recognition (CVPR)}, pp.\  93--102, 2019.

\bibitem[Yuan et~al.(2021)Yuan, Wan, Fu, Liu, Xu, Ji, and
  Ye]{DBLP:conf/cvpr/YuanWFLXJY21}
Tianning Yuan, Fang Wan, Mengying Fu, Jianzhuang Liu, Songcen Xu, Xiangyang Ji,
  and Qixiang Ye.
\newblock Multiple instance active learning for object detection.
\newblock In \emph{Proc. IEEE Conference on Computer Vision and Pattern
  Recognition (CVPR)}, pp.\  5330--5339, 2021.

\bibitem[Zhang et~al.(2020)Zhang, Li, Yang, Wang, Zha, and
  Huang]{DBLP:conf/cvpr/Zhang0YWZH20}
Beichen Zhang, Liang Li, Shijie Yang, Shuhui Wang, Zheng{-}Jun Zha, and
  Qingming Huang.
\newblock State-relabeling adversarial active learning.
\newblock In \emph{Proc. IEEE Conference on Computer Vision and Pattern
  Recognition (CVPR)}, pp.\  8753--8762, 2020.

\bibitem[Zhang \& Plank(2021)Zhang and Plank]{DBLP:conf/emnlp/ZhangP21}
Mike Zhang and Barbara Plank.
\newblock Cartography active learning.
\newblock In Marie{-}Francine Moens, Xuanjing Huang, Lucia Specia, and
  Scott~Wen{-}tau Yih (eds.), \emph{Proc. Findings of the Association for
  Computational Linguistics (EMNLP)}, pp.\  395--406. Association for
  Computational Linguistics, 2021.

\bibitem[Zhang et~al.(2022)Zhang, Hu, Xu, Ma, Wan, and Guo]{IA-SSD}
Yifan Zhang, Qingyong Hu, Guoquan Xu, Yanxin Ma, Jianwei Wan, and Yulan Guo.
\newblock Not all points are equal: Learning highly efficient point-based
  detectors for 3d lidar point clouds.
\newblock In \emph{Proc. IEEE Conference on Computer Vision and Pattern
  Recognition (CVPR)}, pp.\  18953--18962, 2022.

\end{thebibliography}
\bibliographystyle{iclr2023_conference}

\appendix

\end{document}